\documentclass[a4paper,11pt]{article}

\usepackage{verbatim}

\usepackage{cite}
\usepackage{caption, subcaption}
\usepackage{mathtools}
\usepackage{float}
\usepackage{wrapfig,lipsum,booktabs}

\usepackage{fourier}
\usepackage{color}
 \usepackage{graphicx}
\usepackage{url}
\usepackage[affil-it]{authblk}
\usepackage{amsmath}
\usepackage{wrapfig}

\usepackage[T1]{fontenc}
\usepackage{times}

\begin{document}

\title{A Study on Visual Perception of Light Field Content}

\author{Ailbhe Gill, Emin Zerman, Cagri Ozcinar, Aljosa Smolic\thanks{This publication has emanated from research conducted with the financial support of Science Foundation Ireland (SFI) under the Grant Number 15/RP/27760.}}
%\author{Anonymous Authors}
\affil{V-SENSE, School of Computer Science, Trinity College Dublin, Dublin, Ireland}
%\affil{Affiliation}
\date{}
\maketitle
\thispagestyle{empty}

\begin{abstract}
%Understanding of visual attention or saliency is important for the design of visual computing systems. 
The effective design of visual computing systems depends heavily on the anticipation of visual attention, or saliency. While visual attention is well investigated for conventional 2D images and video, it is nevertheless a very active research area for emerging immersive media. In particular, visual attention of light fields (light rays of a scene captured by a grid of cameras or micro lenses) has only recently become a focus of research. As they may be rendered and consumed in various ways, a primary challenge that arises is the definition of what visual perception of light field content should be. In this work, we present a visual attention study on light field content. We conducted perception experiments displaying them to users in various ways and collected corresponding visual attention data. Our analysis highlights characteristics of user behaviour in light field imaging applications. The light field data set and attention data are provided with this paper.%to support further research.
\end{abstract}

%%%%%%%%%%%%%%%%%%%%%%
\section{Introduction}
\label{sec:intro}

New developments in capture~\cite{broxton2019low} and display technologies~\cite{lee2016additive, overbeck2018welcome} introduced a novel way of visual media representation - the light field.  In contrast to traditional imaging systems, which capture a 3D scene by projecting it onto a 2D surface, light fields~\cite{Levoy1996, Gortler1996} encode all angular, directional and intensity information of light rays travelling within a 3D-space. Light fields can be displayed in 3D on 360-degree displays~\cite{jones2007rendering} or specialised light field displays~\cite{lanman2011polarization}. They can also be viewed in 2D, as sheared perspective views or as focal stacks - where images with differing focal planes of the light field appear sharp or ``\textit{in focus}". Focal stacks are computed using digital refocusing \cite{ng2005light, lePendu2019fourier}.
%New developments in capture~\cite{broxton2019low} and display technologies~\cite{lee2016additive, overbeck2018welcome} introduced a novel way of visual media representation - the light field.  In contrast to traditional imaging systems, which capture a 3D scene by projecting it onto a 2D surface, light fields~\cite{Levoy1996, Gortler1996} encode all angular, directional and intensity information of light rays travelling within a 3D-space are captured. Light fields can be displayed in 3D on 360-degree displays~\cite{jones2007rendering} or specialised light field displays~\cite{lanman2011polarization}. They can also be viewed in 2D, as sheared perspective views or as focal stacks - where images with differing focal planes of the light field appear sharp or ``\textit{in focus}". Focal stacks are computed using digital refocusing \cite{ng2005light, lePendu2019fourier}.

Light fields hold more information than a regular image and can be used in various applications~\cite{matysiak2020high} including refocusing~\cite{lePendu2019fourier} and streaming~\cite{lfstreaming}. It's plausible that visual attention (where people look when they view a scene) varies according to media type. As a type distinct from 2D-image captures of scenes, light fields' relationship with visual attention may differ from conventional images. To our knowledge, saliency of varied renderings of light fields has not been previously investigated.

Hence, in this study, we built a light field visual attention database by bringing light fields from different sources together and collecting eye tracking data for different rendering scenarios of them. Our goal was to obtain ground truth visual attention data for light fields and analyse if it differs from attention in 2D images not generated from light fields. Light field refocusing was our chosen method to render the light fields which was representative of their 3D nature, but for a 2D display. We subsequently examined how changes in focus affected participants' visual attention, treating focus as a cue characteristic of light fields. 

The rest of this paper is structured as follows. Section~\ref{sec:relWork} discusses related work. Section~\ref{sec:database} introduces the selected light fields and rendering scenarios considered in this study. Details related to the collection of eye-tracking data and user study are given in Section~\ref{sec:userStudy}. Section~\ref{sec:results} provides an analysis and presents the results. We outline our conclusions in Section~\ref{sec:conclusion}.

\section{Related Work}
\label{sec:relWork}
Visual saliency is the subjective term describing perceived pertinent regions or elements of a scene which stand out in the scene context. It has been studied for images~\cite{Judd_2009} and videos~\cite{itti2005quantifying} on planar surfaces. Humans have been found to fixate on regions with greater edge density and local contrast~\cite{reinagel1999natural}. Low level features such as intensity, orientation and color contrast have been found to guide visual attention~\cite{itti1998model} as well as high level features like faces~\cite{buswell1935people, yarbus2013eye}. Viewers in visual attention experiments tend to move their gaze to targets near where they are currently looking. There is also a tendency to look at centrally located targets in their field of view.~\cite{parkhurst2002modeling}.

Previous work in light field saliency has focused on object-based methods of visual saliency prediction~\cite{li2016saliency, zhang2015saliency, sheng2016relative, zhang2017saliency, wang2019deep, Zhang2020}. These works represent saliency ground truth as binary maps. These maps are obtained by manually segmenting objects that stand out in all-in-focus renderings of the light fields and human-labelling those segments as 1 and all other regions as 0. They focus on the localisation of instances of dominant objects, not taking into account tracked human gaze. Only one type of light field rendering is considered, effectively ignoring the 3D nature of light fields.

Eye fixation data collected using eye-tracking devices provides a more meaningful form of ground truth for visual attention compared to binary maps, since it represents the statistical distribution of fixation data. Saliency maps~\cite{koch1987shifts}, which are continuous density maps that represent the probability of fixation at every point in an image, can be computed from this data and can be analysed to make more accurate models for predicting visual attention in light fields. Our work addresses the visual saliency of all spatial locations that attract visual attention, be they regions, objects or points of interest, which is in contrast to previous work. We investigated how attention is affected by changes in focus. Our results show that characteristics specific to light fields influence visual attention, as refocusing is a distinctive feature of light field technology.

%====================
%      DATABASE
%====================
\section{Database}
\label{sec:database}

The data was selected from four main light field datasets: Stanford (New) Light Field Archive~\cite{StanfordLF}, EPFL Light Field Image Dataset~\cite{Rerabek:218363}, Disney High Spatio-Angular Resolution Light Fields \cite{kim2013scene} and HCI Heidelberg 4D Light Field Dataset~\cite{honauer2016dataset}. 
We believe that the selection of these four datasets makes the collected data representative as the light fields were acquired using a camera array~\cite{StanfordLF}, a single camera with microlens array~\cite{Rerabek:218363}, a camera on a gantry~\cite{kim2013scene}, and computer generated imagery~\cite{honauer2016dataset} respectively.

We selected 20 light fields from these datasets according to the following criteria: they contained multiple regions or objects with high colour contrast between each other and contained regions with great edge density and local contrast at varied depths and spatial locations. 

Slices of a light field focused at a sequence of depths form what is known as a focal stack. We generated these focal stacks for each of our light fields using the Fourier Disparity Layers method \cite{lePendu2019fourier} and used them to simulate traversing a 3D scene on a 2D display. We considered three different scenarios for light field rendering and rendered each light field in five ways as follows:
 \begin{enumerate}
     \item \textbf{all-in-focus:} all the points in the rendered image are in focus.
     \item \textbf{region-in-focus:} one slice/image of the focal stack is rendered so only objects at that slice's specific depth of focus appear sharp. We rendered two regions \textit{region\nobreakdash-1} and \textit{region\nobreakdash-2} which have objects in opposite positions of the frame eg. left/right, top/bottom, foreground/background.
     \item \textbf{focal-sweep:} all the images of the focal stack are rendered in sequence. We rendered two focal sweeps \textit{front-to-back} with region of focus moving from foreground to background and \textit{back-to-front} with region of focus moving from background to foreground.
 \end{enumerate}

This created database is made publicly available on our project webpage%\footnote{The link is removed to preserve anonymity. It will be shared here upon acceptance.}, 
\footnote{\url{https://v-sense.scss.tcd.ie/research/light-fields/visual-attention-for-light-fields/}}, 
in order to support further scientific studies in this field. On this webpage, we also share some additional results which we could not add to this paper due to page limitations.

%====================
%   SUBJECTIVE TEST
%====================
\section{Eye Tracking Data Collection}
\label{sec:userStudy}

%In this section, we describe the details of the data collection of the eye tracking data for the light field database we detailed in Section~\ref{sec:database}. 

\subsection{Apparatus \& Setup}
\label{subsec:setup}

We used a desktop mounted eye-tracker, the Eyelink 1000 plus~\cite{Eyelink2016} which records eye movements with a sampling rate of 1000Hz. The visual stimuli were presented on a 23.8 inch Dell P2415Q monitor (height $\times$ width: 29.6 $\times$ 52.7 cm; native resolution: 4K/UHD/2160p; refresh rate: 60Hz). The monitor was placed at 67cm from the users eye which kept the visual angle of the stimuli between 39$^{\circ}$ and 24$^{\circ}$. The resolution of the monitor was set to be 1920 $\times$ 1080 pixels (16:9 aspect ratio).

The experiment was held in a quiet, well lit room with white walls. We used the standard Eyelink 1000 chin rest to minimise head movement. We specified the width and height of the monitor as well as the resolution and our measured eye to screen distance in the eye-tracker configuration files.
The test script was written in Matlab (R2019b) using the EyeLink Toolbox within Psychtoolbox~3~\cite{kleiner2007s}. %Psychophysics Toolbox~3~\cite{kleiner2007s}. 

\subsection{Participants \& Methodology}
\label{subsec:methodology}
We conducted the experiment, following ethics approval, on 21 participants (16 male and 5 female), aged between 18 and 37 with a mean age of 25.3. A department wide email was sent to students and staff for recruitment.
All participants had normal vision or corrected-to-normal vision. The experiment lasted between 25 and 35 minutes for each participant. A brief oral overview of the experiment as well as an information sheet and consent form were provided to participants. They were instructed to view the stimuli freely and naturally while keeping their heads as still as possible. The chin rest position was fixed but the participants could raise or lower their chair until they were comfortable. The distance from the eye to eyetracker was kept at 53cm.

Eye movements of the left eye only were recorded. We used the Eyelink default monocular nine-point calibration and validation procedure. We showed the participants the light fields rendered all-in-focus for 4 seconds each to acquaint them with the data. They were then shown the five renderings of each light field, for 10 seconds each (120 frames with 12 fps), in randomised order, with a 2 seconds interval between. The interval screen was to ensure fixation was re-centered. Randomisation was used to avoid carryover~\cite{greenwald1976within}.

The eye-tracker records eye events, \textit{saccades}, \textit{fixations} and blinks.
%as well as other relevant data such as blinks. 
Fixations are periods in which an area of a visual scene is kept on the fovea. Saccades are rapid movements of the eyes whose function is to change the point of fixation by directing the fovea towards an area of visual interest~\cite{yarbus2013eye}. In the subsequent analysis section, we use fixations recorded by the EyeLink Core System in an EyeLink data file (EDF). %EDF (EyeLink data file)}.%\noteCO{you nicely explain it, but also mention that we filter sacades and others..}

%====================
%       Results
%====================
\section{Results}
\label{sec:results}
%Having collected eye tracking data, in this section, we will explain the analysis we conducted and our results. 

\subsection{Qualitative Analysis}
% Below is an example of how to insert images. Delete the ``\vspace'' line,
% uncomment the preceding line ``\centerline...'' and replace ``imageX.ps''
% with a suitable PostScript file name.
% -------------------------------------------------------------------------

\begin{figure*}[t]%[htb]
%%%%%%%%%%%%%%%%%%%%%%%%%%%%%%%%%%%%%%
\begin{comment}
\begin{minipage}[b]{0.015\linewidth}
    \rotatebox[origin = c]{90}{LegoKnights}
\end{minipage}
  \begin{subfigure}{0.155\linewidth}
        \centering
        \includegraphics[width=\linewidth]{center_SAI_lego.jpg}
  \end{subfigure}\hfill
  \begin{subfigure}{0.155\linewidth}
        \centering
        \includegraphics[width=\linewidth]{figures_LegoKnights_allInFocus.jpg}
  \end{subfigure}\hfill
  \begin{subfigure}{0.155\linewidth}
        \centering
        \includegraphics[width=\linewidth]{figures_LegoKnights_focused-obj1.jpg}
  \end{subfigure}\hfill
  \begin{subfigure}{0.155\linewidth}
        \centering
        \includegraphics[width=\linewidth]{figures_LegoKnights_focused-obj2.jpg}
  \end{subfigure}\hfill
  \begin{subfigure}{0.155\linewidth}
        \centering
        \includegraphics[width=\linewidth]{figures_LegoKnights_front2back.jpg}
  \end{subfigure}\hfill
  \begin{subfigure}{0.155\linewidth}
        \centering
        \includegraphics[width=\linewidth]{figures_LegoKnights_back2front.jpg}
  \end{subfigure}%\hfill
  \vspace{0.1cm}
\end{comment}
%%%%%%%%%%%%%%%%%%%%%%%%%%%%%%%%%%%%%%%
%=============================================================================
%\rotatebox[origin = c]{90}{\hspace*{3 mm}Treasure}
  %\hspace*{2.5 mm}
  %\hspace*{30.6 mm}
  \begin{subfigure}{0.155\linewidth}
        \centering
        \includegraphics[width=\linewidth]{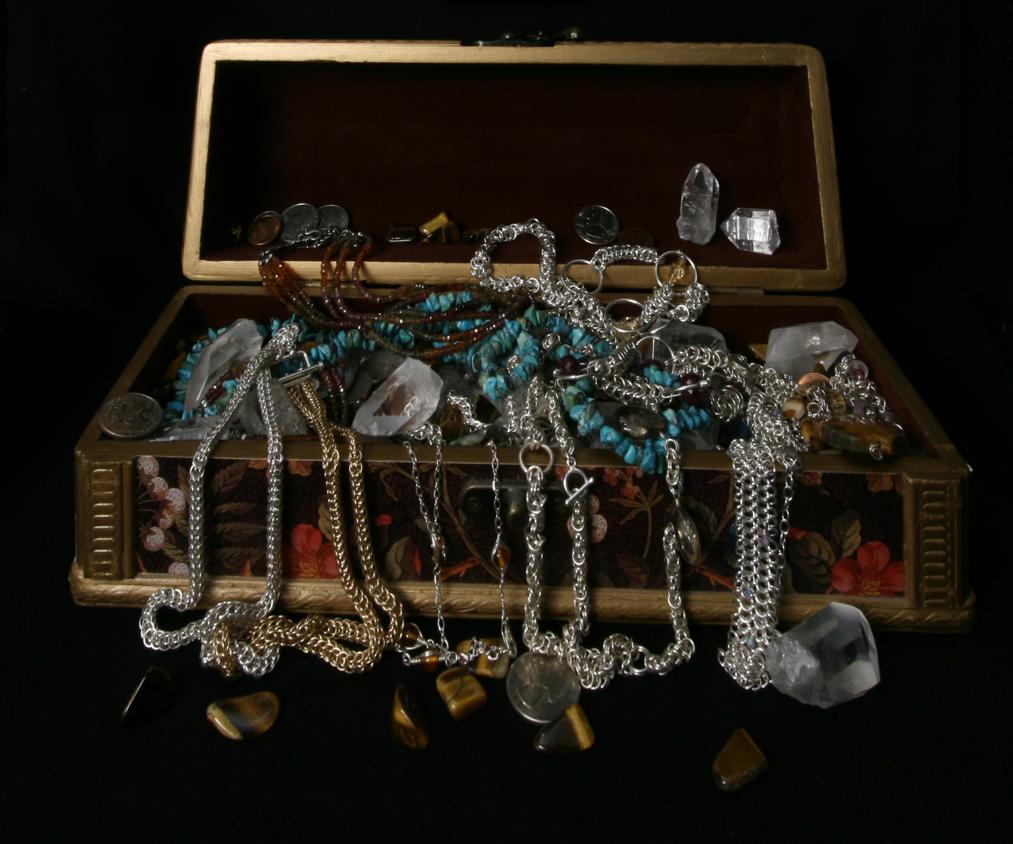}
  \end{subfigure}\hfill
  \begin{subfigure}{0.155\linewidth}
        \centering
        \includegraphics[width=\linewidth]{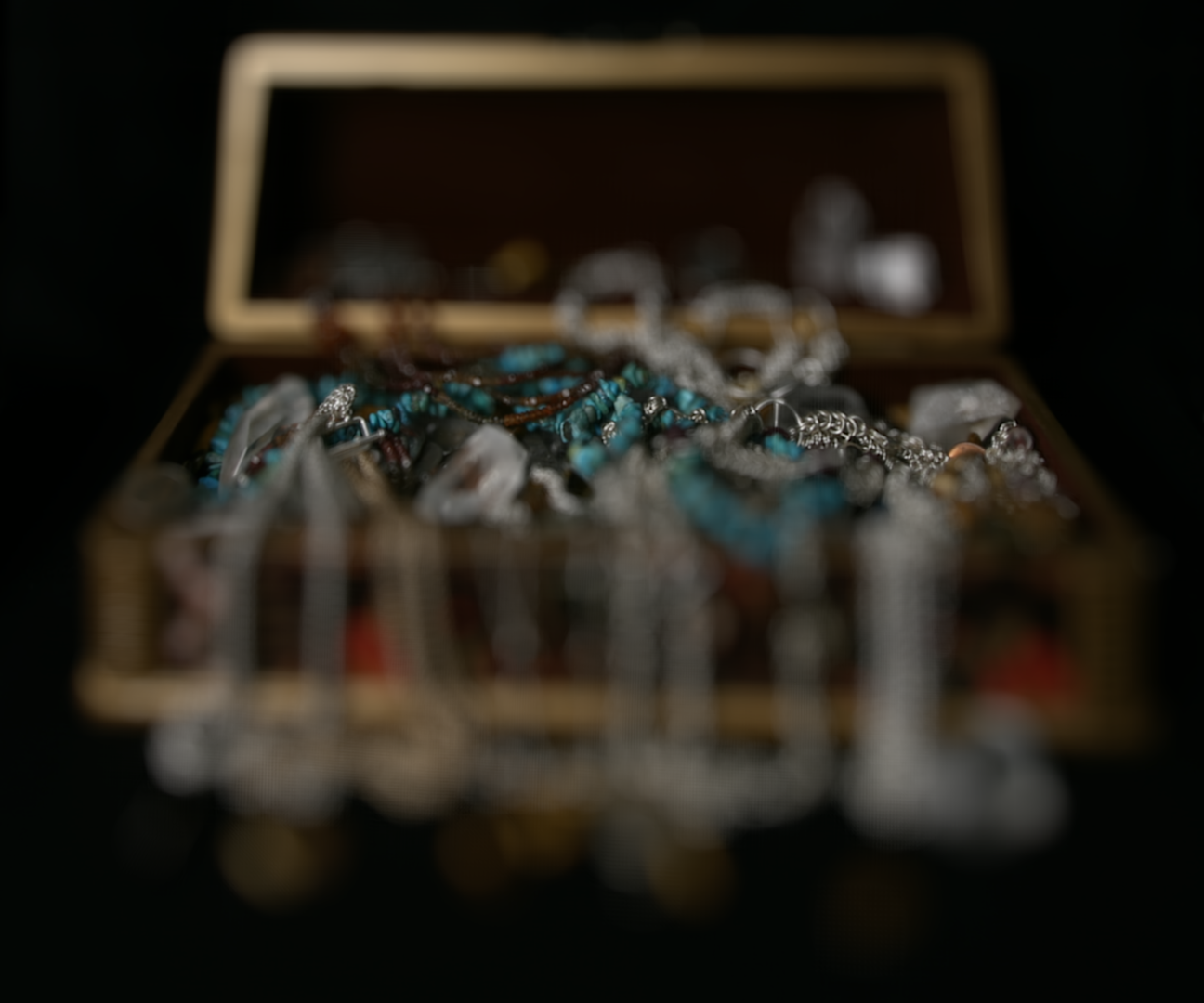}
  \end{subfigure}\hfill
  \begin{subfigure}{0.155\linewidth}
        \centering
        \includegraphics[width=\linewidth]{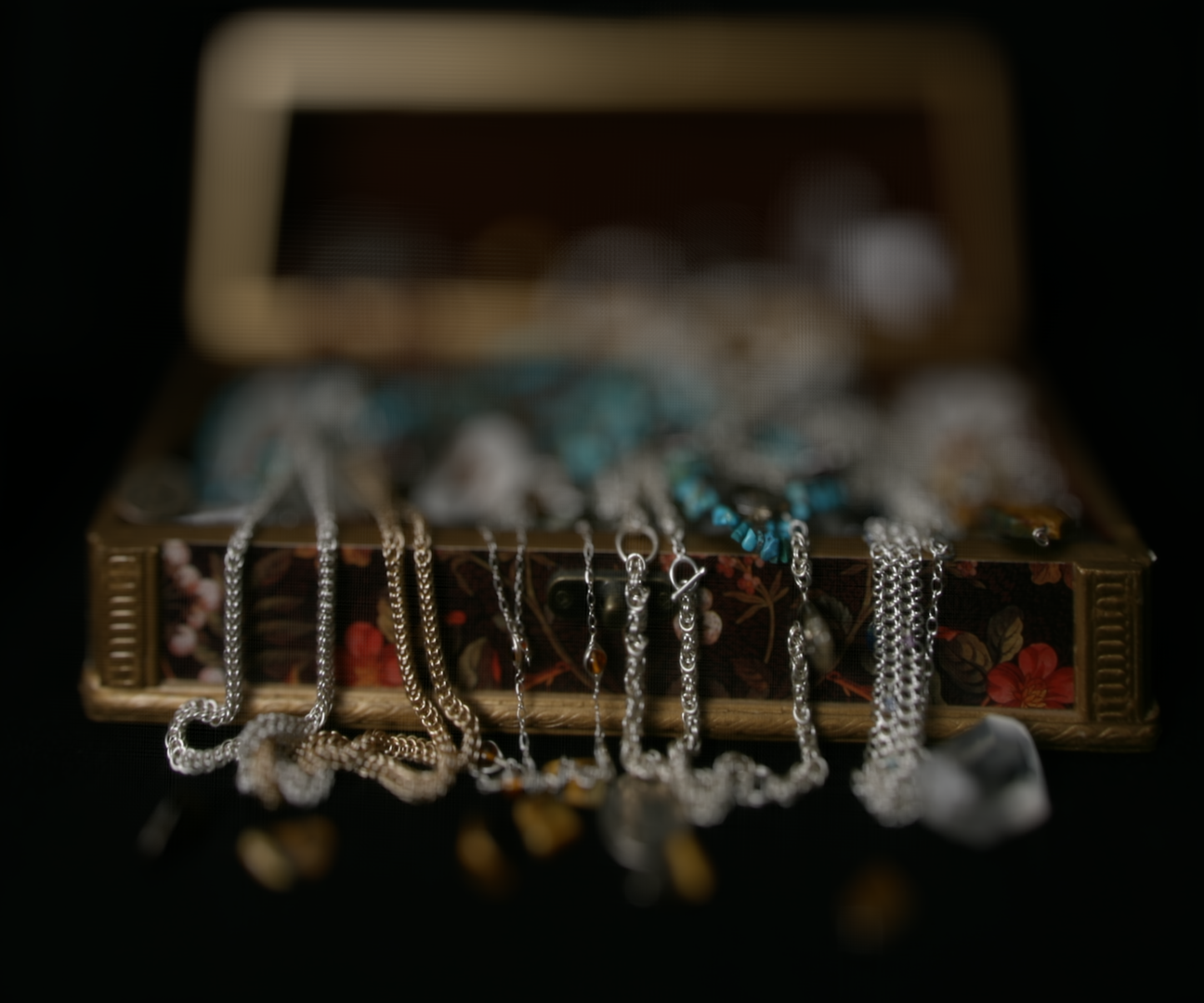}
  \end{subfigure}\hfill
  \begin{subfigure}{0.155\linewidth}
        \centering
        \includegraphics[width=\linewidth]{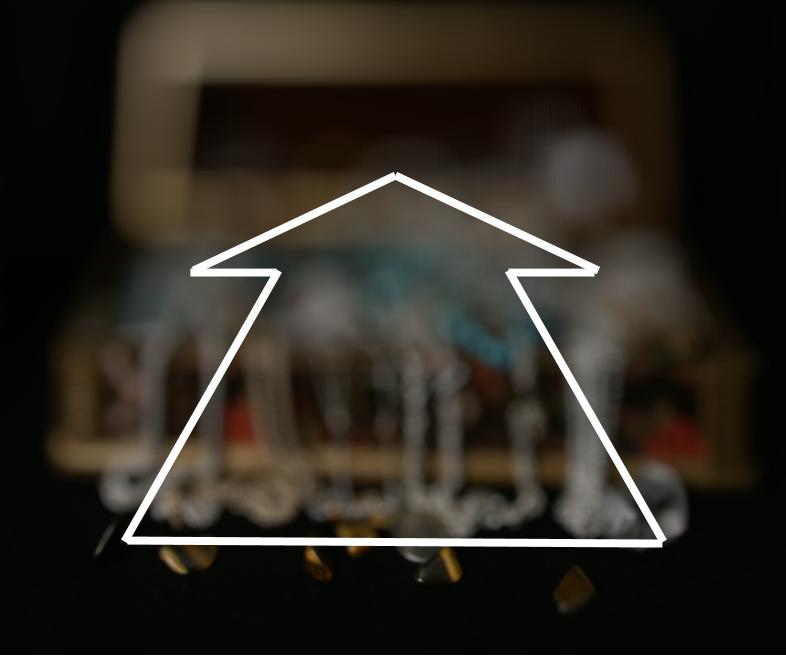}
  \end{subfigure}\hfill
  \begin{subfigure}{0.155\linewidth}
        \centering
        \includegraphics[width=\linewidth]{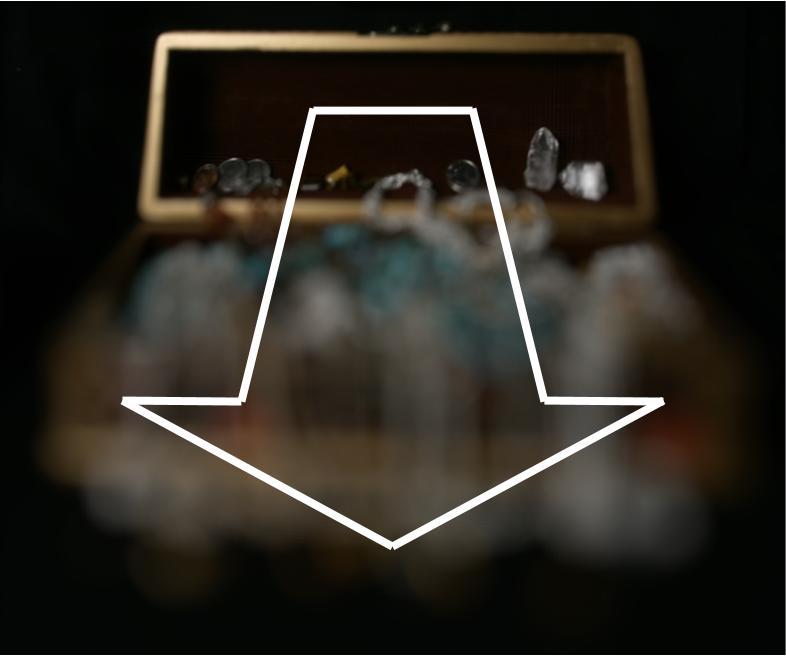}
  \end{subfigure}%\hfill
  \vspace{0.1cm}
  
%\begin{minipage}[b]{0.015\linewidth}
 %   \rotatebox[origin = c]{90}{\hspace*{4 mm}Treasure}
%\end{minipage}
  %\begin{subfigure}{0.155\linewidth}
        %\centering
        %\includegraphics[width=\linewidth]{center_SAI_treasure.jpg}
        %\caption*{(a) Reference}
        %\includegraphics[width=\linewidth]{TreasureChest_focalStack_map.png}
        %\caption*{(a) Depth Map}
 % \end{subfigure}\hfill
  \begin{subfigure}{0.155\linewidth}
        \centering
        \includegraphics[width=\linewidth]{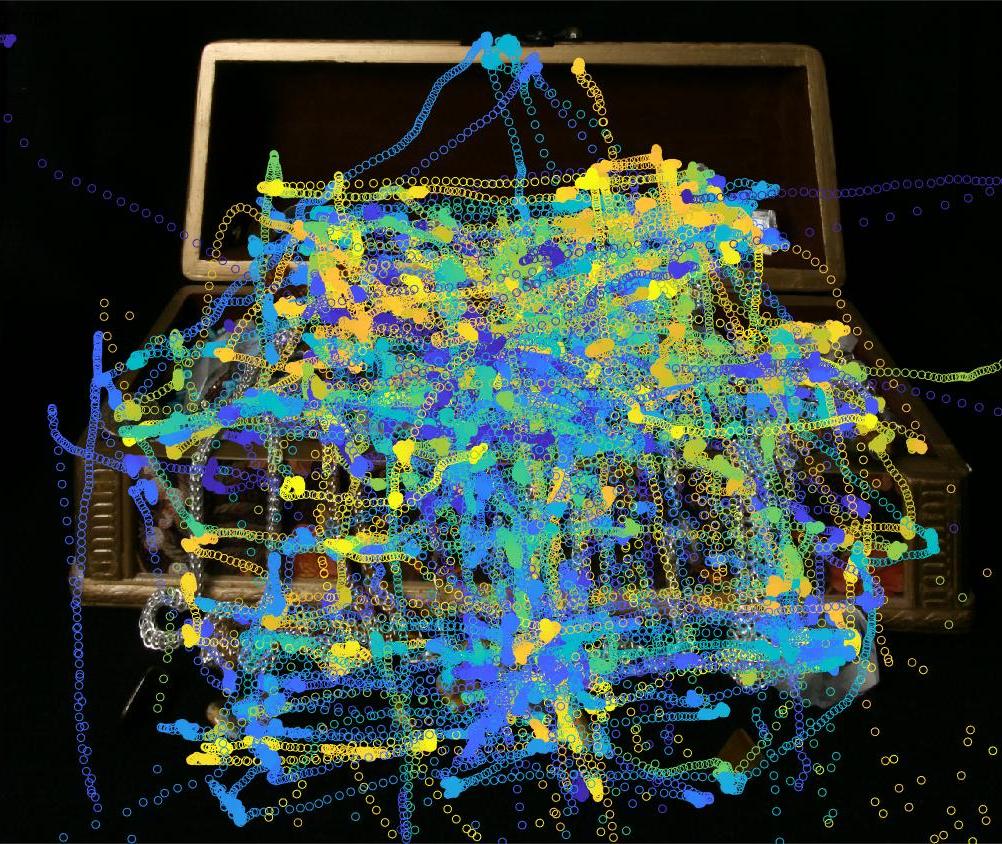}
        \caption*{(a) All-in-focus}
  \end{subfigure}\hfill
  \begin{subfigure}{0.155\linewidth}
        \centering
        \includegraphics[width=\linewidth]{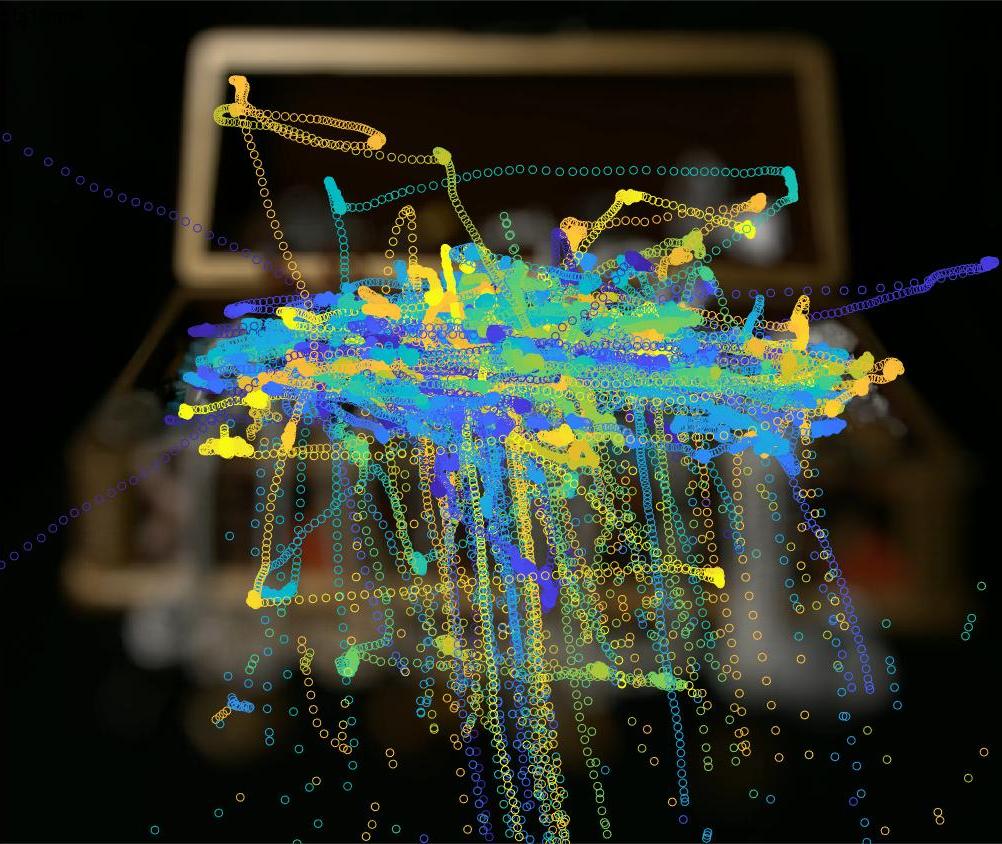}
        \caption*{(b) Region-1}
  \end{subfigure}\hfill
  \begin{subfigure}{0.155\linewidth}
        \centering
        \includegraphics[width=\linewidth]{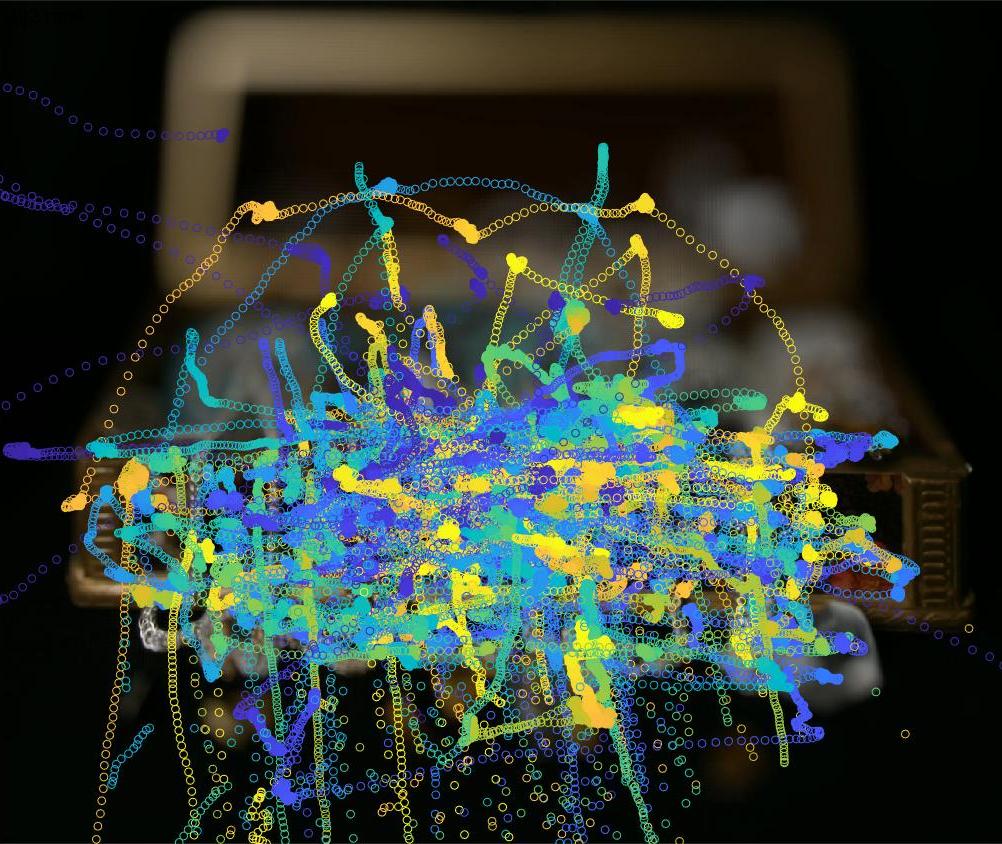}
        \caption*{(c) Region-2}
  \end{subfigure}\hfill
  \begin{subfigure}{0.155\linewidth}
        \centering
        \includegraphics[width=\linewidth]{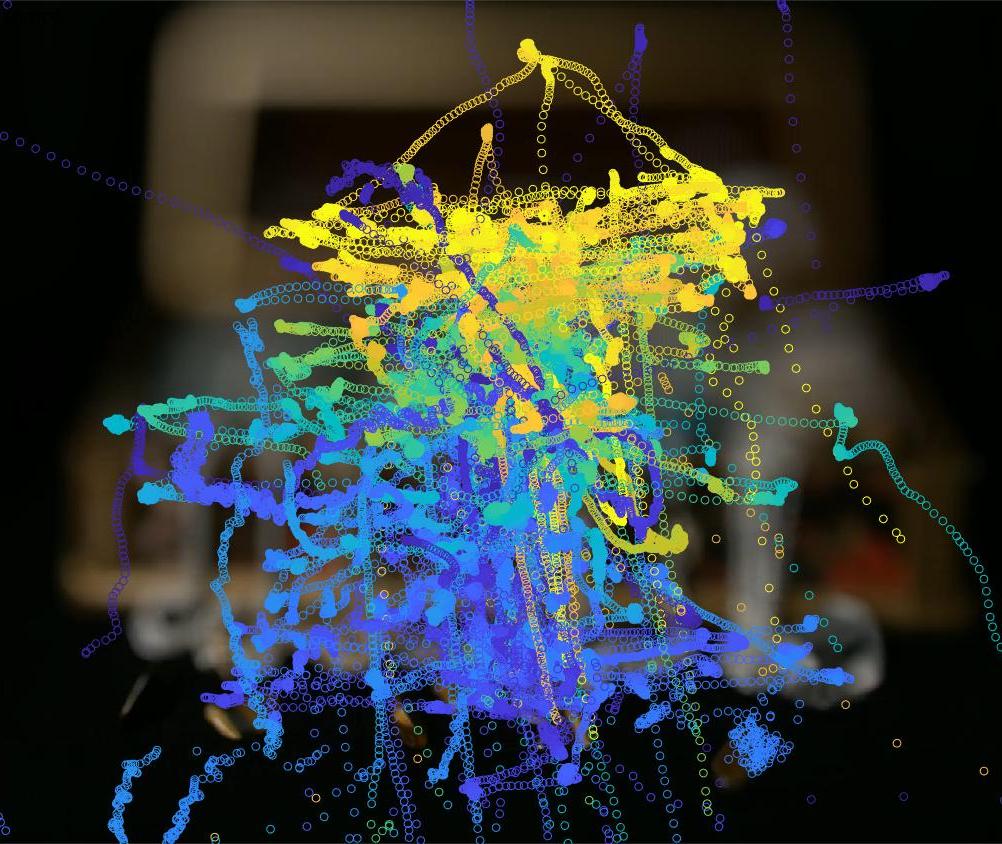}
        \caption*{(d) Front-to-back}
  \end{subfigure}\hfill
  \begin{subfigure}{0.155\linewidth}
        \centering
        \includegraphics[width=\linewidth]{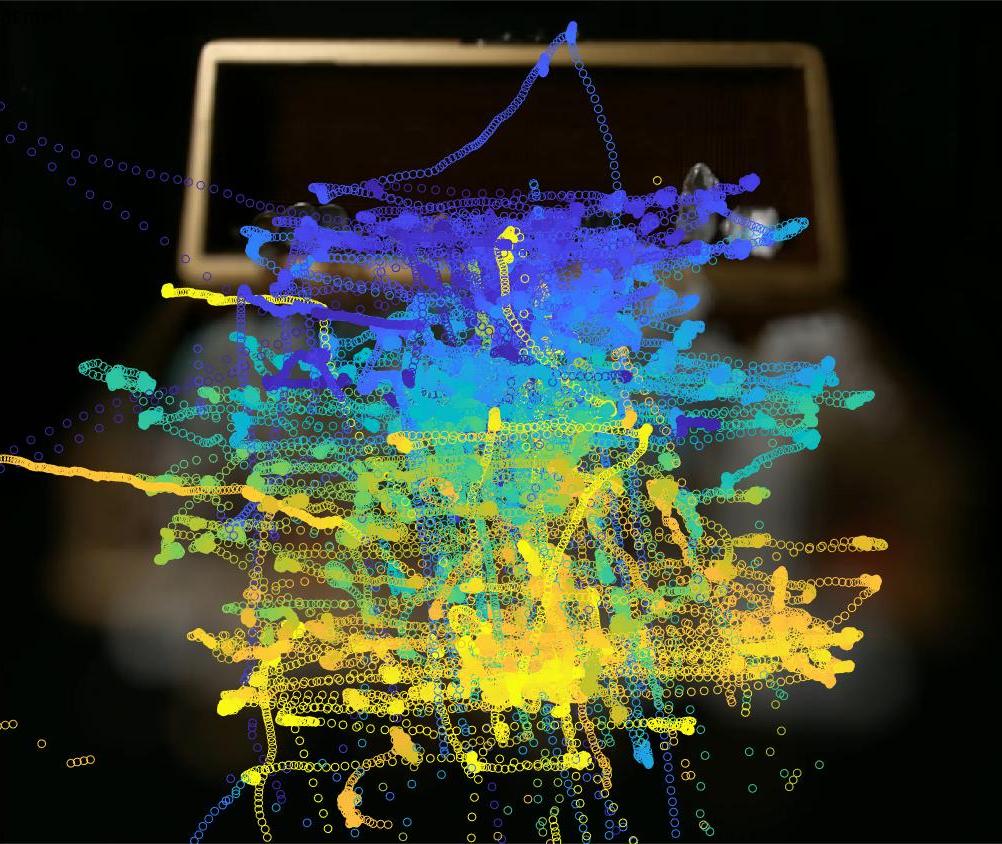}
        \caption*{(e) Back-to-front}
  \end{subfigure}\hfill

%\vspace{-0.4cm}
\vspace{-0.1cm}
\caption{Scanpaths of Treasure light field renderings. Each continuous chain represents a participant, and colour mapping shows the passage of time which starts with blue. Yellow is the most recent time instant.}% and ends with yellow. That is, yellow is the most recent time instant.} % Subfigures (b-f) and (h-l) shows the scanpaths in five different rendering cases for (a) LegoKnights and (g) Treasure light fields, respectively.}
\label{fig:qualAnalysis}
  \vspace{-0.1cm}
\end{figure*}

Fig.~\ref{fig:qualAnalysis} shows the scanpaths of the raw eye tracking data for each rendering of two sample light fields. %the LegoKnights and Treasure light fields. .
Clusters where the colours of the scanpaths are the same reveal a clear path of fixation. This can be observed in the scanpaths of the videos where the focal plane varies over time, as shown in the front-to-back and back-to-front renderings in Fig.~\ref{fig:qualAnalysis}. This is in contrast to the scanpaths of the data collected with non-varying focal planes i.e, all-in-focus, region-1 and region-2, which suggest that each participant views regions of the scene in a different order. This phenomenon is also seen in images rendered from the other light fields in this data set.

We generated saliency maps from the fixation points recorded by our eye-tracker to further analyse the visual saliency patterns in our data. We applied a Gaussian filter to our fixation data to obtain these maps. As our largest image width was 47.11cm and height was 29.60cm, we calculated the visual angle to be 24.91$^\circ$ to 38.74$^\circ$ respectively. We then found that 1$^\circ$ visual angle corresponds to 47.66 pixels horizontally and 42.67 pixels vertically and used these values as our standard deviations $\sigma_x$ and $\sigma_y$~\cite{le2013methods}. We used the duration of the fixations as a weight when computing the Gaussian.

We created saliency maps for each light field and corresponding rendering in two ways. The first method involved computing maps using the fixations of all participants for the full 10 second video. The second split each video over time into 5 segments (of 2 seconds each). For each of these, we generated a saliency map per segment using the fixations of all participants. This allowed us to see changes in visual attention over time.

\begin{figure}[htb]
%%%%%%%%%%%%%%%%%%%%%%%%%%%%%%5
\begin{comment}
  \rotatebox[origin = c]{90}{\hspace*{3 mm}Dino}
  \begin{subfigure}{0.16\linewidth}
    \centering
    \includegraphics[width=\linewidth]{center_SAI_dino.jpg}
    \label{fig:1}
  \end{subfigure}\hfill
  \begin{subfigure}{0.16\linewidth}
    \centering
    \includegraphics[width=\linewidth]{dino_allInFocus.jpg}
    \label{fig:2}
  \end{subfigure}\hfill
  \begin{subfigure}{0.16\linewidth}
    \centering
    \includegraphics[width=\linewidth]{dino_front2back.jpg}
    \label{fig:3}
  \end{subfigure}\hfill
  \begin{subfigure}{0.16\linewidth}
    \centering
    \includegraphics[width=\linewidth]{dino_back2front.jpg}
    \label{fig:4}
  \end{subfigure}\hfill
  \begin{subfigure}{0.16\linewidth}
    \centering
    \includegraphics[width=\linewidth]{dino_focused-obj1.jpg}
    \label{fig:5}
  \end{subfigure}\hfill
  \begin{subfigure}{0.16\linewidth}
    \centering
    \includegraphics[width=\linewidth]{dino_focused-obj2.jpg}
    \label{fig:6}
  \end{subfigure}\hfill
\end{comment}
%%%%%%%%%%%%%%%%%%%%%
  %\hspace*{30.5 mm}
  \begin{subfigure}{0.16\linewidth}
    \centering
    \includegraphics[width=\linewidth]{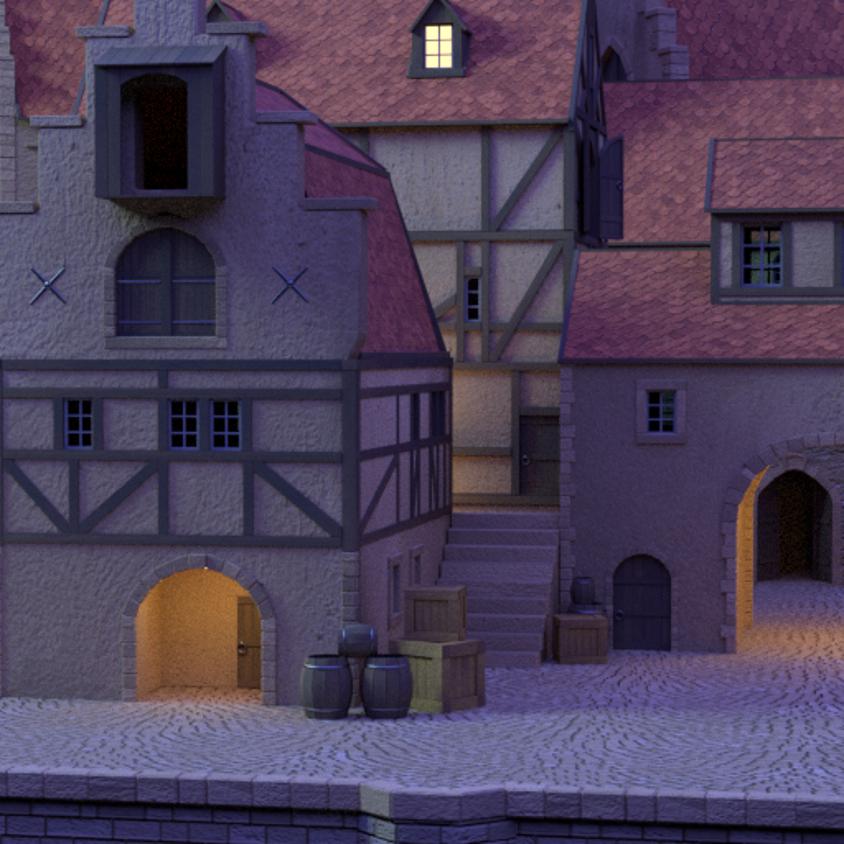}
    \label{fig:2}
  \end{subfigure}\hfill
  \begin{subfigure}{0.16\linewidth}
    \centering
    \includegraphics[width=\linewidth]{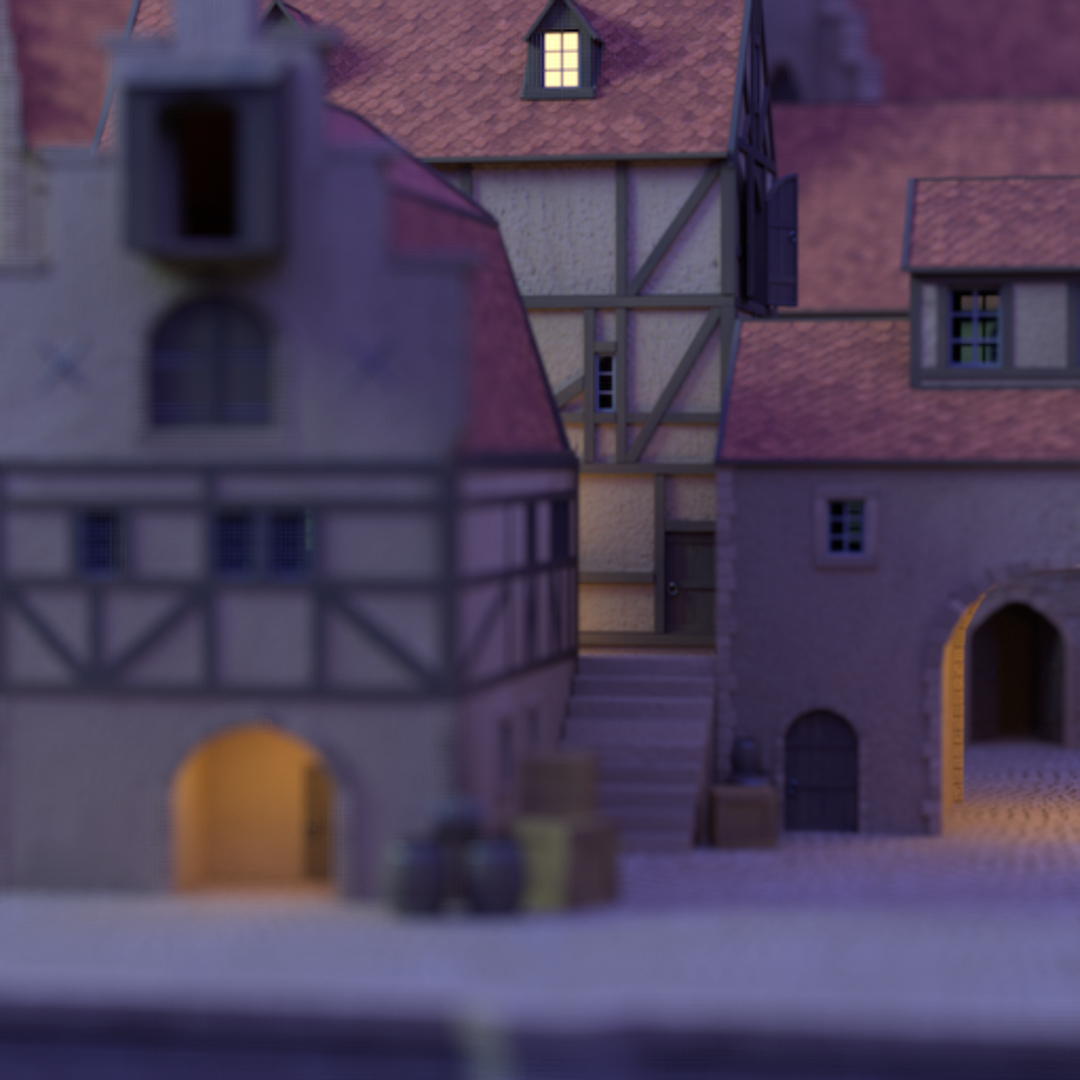}
    \label{fig:5}
  \end{subfigure}\hfill
  \begin{subfigure}{0.16\linewidth}
    \centering
    \includegraphics[width=\linewidth]{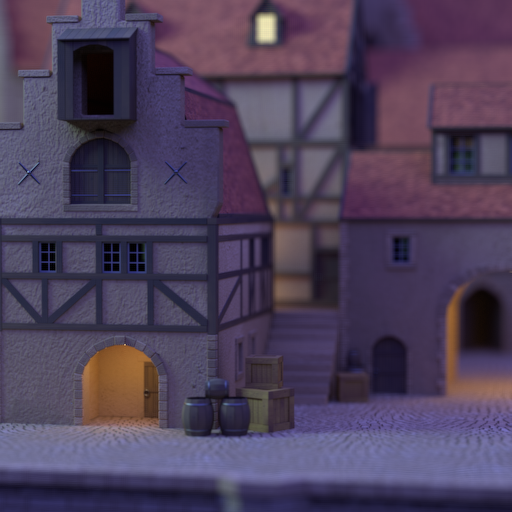}
    \label{fig:6}
  \end{subfigure}\hfill
    \begin{subfigure}{0.16\linewidth}
    \centering
    \includegraphics[width=\linewidth]{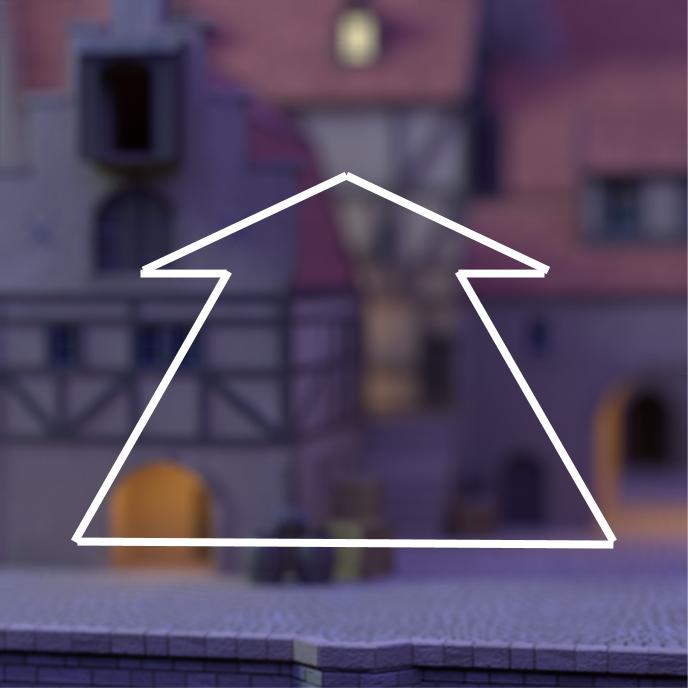}
    \label{fig:3}
  \end{subfigure}\hfill
  \begin{subfigure}{0.16\linewidth}
    \centering
    \includegraphics[width=\linewidth]{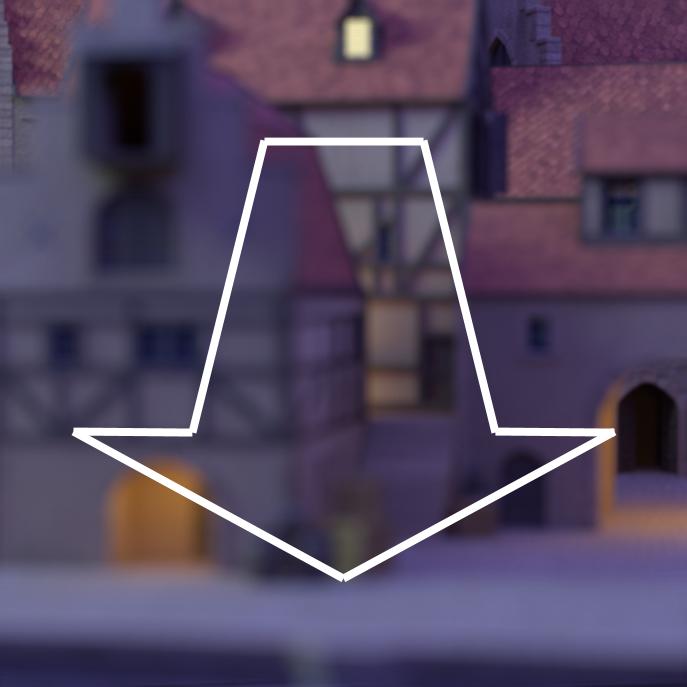}
    \label{fig:4}
  \end{subfigure}\hfill
  
  \vspace{-4mm}  
  %\rotatebox[origin = c]{90}{\hspace*{4 mm}Medieval}
  %\begin{subfigure}{0.16\linewidth}
    %\centering
    %\includegraphics[width=\linewidth]{center_SAI_medieval.jpg}
    %\includegraphics[width=\linewidth]{medieval2_focalStack_map.png}
    %\caption{Reference}
    %\caption{Depth Map}
   % \label{fig:7}
 % \end{subfigure}\hfill
  \begin{subfigure}{0.16\linewidth}
    \centering
    \includegraphics[width=\linewidth]{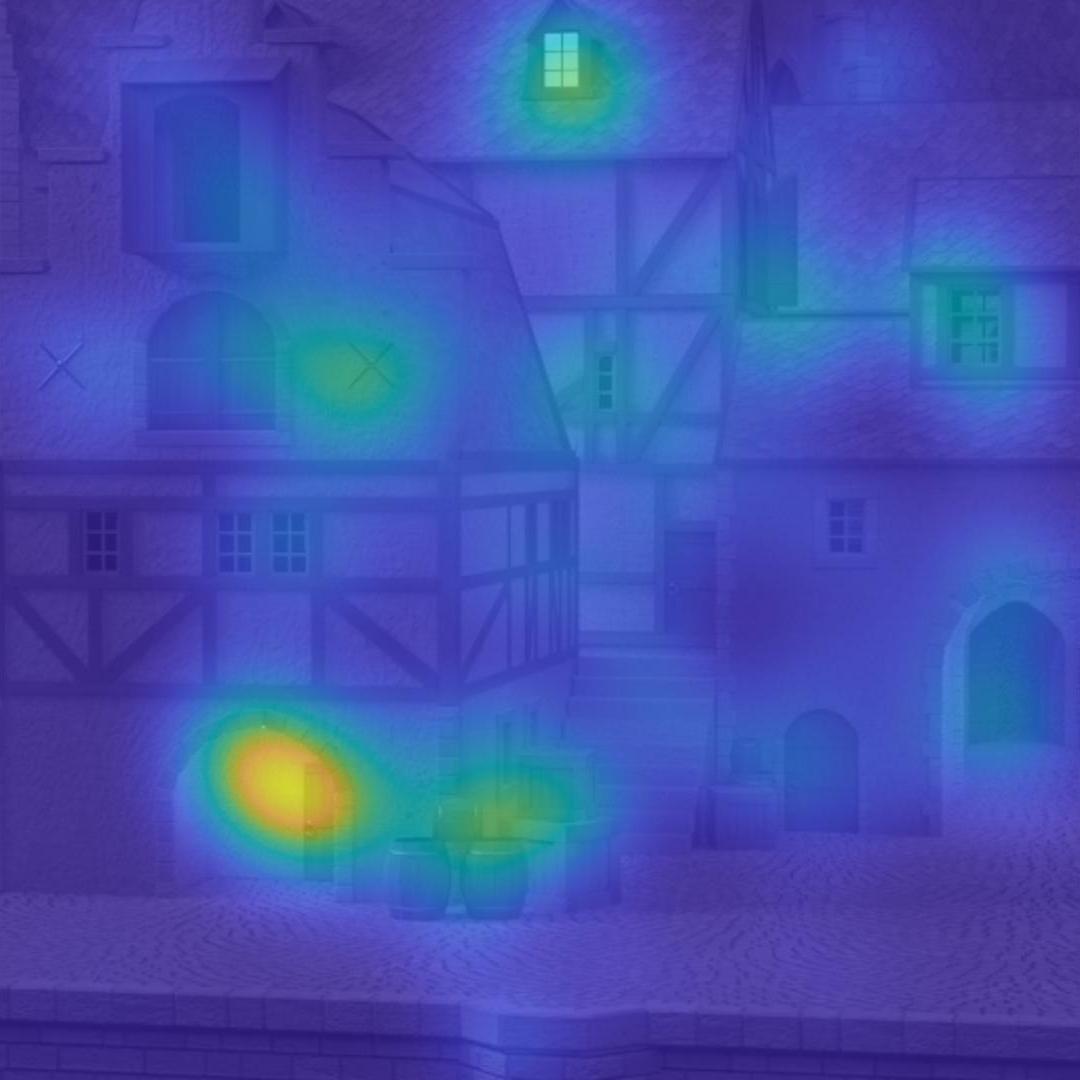}
    \caption{All-in-focus}
    \label{fig:8}
  \end{subfigure}\hfill
  \begin{subfigure}{0.16\linewidth}
    \centering
    \includegraphics[width=\linewidth]{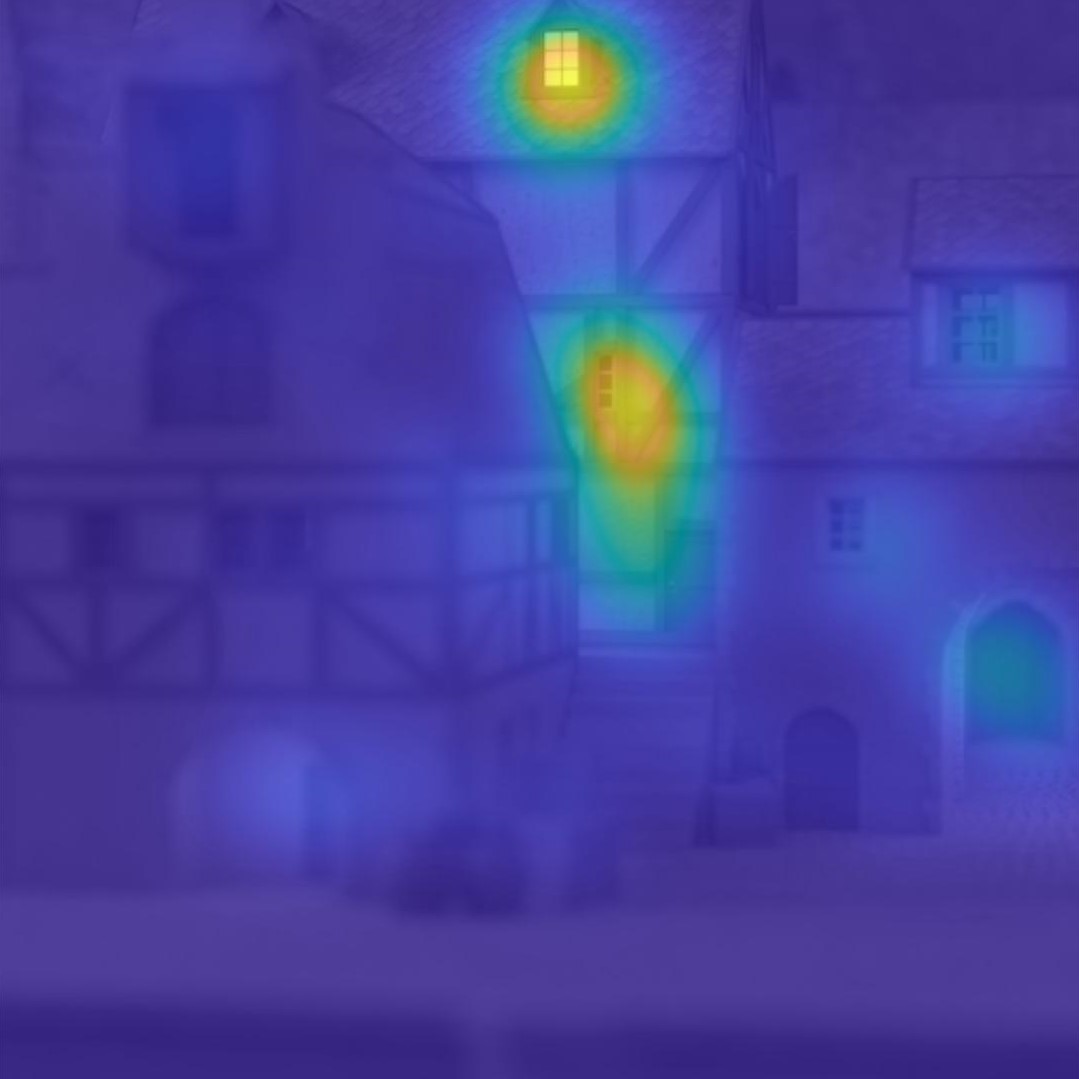}
    \caption{Region-1}
    \label{fig:11}
  \end{subfigure}\hfill
  \begin{subfigure}{0.16\linewidth}
    \centering
    \includegraphics[width=\linewidth]{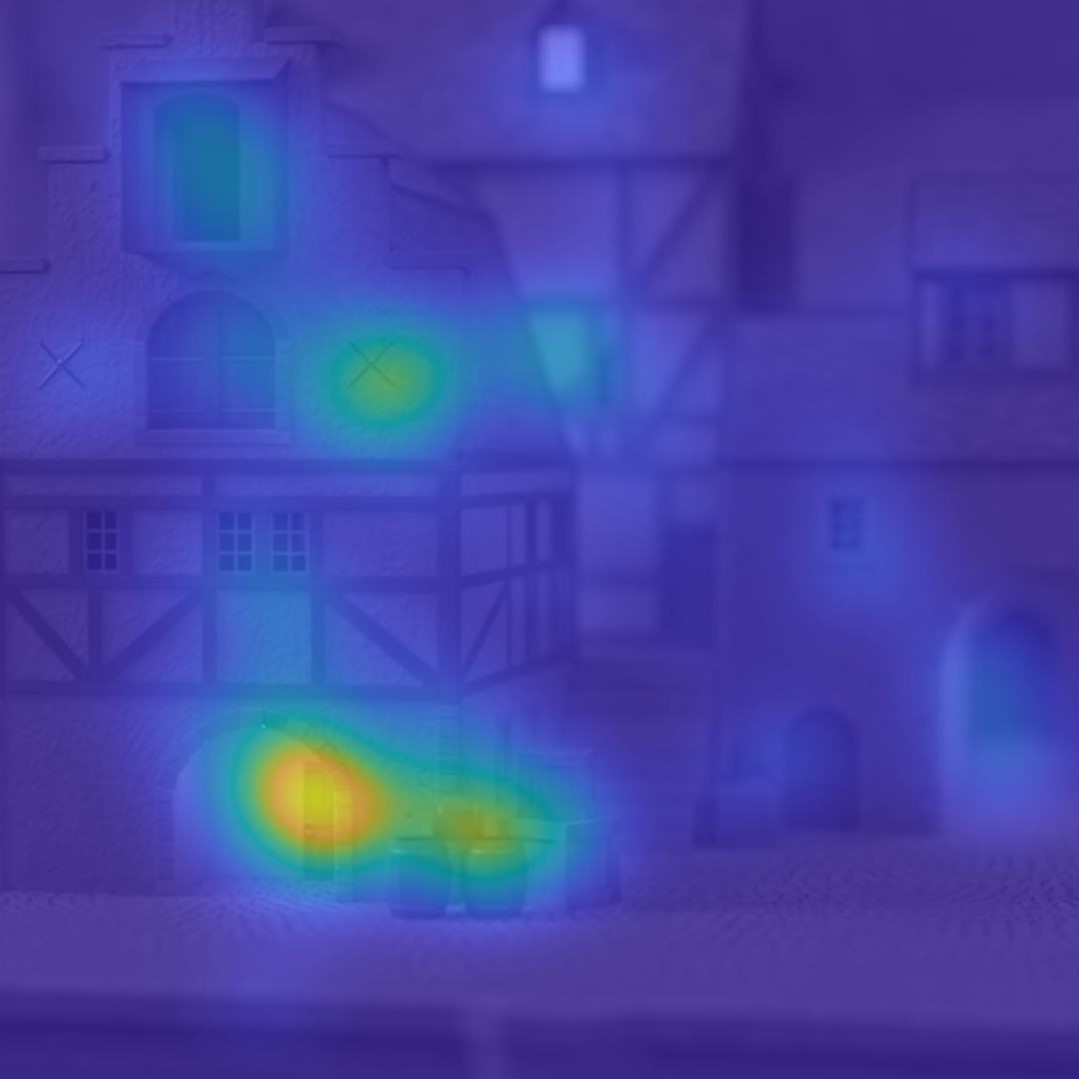}
    \caption{Region-2}
    \label{fig:12}
  \end{subfigure}\hfill
  \begin{subfigure}{0.16\linewidth}
    \centering
    \includegraphics[width=\linewidth]{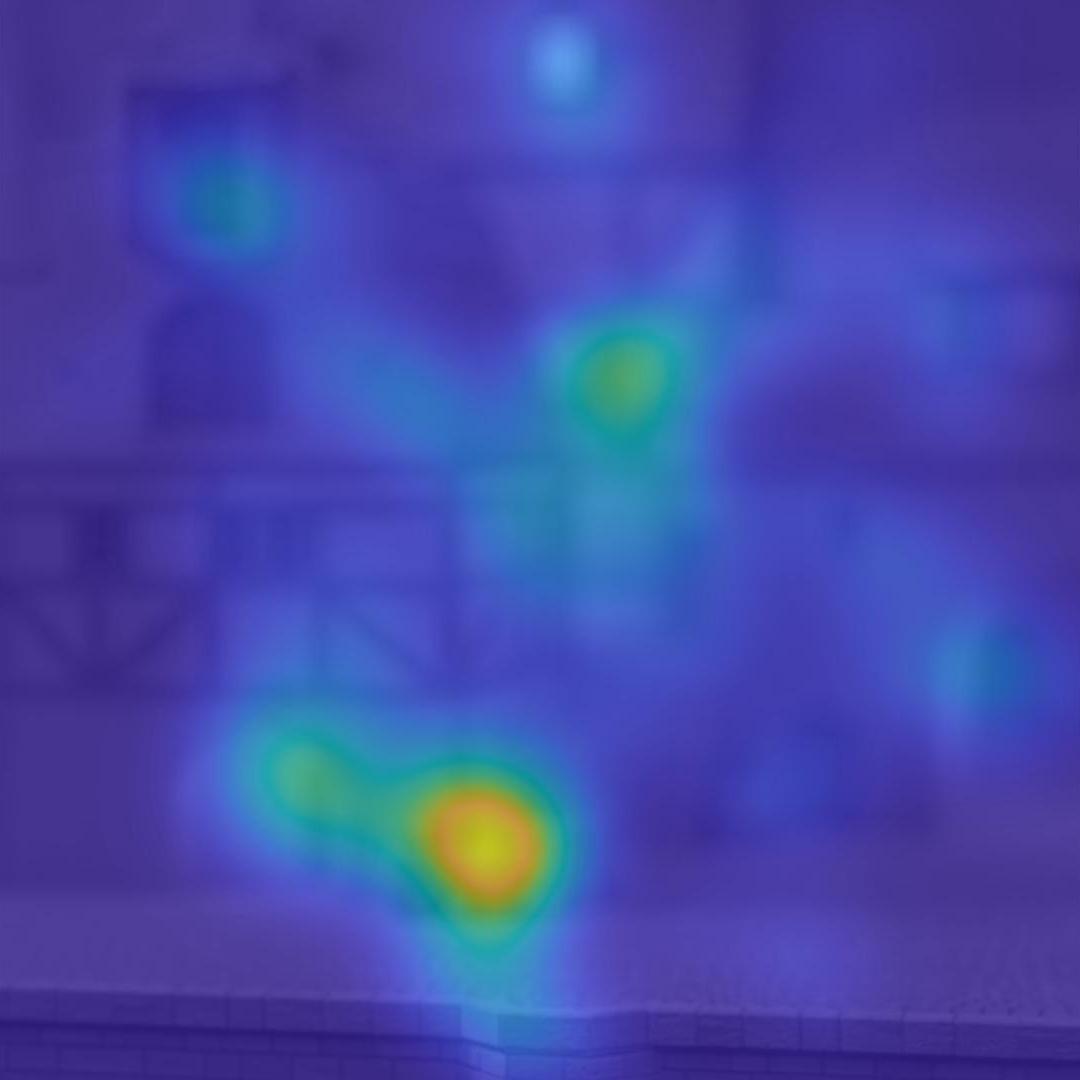}
    \caption{Front-to-back}
    \label{fig:9}
  \end{subfigure}\hfill
  \begin{subfigure}{0.16\linewidth}
    \centering
    \includegraphics[width=\linewidth]{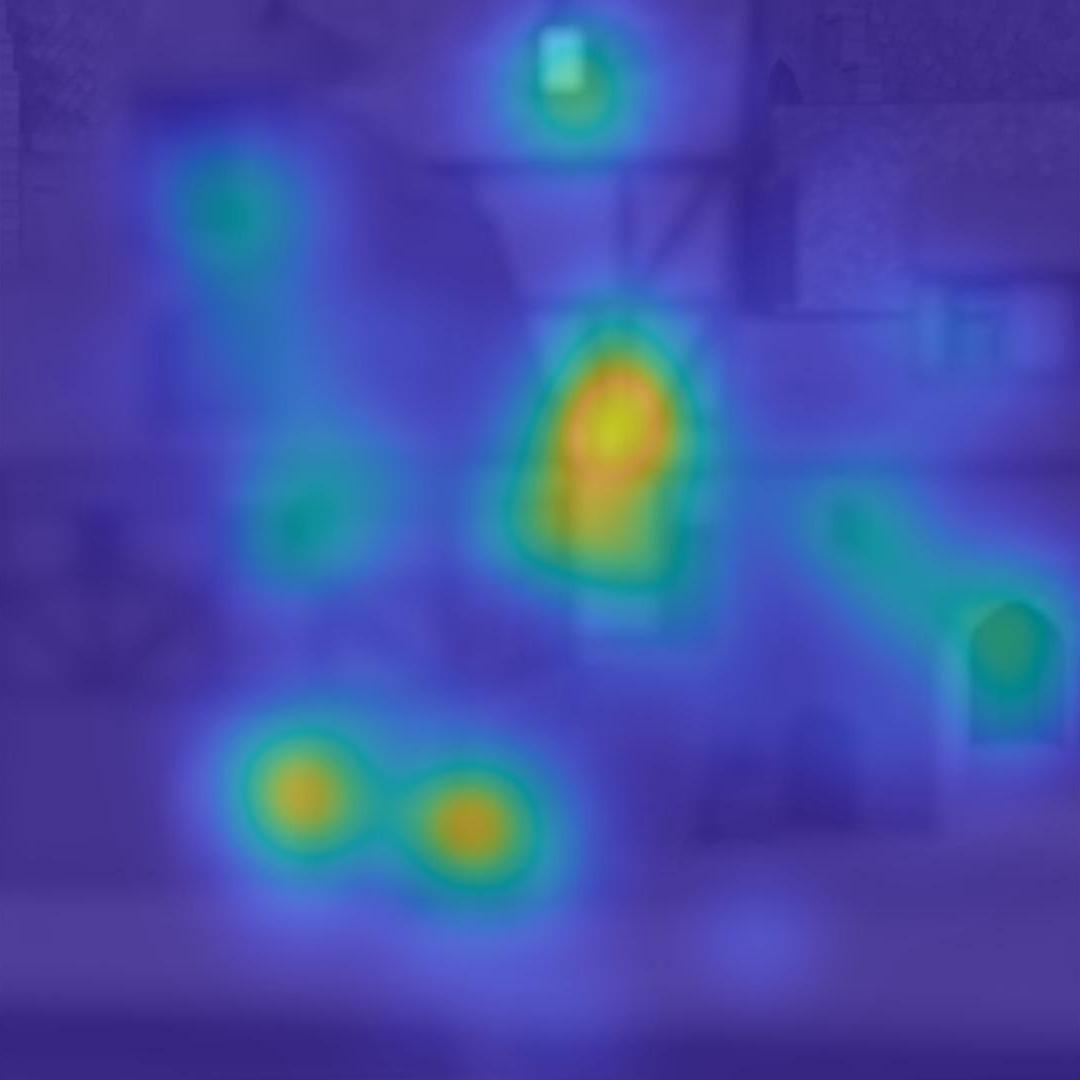}
    \caption{Back-to-front}
    \label{fig:10}
  \end{subfigure}\hfill
  \caption{Medieval light field rendered three ways all-in-focus, front-to-back and back-to-front overlayed with a heatmap, generated from all participants and averaged over entire video.}
  \vspace{-5pt}
  \label{fig:WholeVidExamples}
\end{figure}

We studied the heatmaps averaged over the entire 10-second video. We found that some had very similar saliency maps for the all-in-focus and focal-sweeps. %which is seen in the Dino light field in Fig.~\ref{fig:WholeVidExamples}. 
However, there were also many saliency maps where focal-sweeps revealed other salient regions. For example, in the Medieval light field in Fig.~\ref{fig:WholeVidExamples}, the centre building is salient in the back-to-front rendering whereas it is not in the all-in-focus rendering. This shows that a static rendering of a scene may not reveal all the salient regions present in the 3-dimensional light field data.

\begin{figure*}[tb]
\rotatebox[origin = c]{90}{(a)}
  \rotatebox[origin = c]{90}{Sideboard}
\rotatebox[origin = c]{90}{RiF1}
  \begin{subfigure}{0.1555\linewidth}
    \centering
    \includegraphics[width=\linewidth]{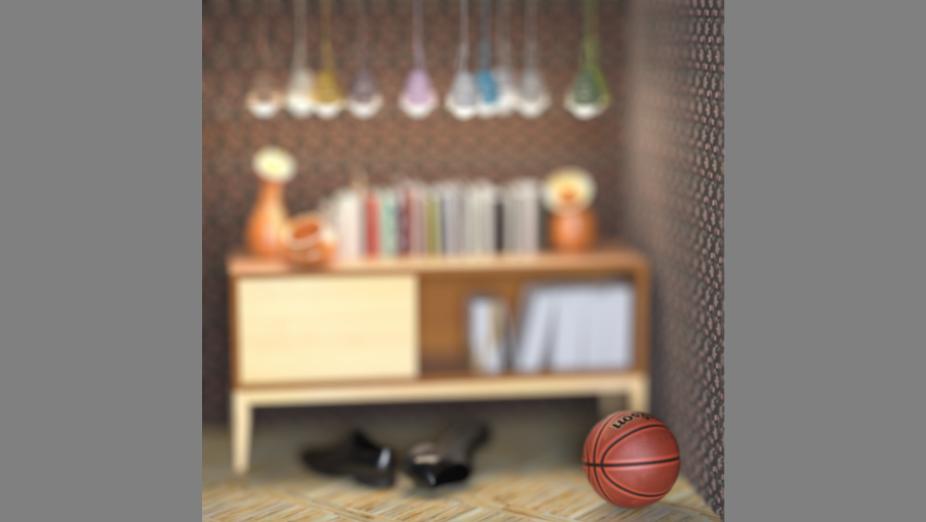}
  \end{subfigure}\hfil
  \begin{subfigure}{0.1555\linewidth}
    \centering
    \includegraphics[width=\linewidth]{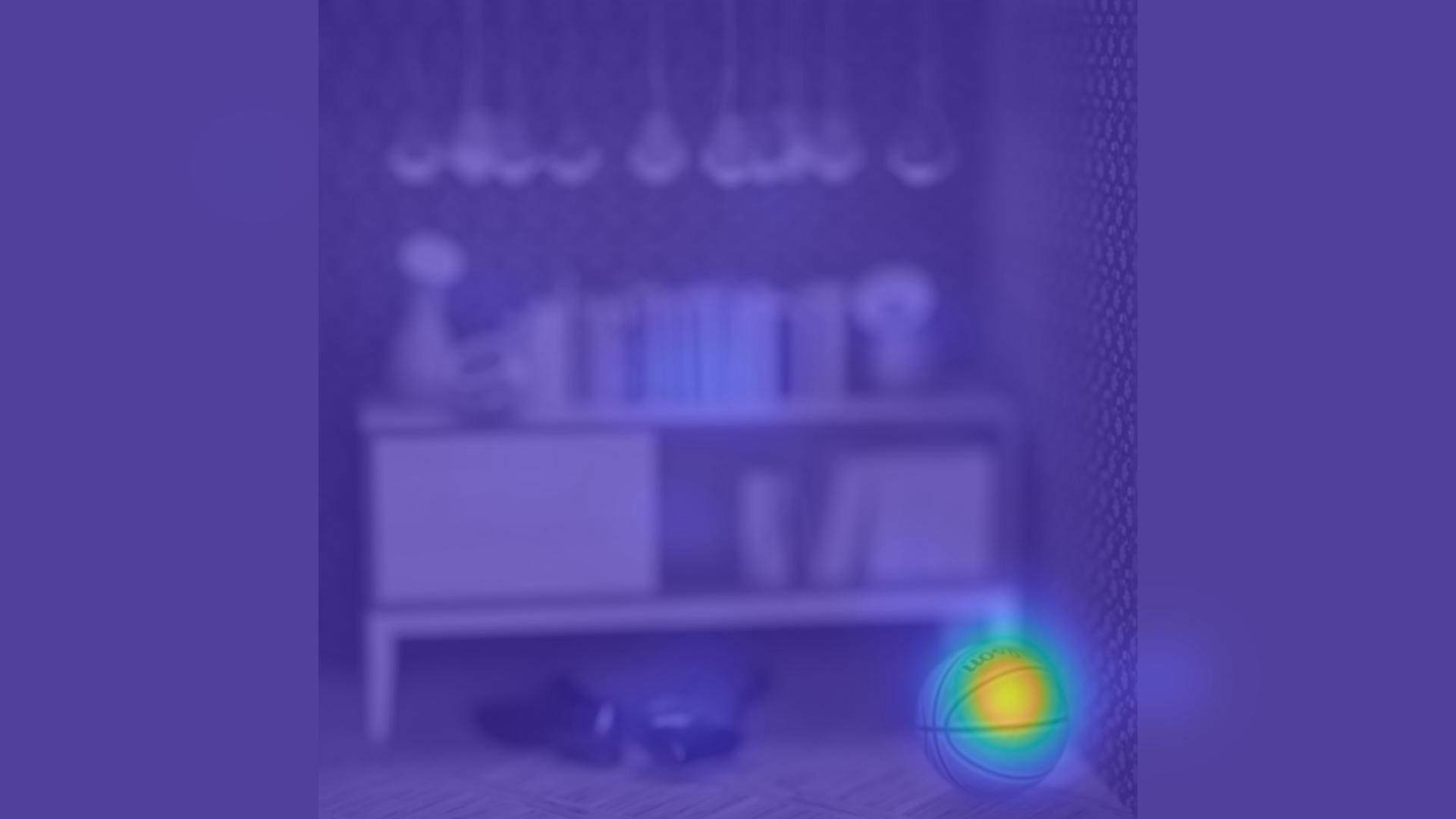}
  \end{subfigure}\hfil
  \begin{subfigure}{0.1555\linewidth}
    \centering
    \includegraphics[width=\linewidth]{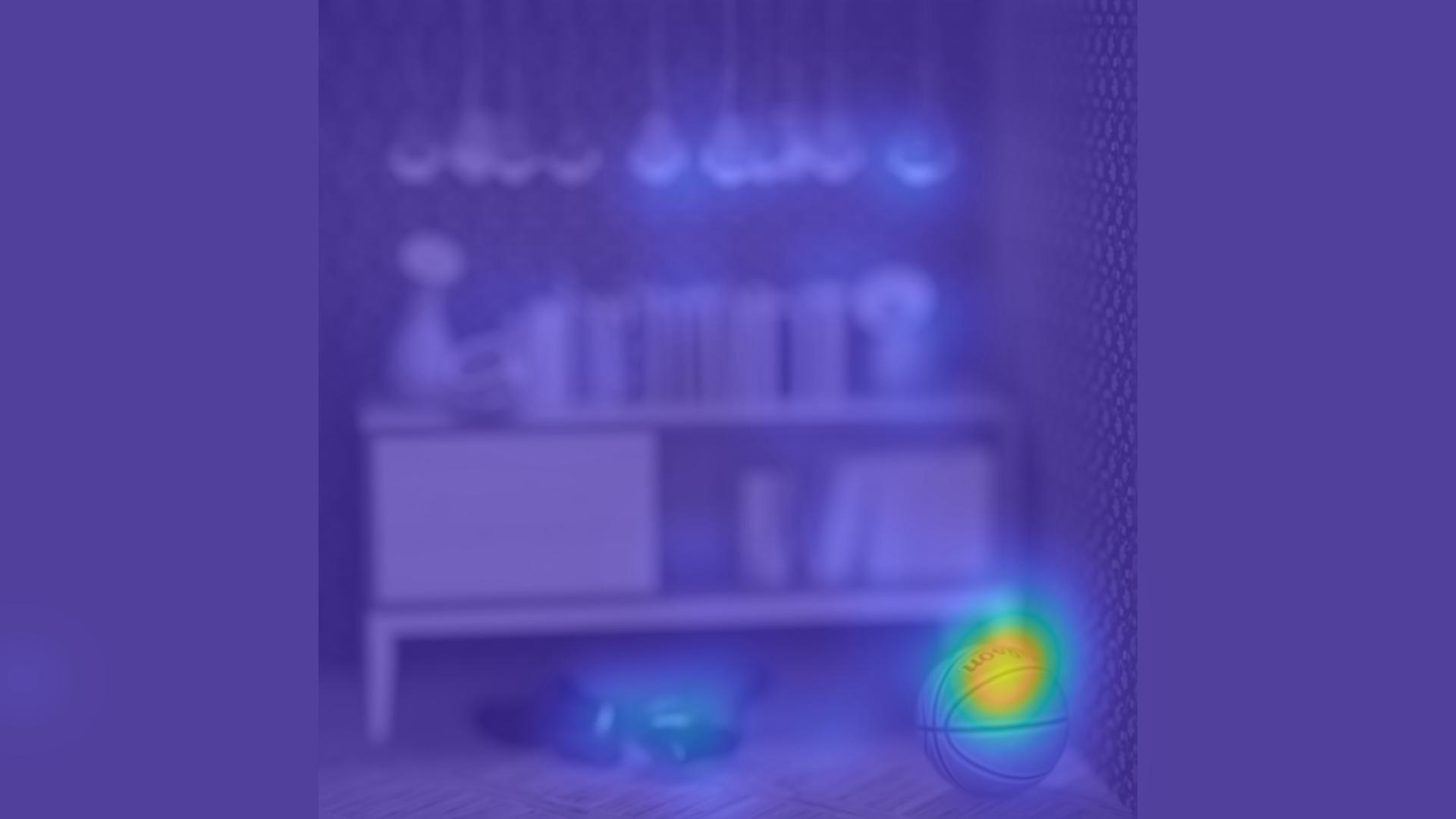}
  \end{subfigure}\hfil
  \begin{subfigure}{0.1555\linewidth}
    \centering
    \includegraphics[width=\linewidth]{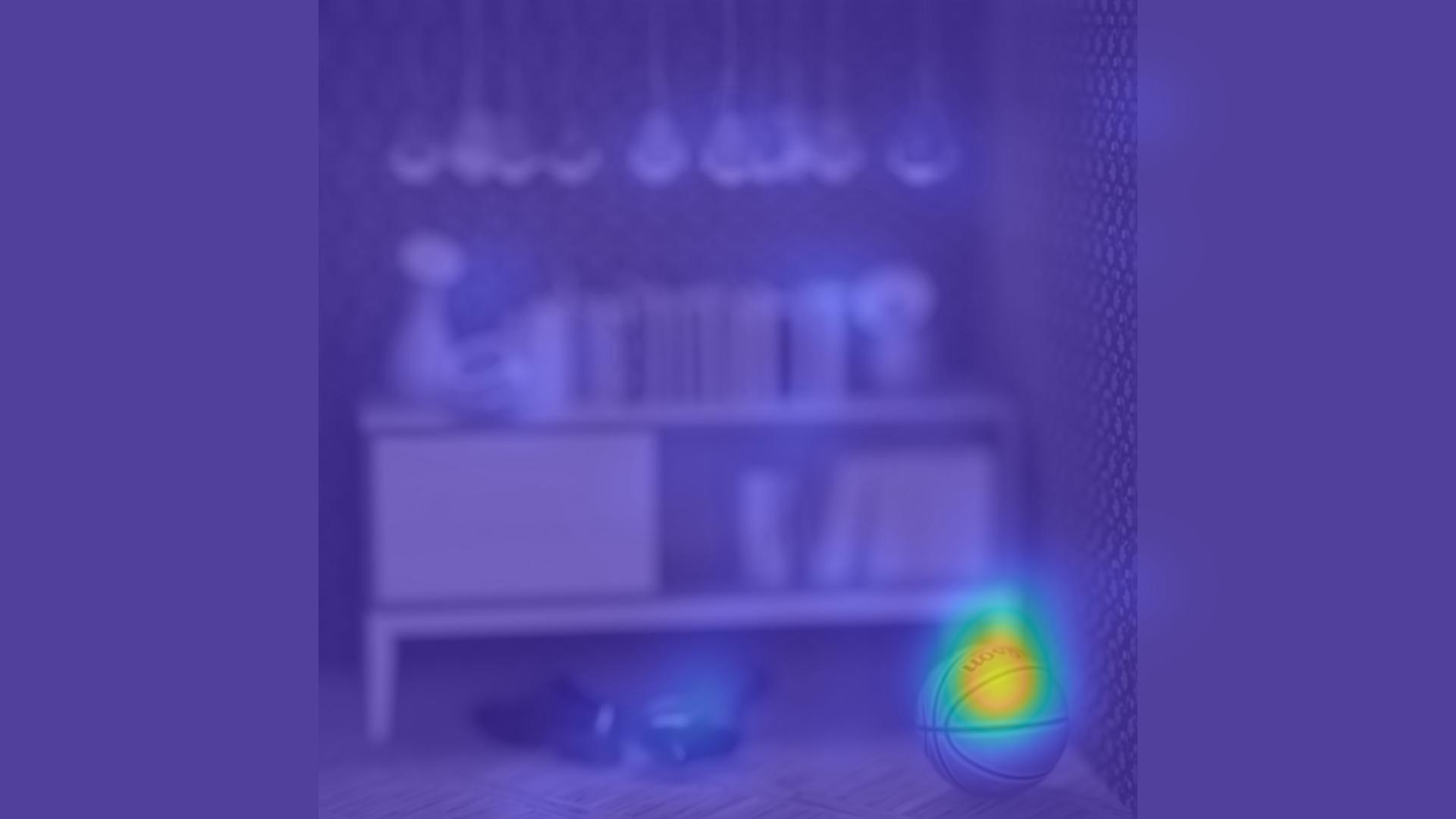}
  \end{subfigure}\hfil
  \begin{subfigure}{0.1555\linewidth}
    \centering
    \includegraphics[width=\linewidth]{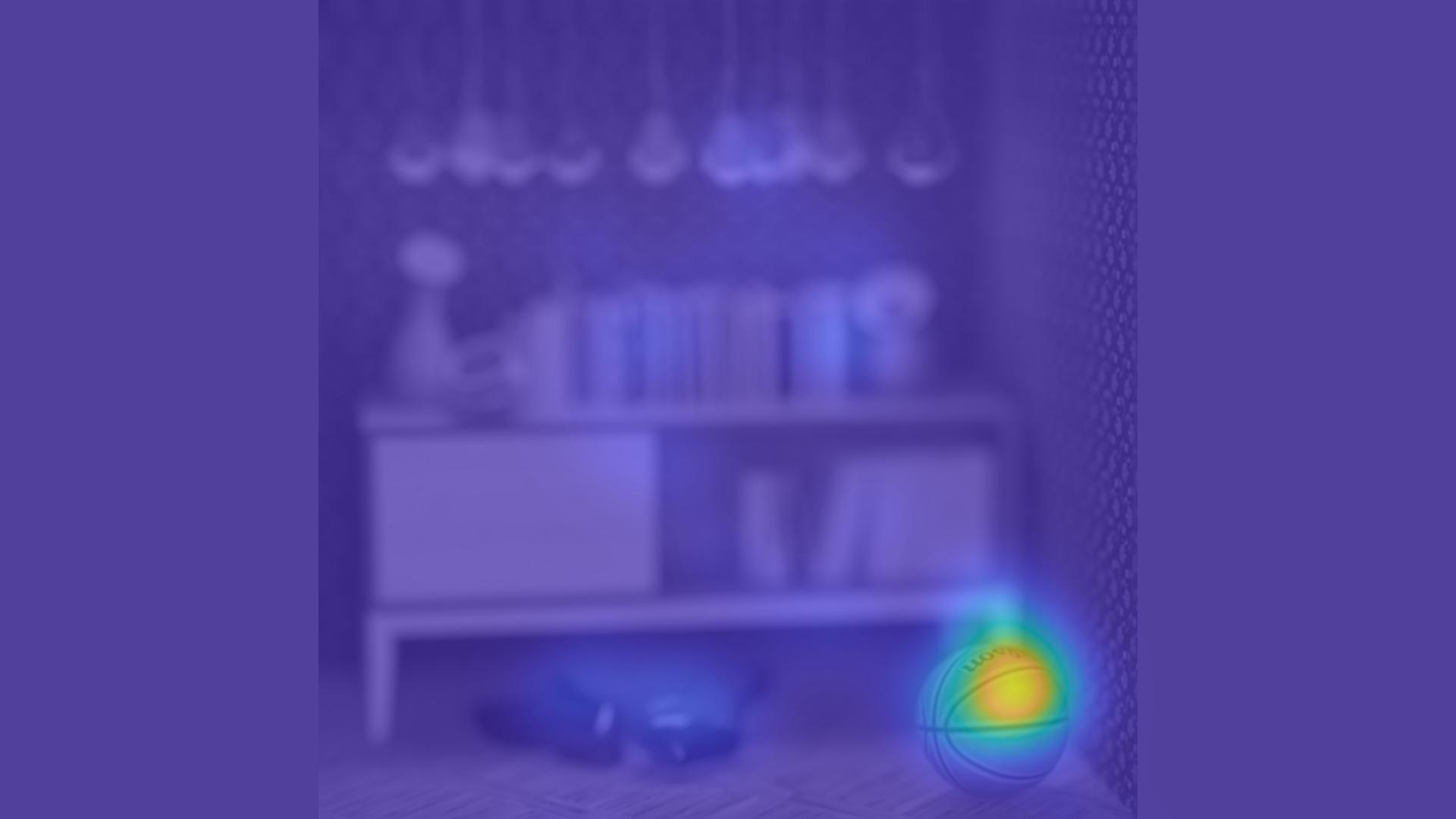}
  \end{subfigure}\hfil
  \begin{subfigure}{0.1555\linewidth}
    \centering
    \includegraphics[width=\linewidth]{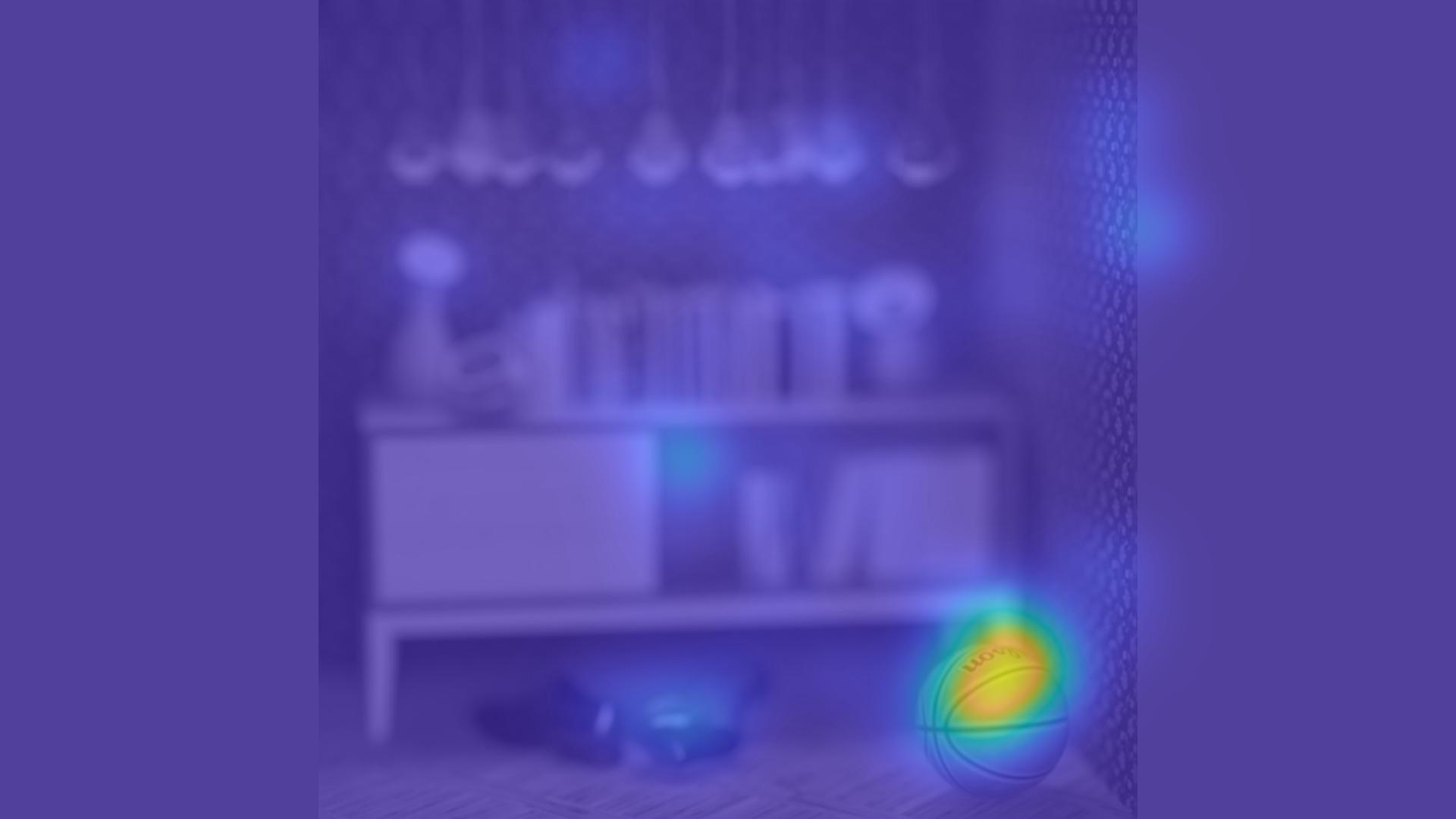}
  \end{subfigure}\hfil

  \vspace{0.3 mm} 
  \rotatebox[origin = c]{90}{(b)}
  \rotatebox[origin = c]{90}{Sideboard}
\rotatebox[origin = c]{90}{AiF}
  \begin{subfigure}{0.1555\linewidth}
    \centering
    \includegraphics[width=\linewidth]{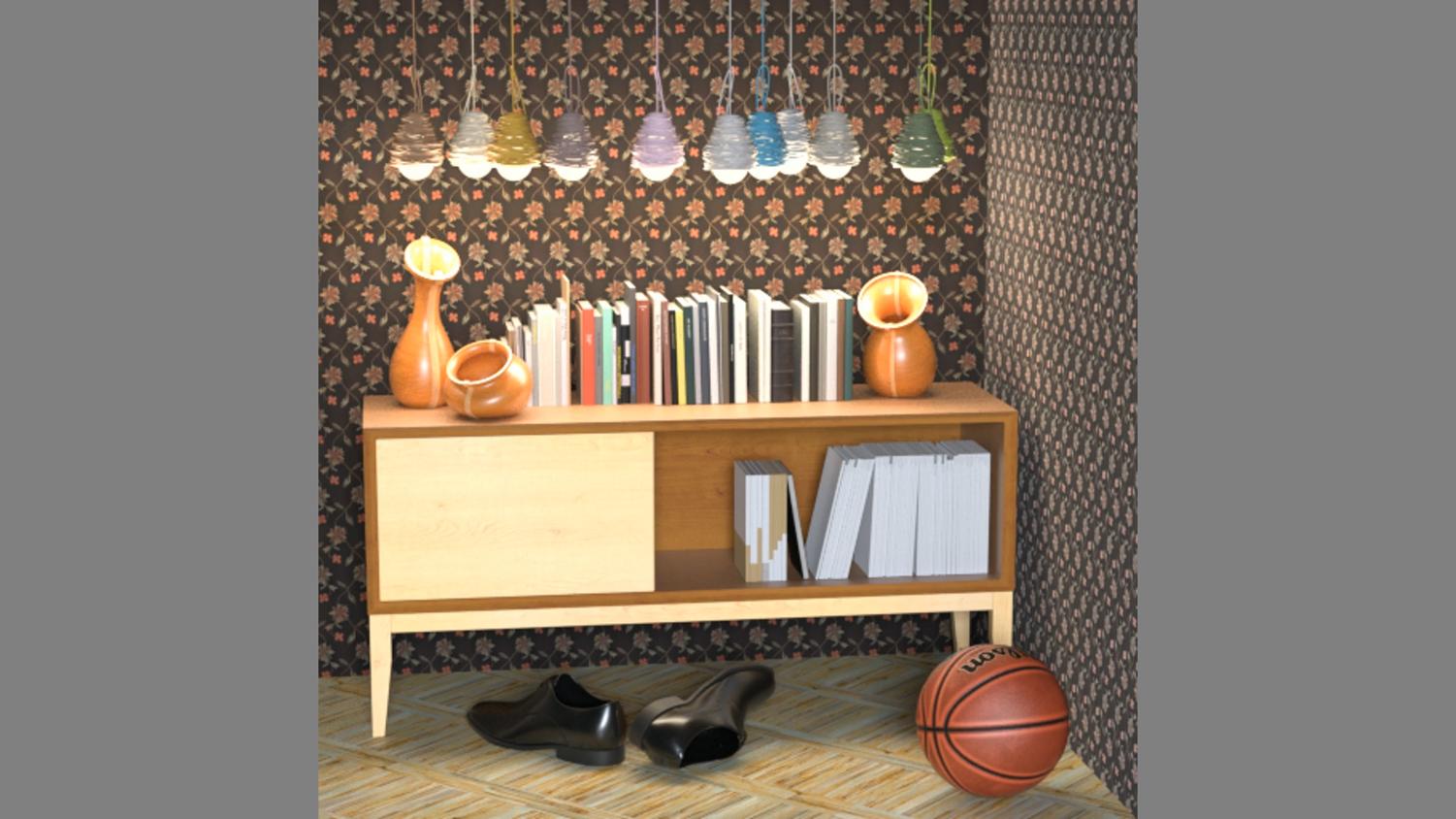}
  \end{subfigure}\hfil
  \begin{subfigure}{0.1555\linewidth}
    \centering
    \includegraphics[width=\linewidth]{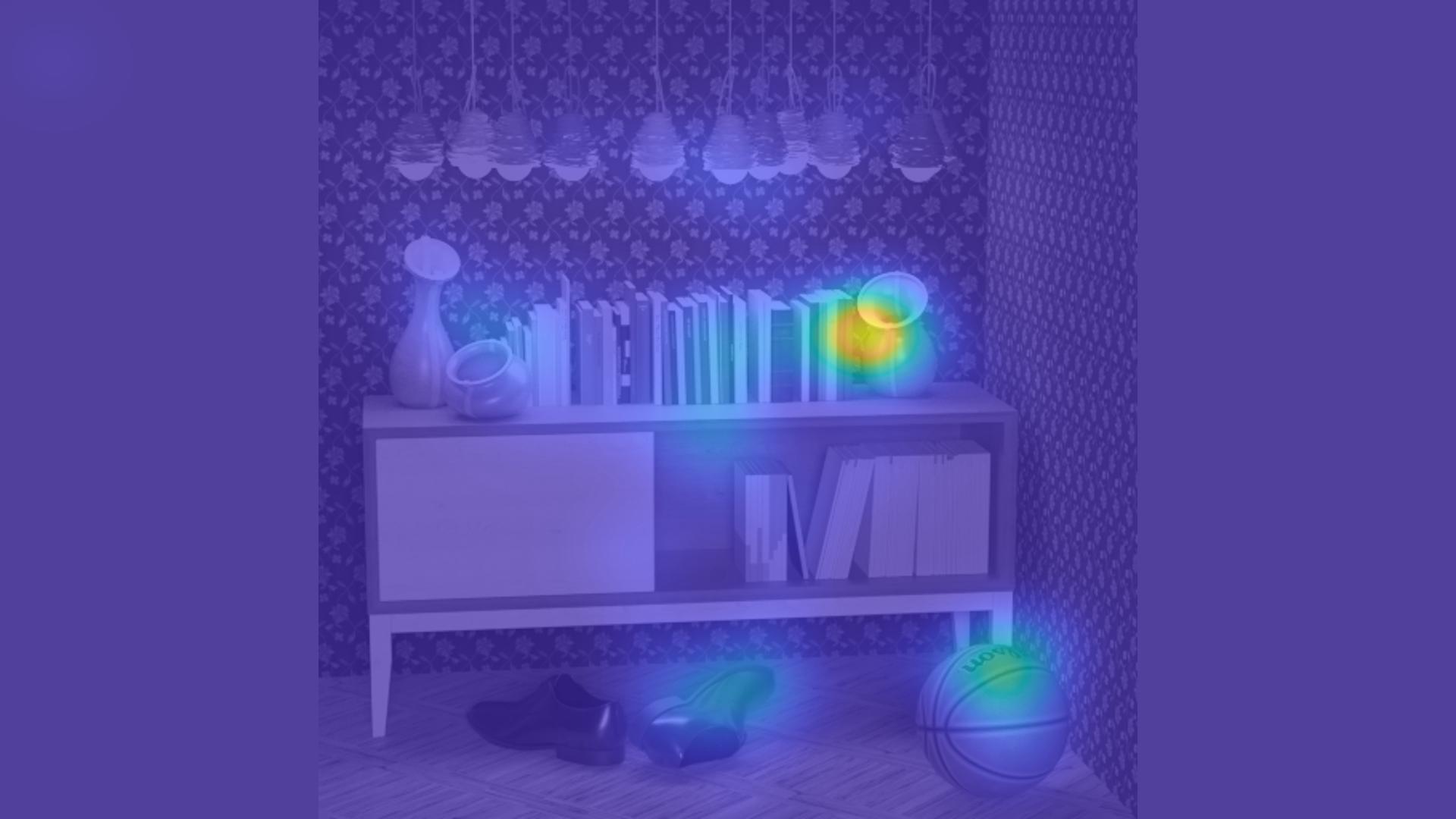}
  \end{subfigure}\hfil
  \begin{subfigure}{0.1555\linewidth}
    \centering
    \includegraphics[width=\linewidth]{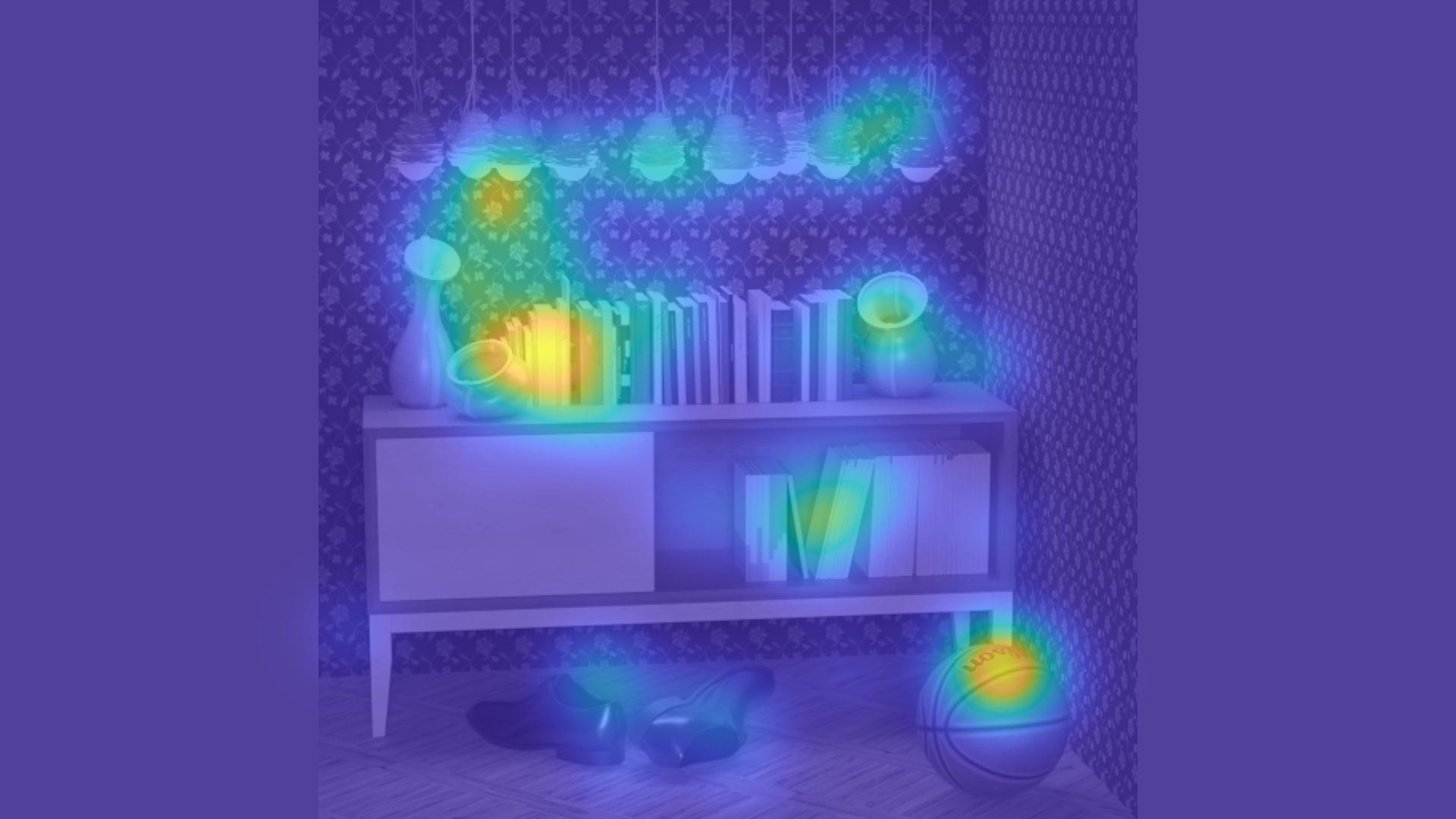}
  \end{subfigure}\hfil
  \begin{subfigure}{0.1555\linewidth}
    \centering
    \includegraphics[width=\linewidth]{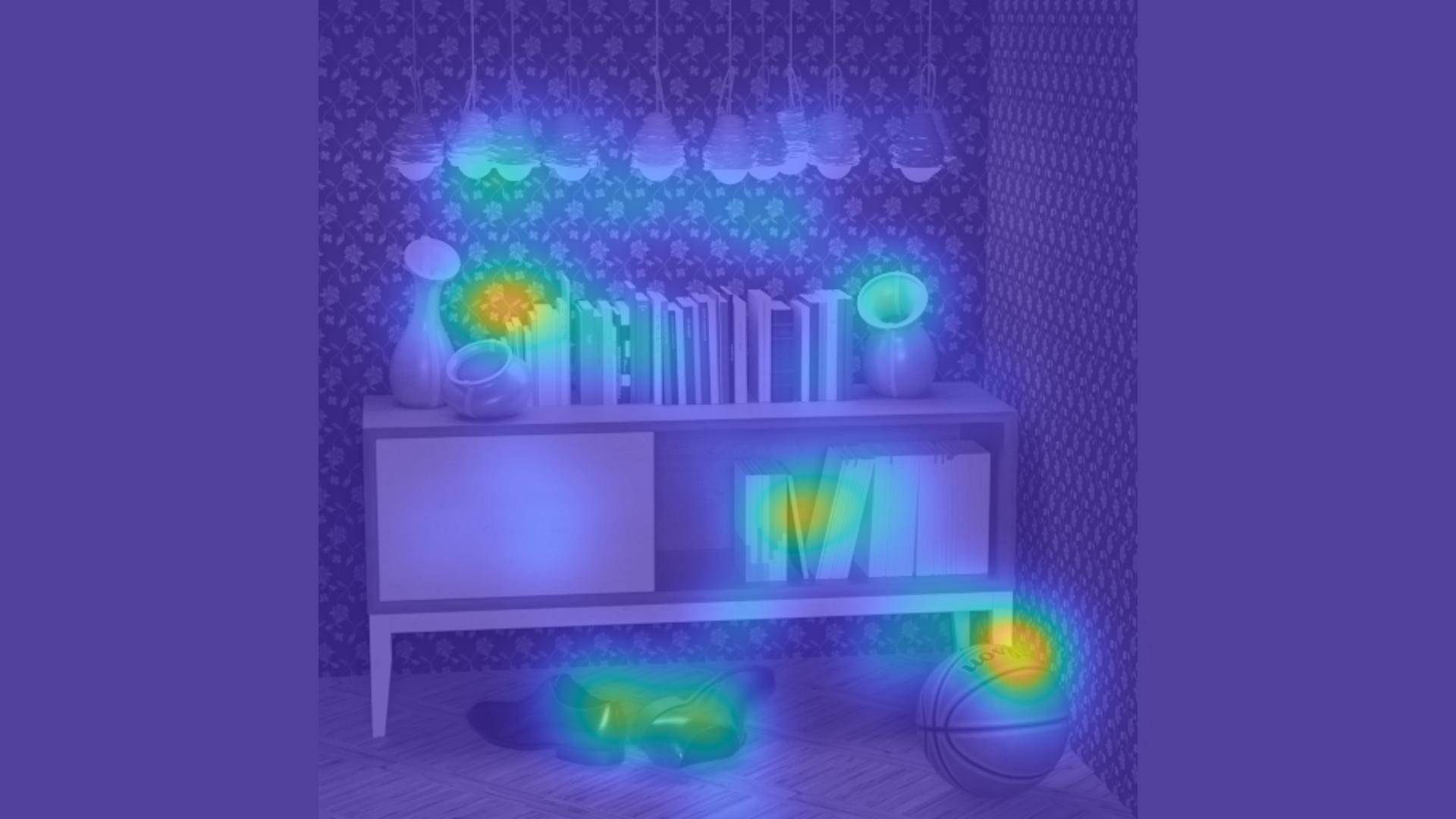}
  \end{subfigure}\hfil
  \begin{subfigure}{0.1555\linewidth}
    \centering
    \includegraphics[width=\linewidth]{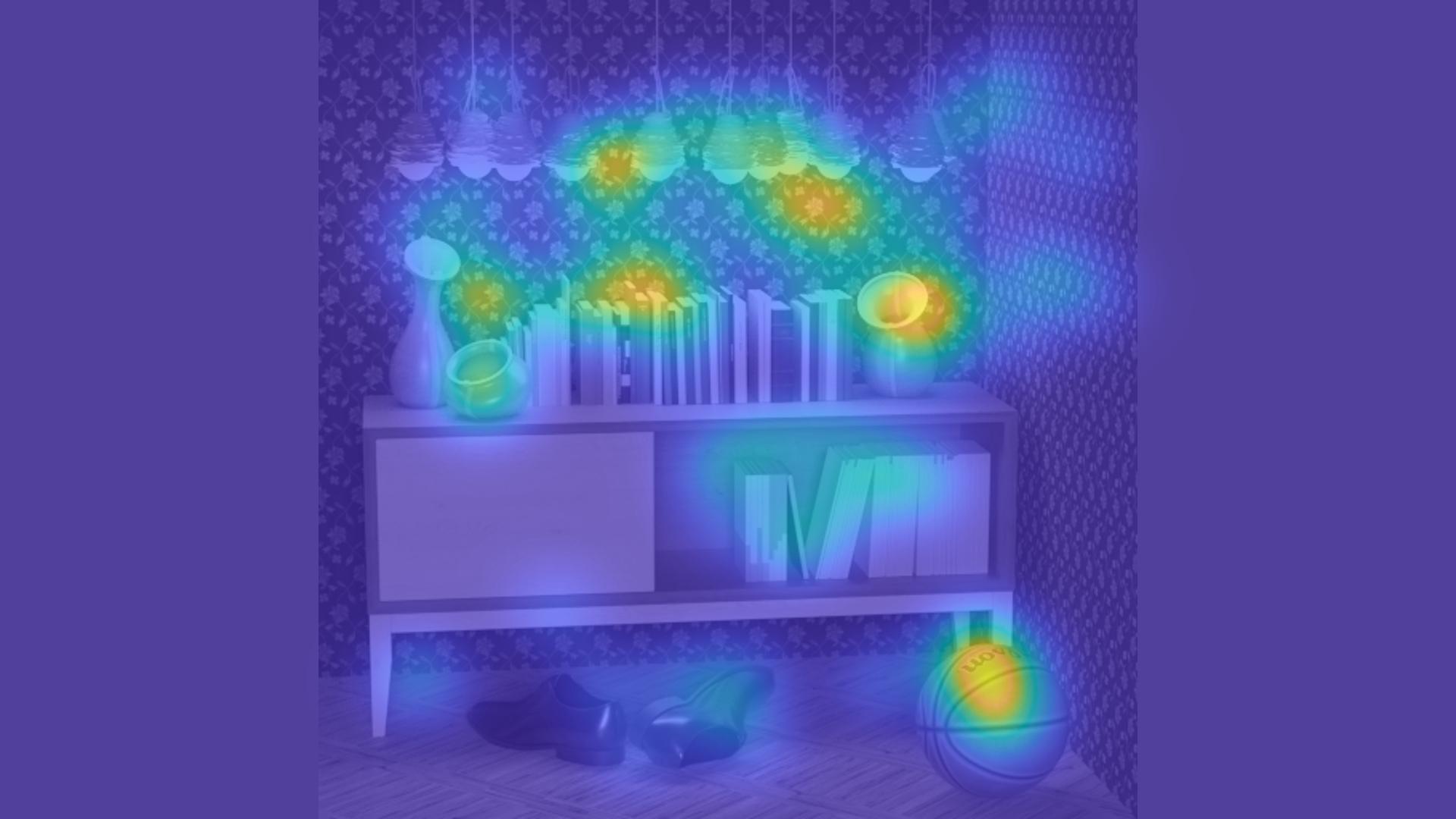}
  \end{subfigure}\hfil
  \begin{subfigure}{0.1555\linewidth}
    \centering
    \includegraphics[width=\linewidth]{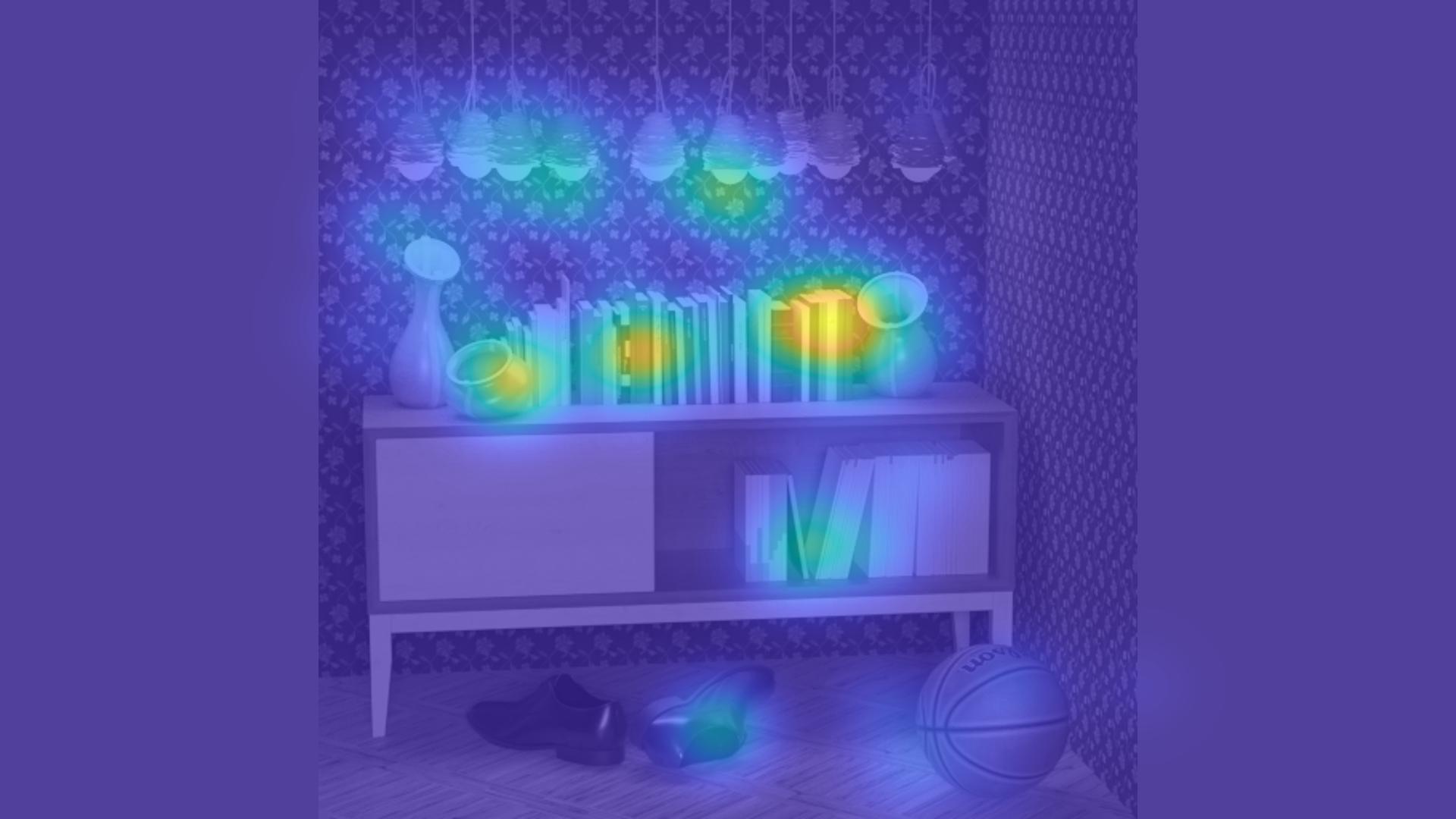}
  \end{subfigure}\hfil

  \vspace{0.3 mm} 
  \hrule %\noindent\hrulefill \\
  \vspace{0.3 mm} 
  \rotatebox[origin = c]{90}{(c)}
\rotatebox[origin = c]{90}{Tarot-S}
\rotatebox[origin = c]{90}{RiF2}
  \begin{subfigure}{0.1555\linewidth}
    \centering
    \includegraphics[width=\linewidth]{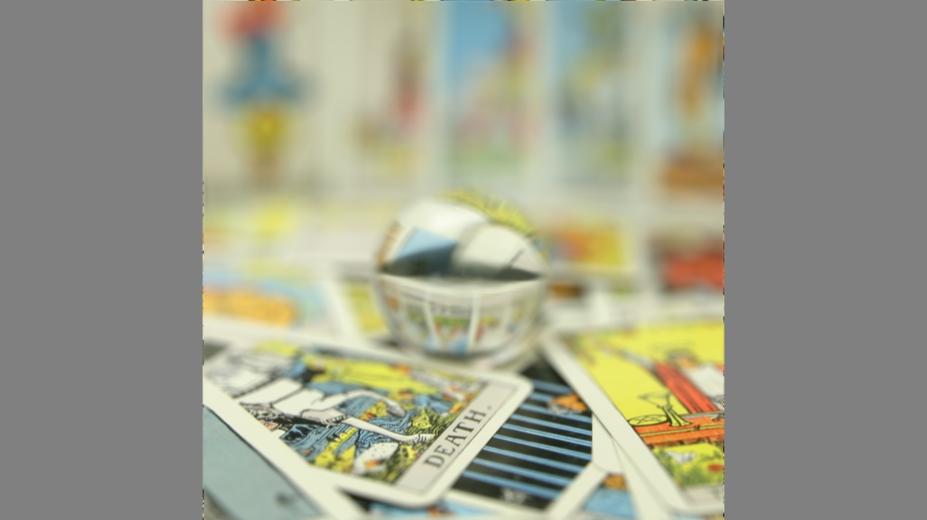}
  \end{subfigure}\hfil
  \begin{subfigure}{0.1555\linewidth}
    \centering
    \includegraphics[width=\linewidth]{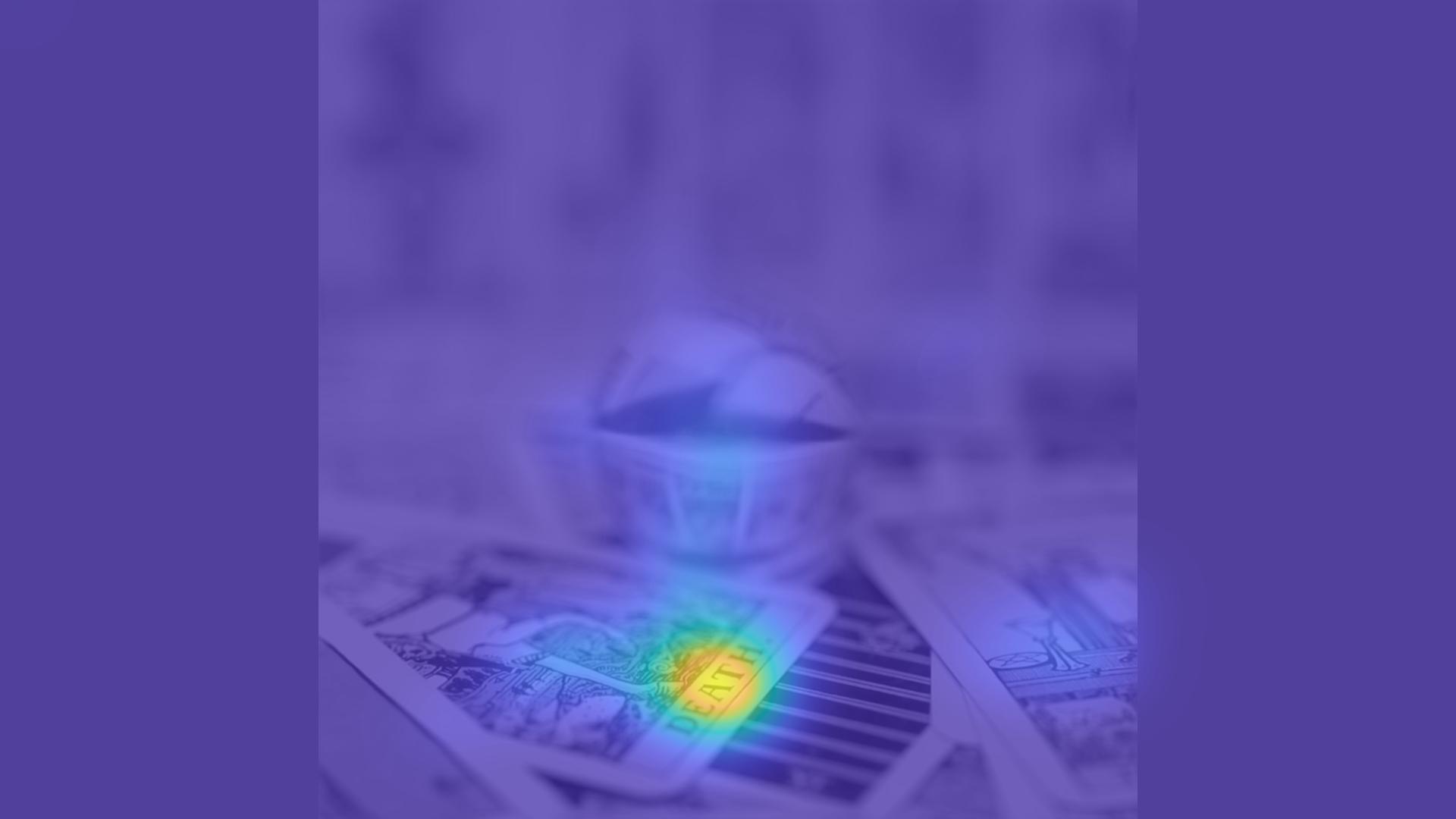}
  \end{subfigure}\hfil
  \begin{subfigure}{0.1555\linewidth}
    \centering
    \includegraphics[width=\linewidth]{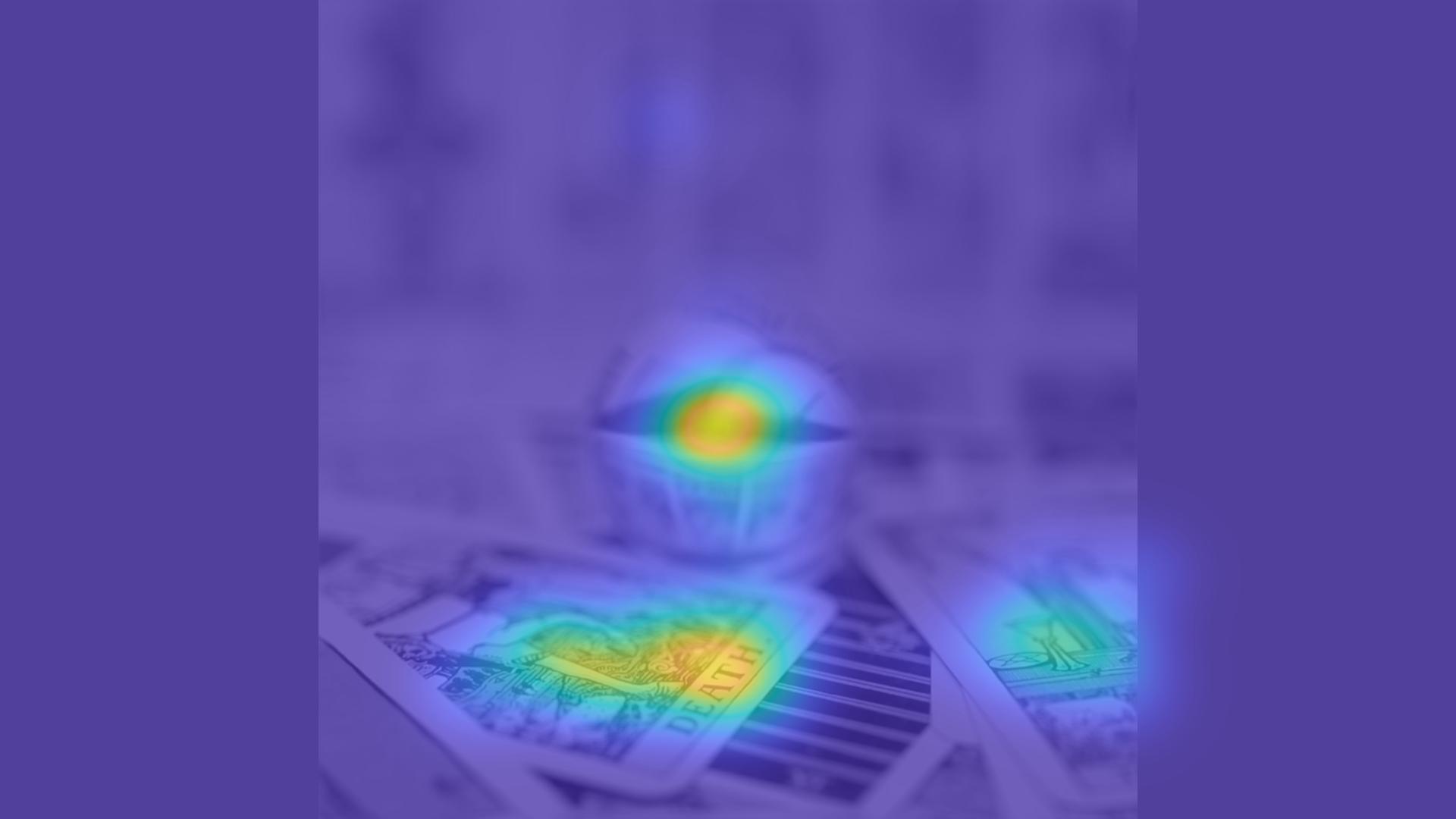}
  \end{subfigure}\hfil
  \begin{subfigure}{0.1555\linewidth}
    \centering
    \includegraphics[width=\linewidth]{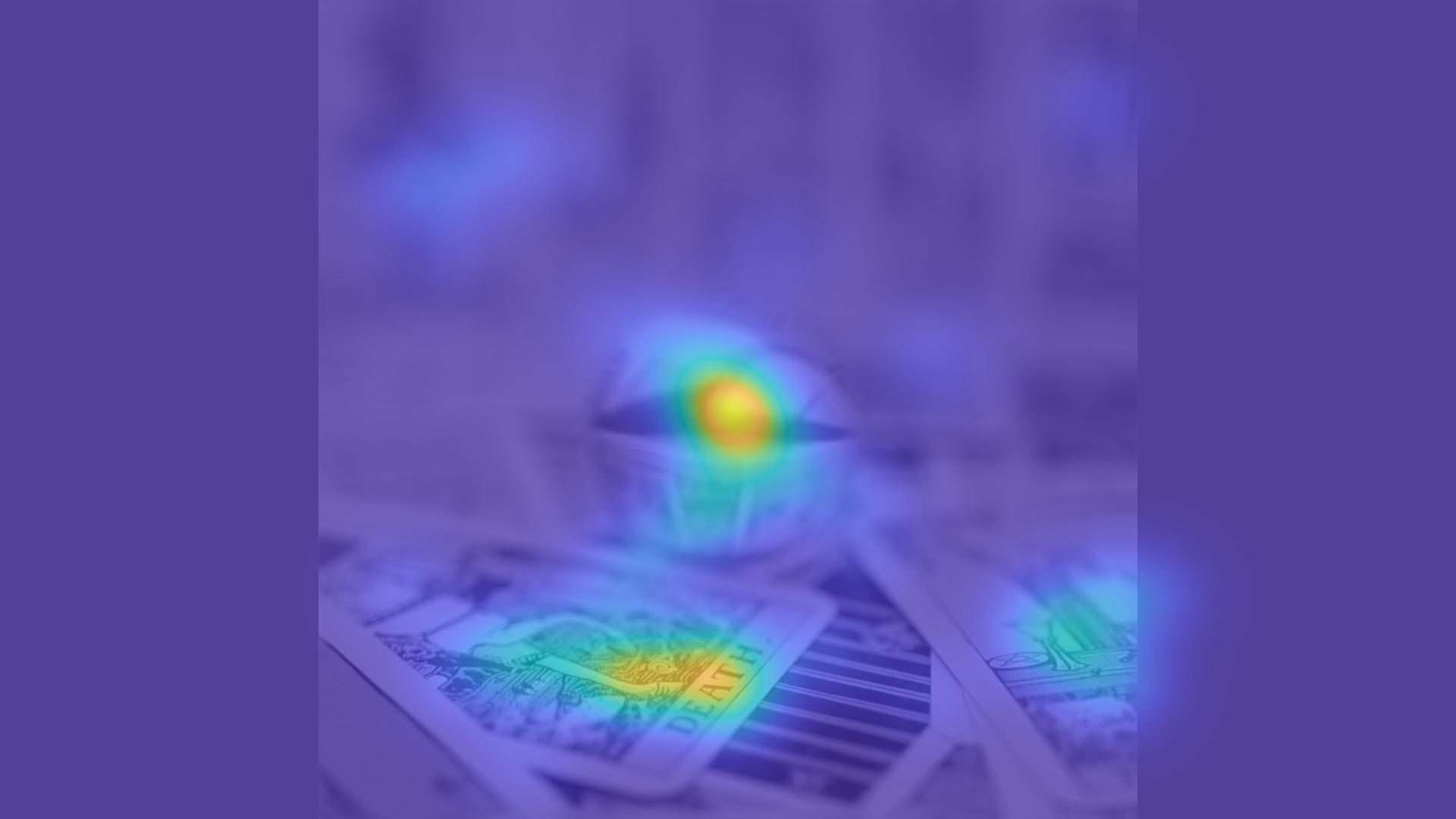}
  \end{subfigure}\hfil
  \begin{subfigure}{0.1555\linewidth}
    \centering
    \includegraphics[width=\linewidth]{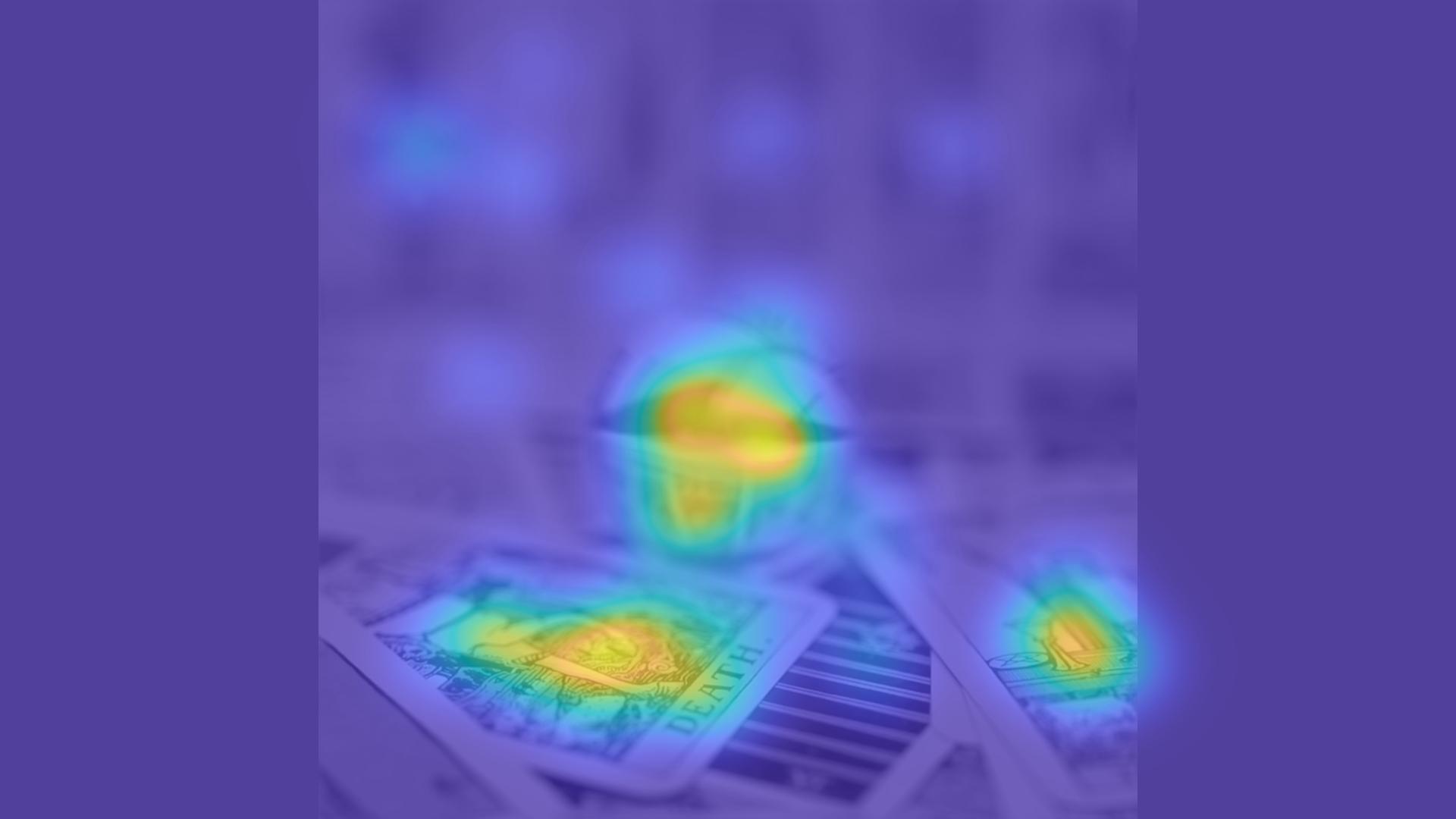}
  \end{subfigure}\hfil
  \begin{subfigure}{0.1555\linewidth}
    \centering
    \includegraphics[width=\linewidth]{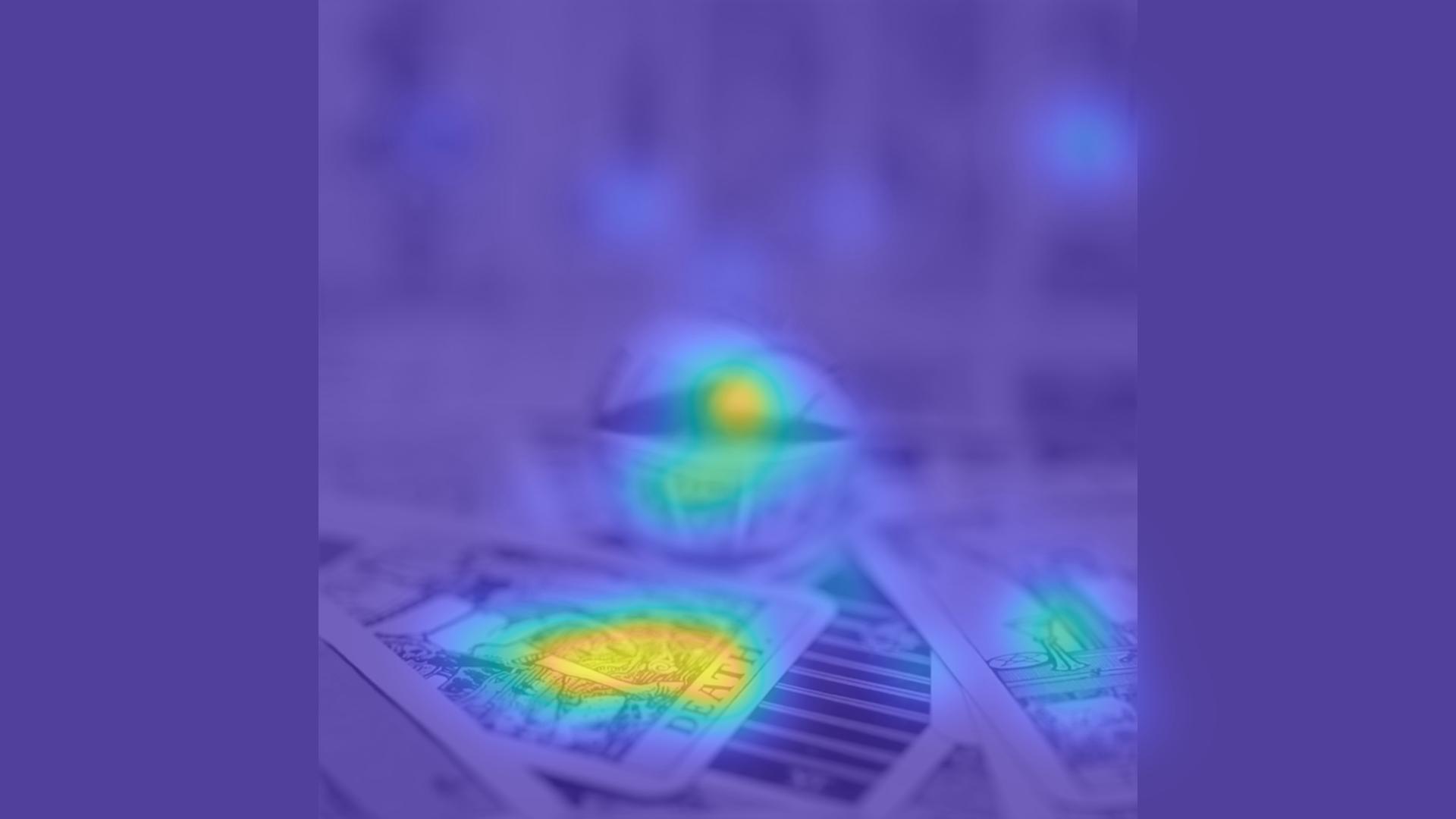}
  \end{subfigure}\hfil

  \vspace{0.3 mm} 
  \rotatebox[origin = c]{90}{(d)}
\rotatebox[origin = c]{90}{Tarot-S}
\rotatebox[origin = c]{90}{AiF}
  \begin{subfigure}{0.1555\linewidth}
    \centering
    \includegraphics[width=\linewidth]{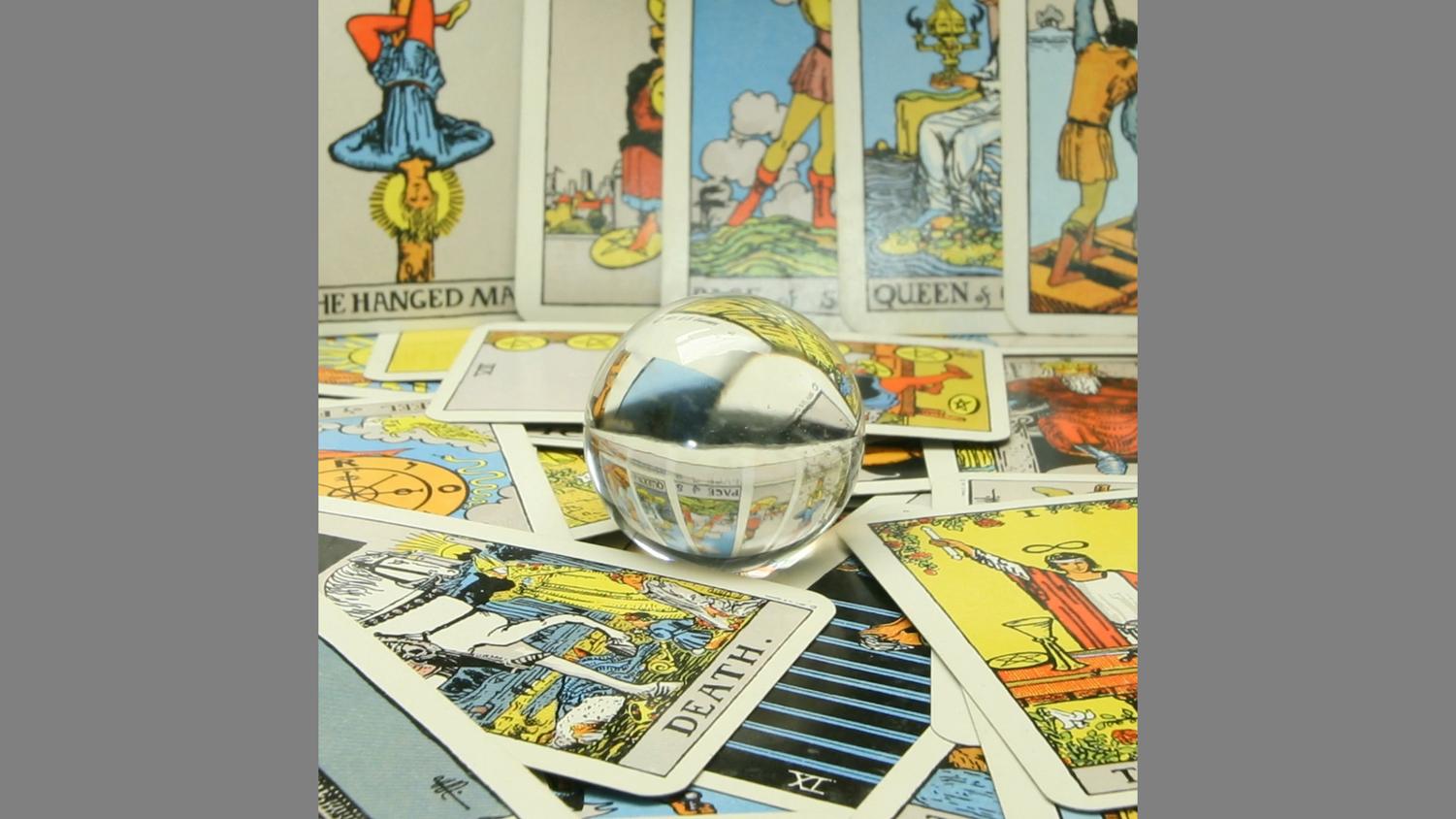}
  \end{subfigure}\hfil
  \begin{subfigure}{0.1555\linewidth}
    \centering
    \includegraphics[width=\linewidth]{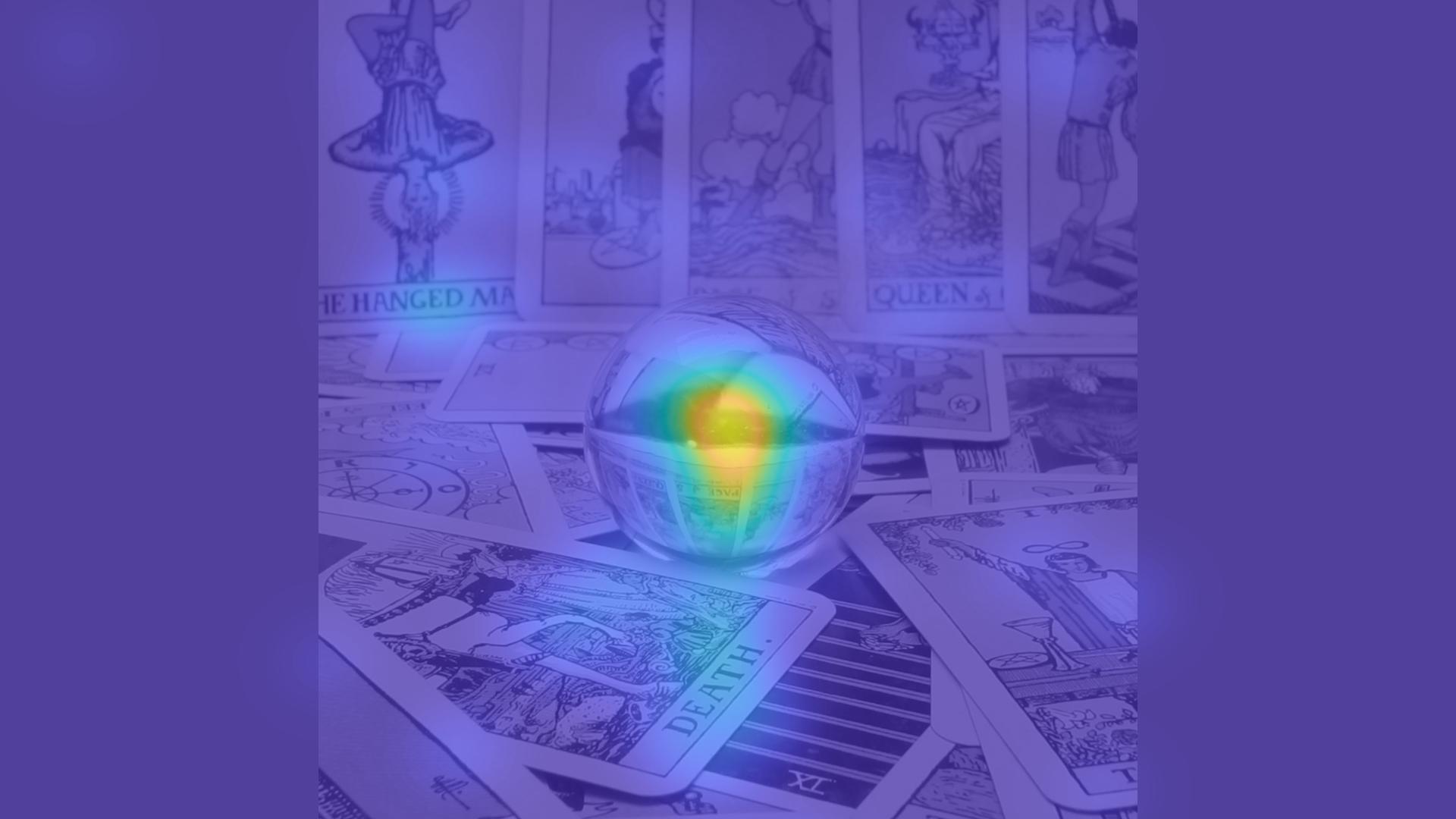}
  \end{subfigure}\hfil
  \begin{subfigure}{0.1555\linewidth}
    \centering
    \includegraphics[width=\linewidth]{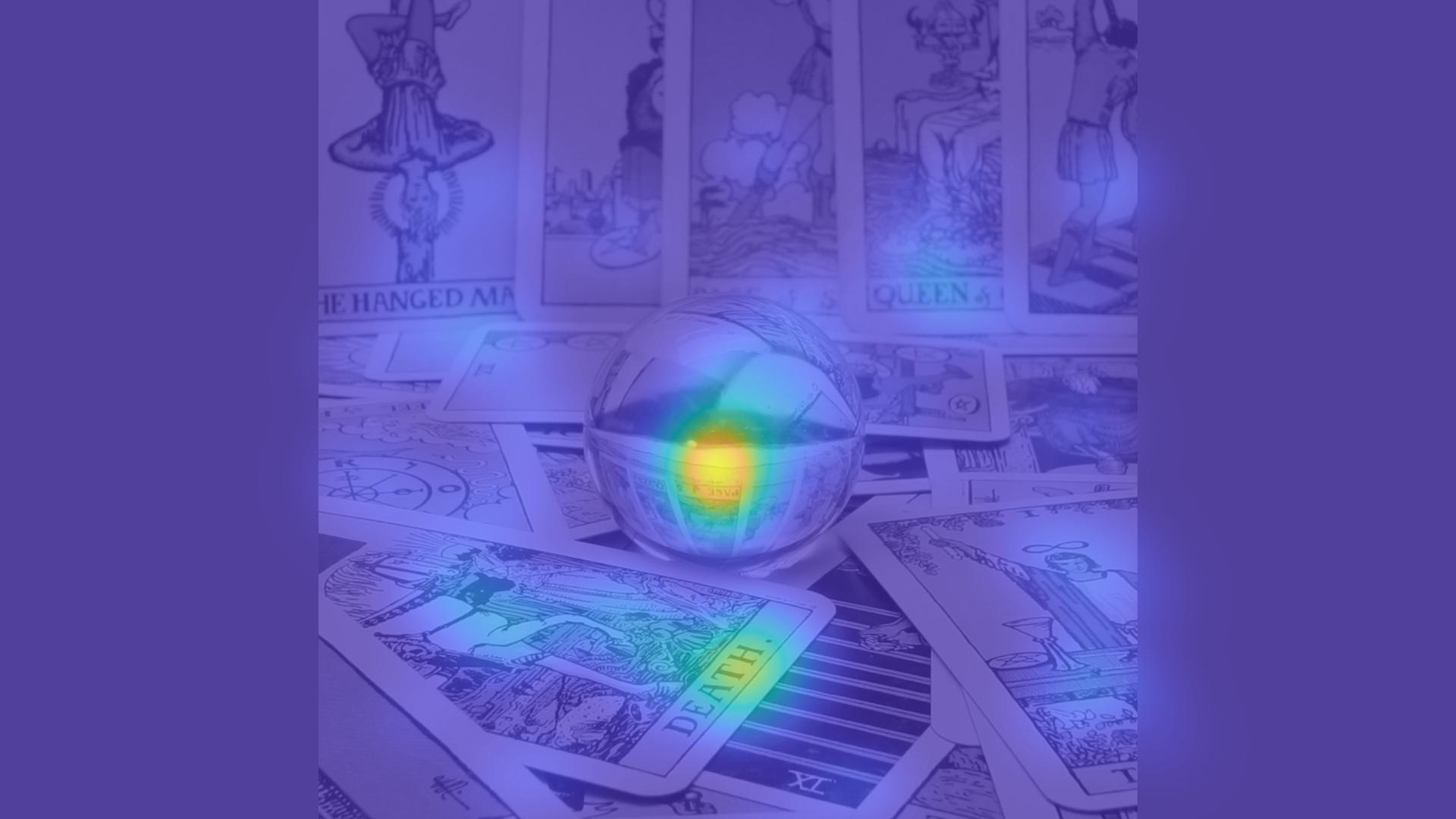}
  \end{subfigure}\hfil
  \begin{subfigure}{0.1555\linewidth}
    \centering
    \includegraphics[width=\linewidth]{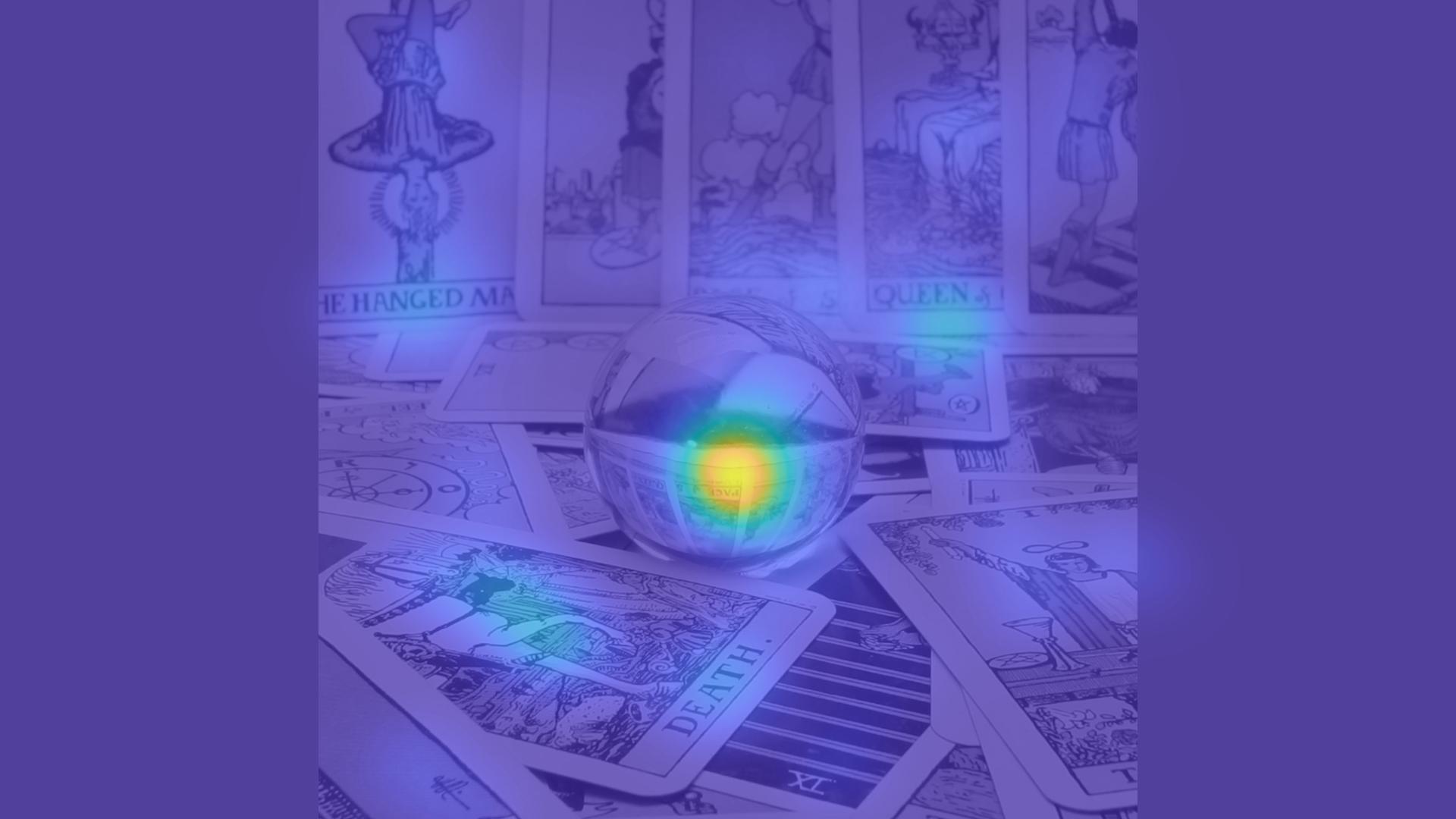}
  \end{subfigure}\hfil
  \begin{subfigure}{0.1555\linewidth}
    \centering
    \includegraphics[width=\linewidth]{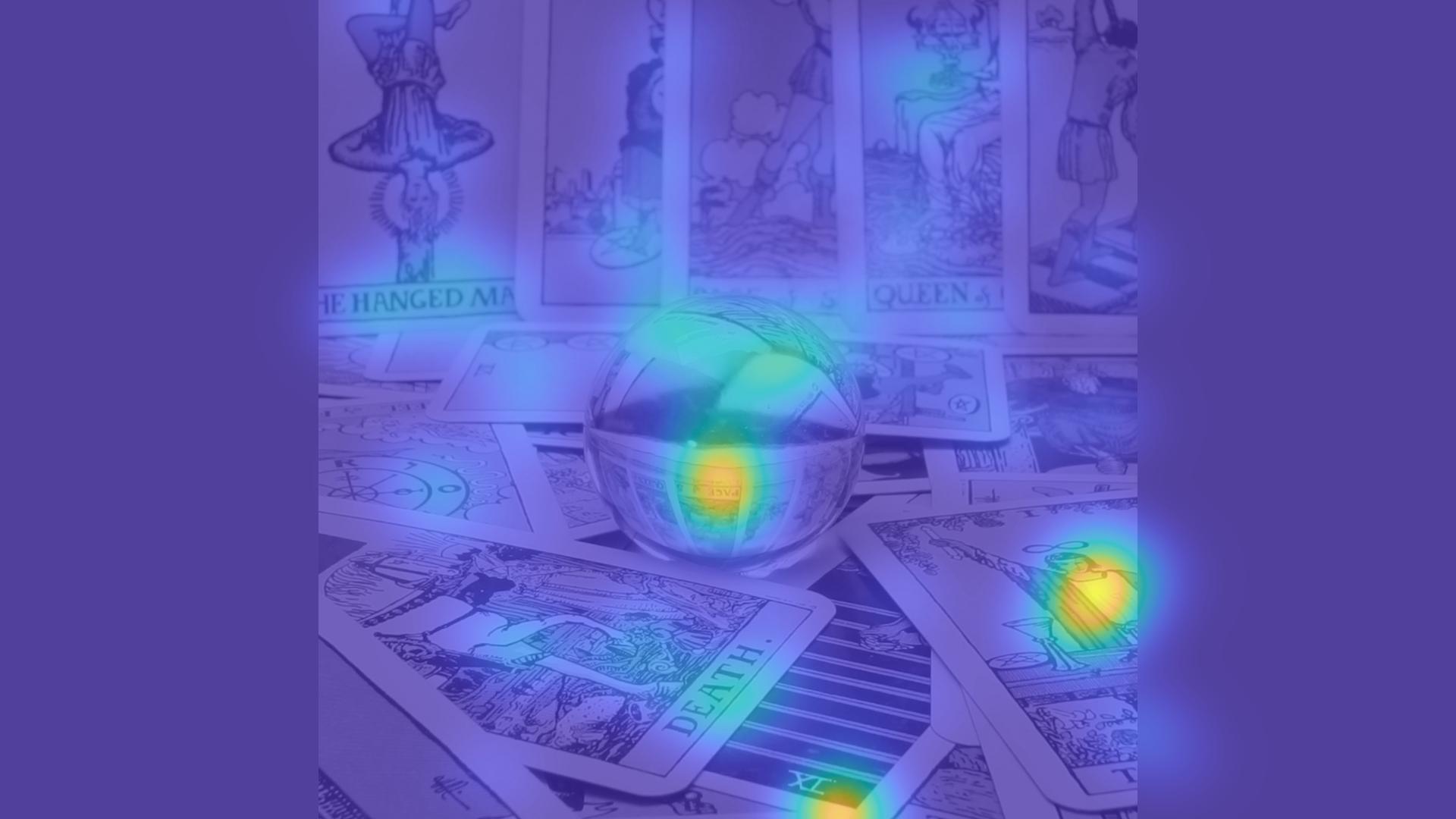}
  \end{subfigure}\hfil
  \begin{subfigure}{0.1555\linewidth}
    \centering
    \includegraphics[width=\linewidth]{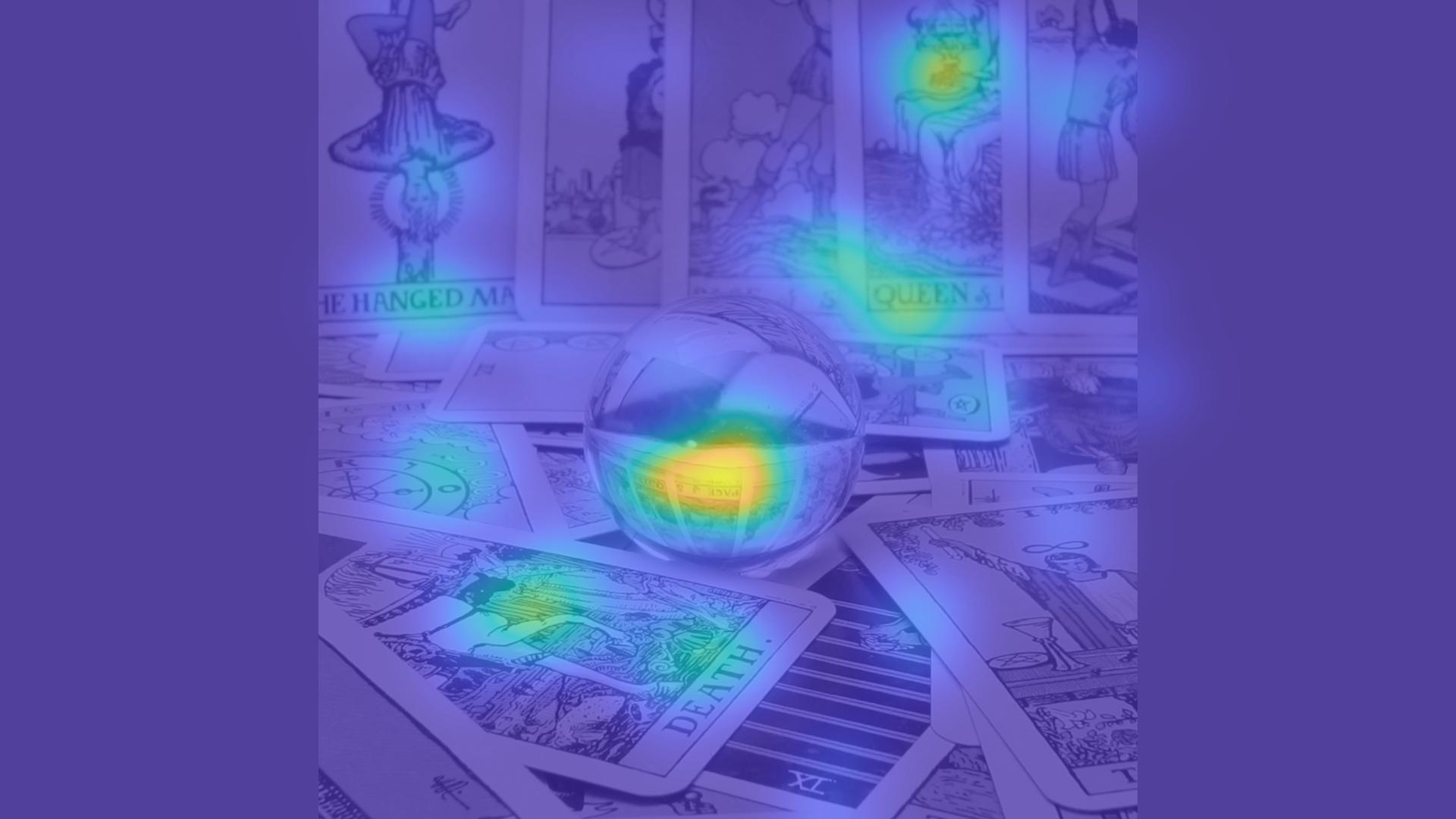}
  \end{subfigure}\hfil

  \vspace{0.3 mm} 
  \hrule %\noindent\hrulefill \\
  \vspace{0.3 mm} 
  \rotatebox[origin = c]{90}{(e)}
\rotatebox[origin = c]{90}{Treasure}
\rotatebox[origin = c]{90}{AiF}
  \begin{subfigure}{0.1555\linewidth}
    \centering
    \includegraphics[width=\linewidth]{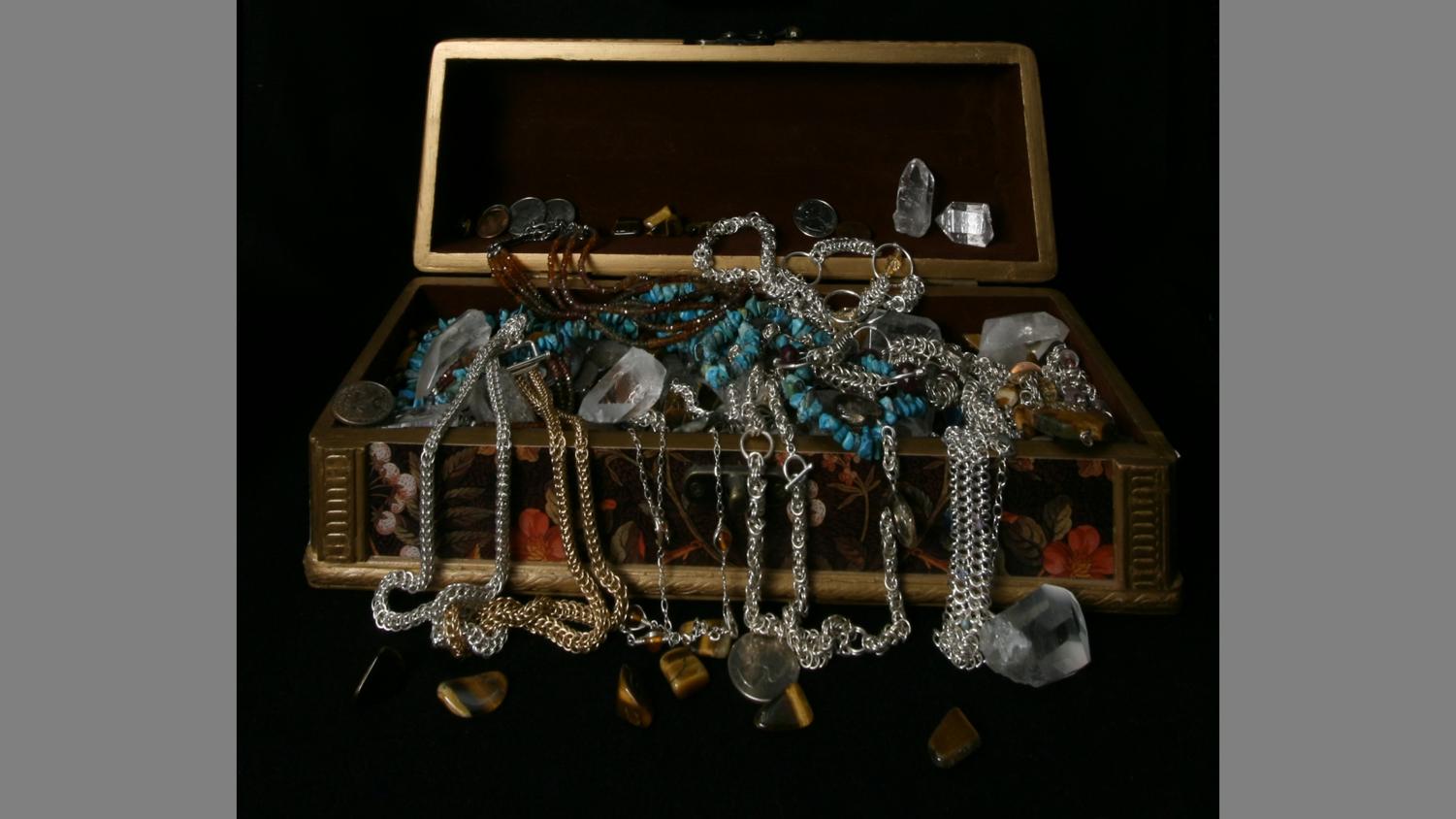}
  \end{subfigure}\hfil
  \begin{subfigure}{0.1555\linewidth}
    \centering
    \includegraphics[width=\linewidth]{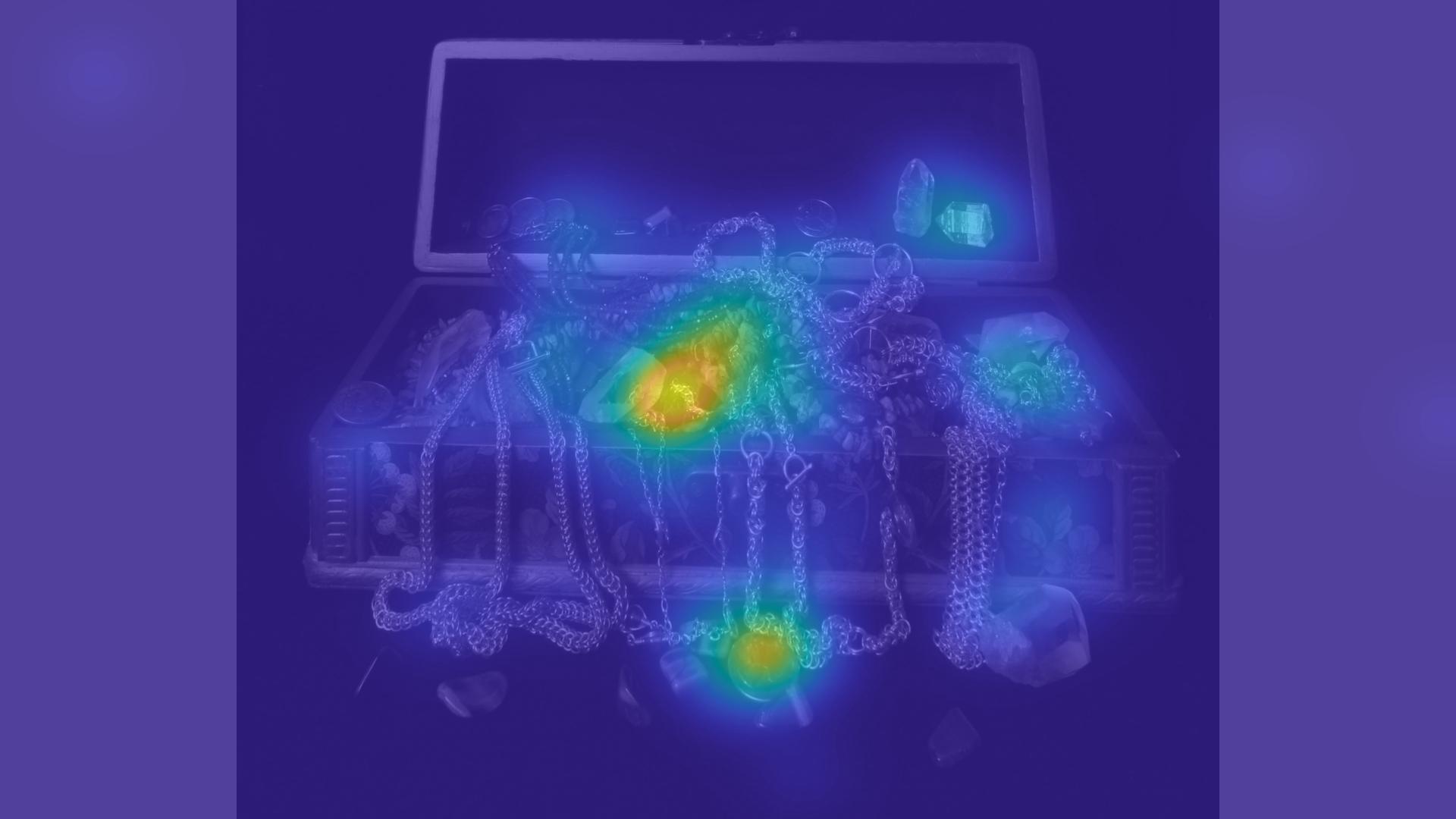}
  \end{subfigure}\hfil
  \begin{subfigure}{0.1555\linewidth}
    \centering
    \includegraphics[width=\linewidth]{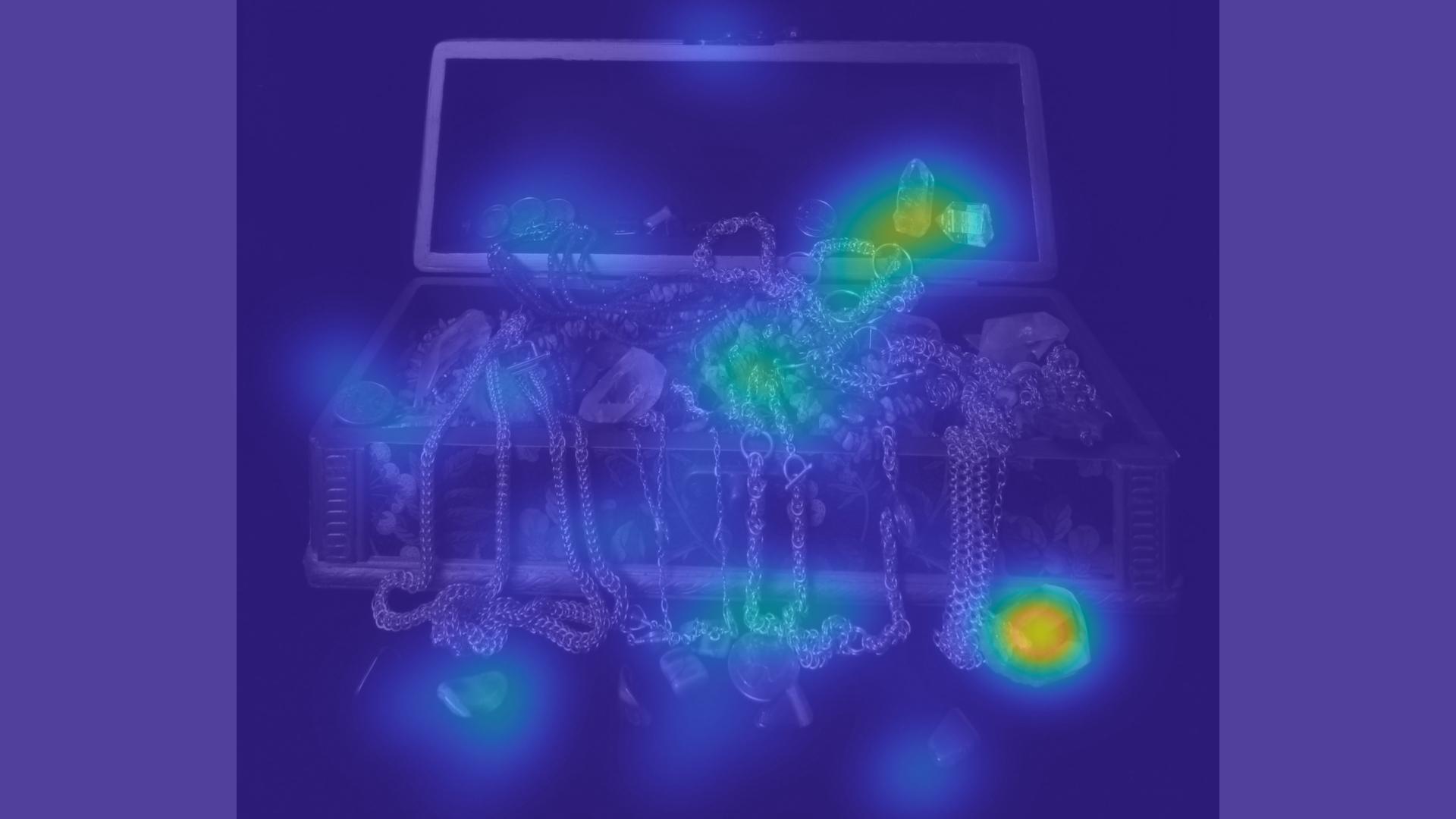}
  \end{subfigure}\hfil
  \begin{subfigure}{0.1555\linewidth}
    \centering
    \includegraphics[width=\linewidth]{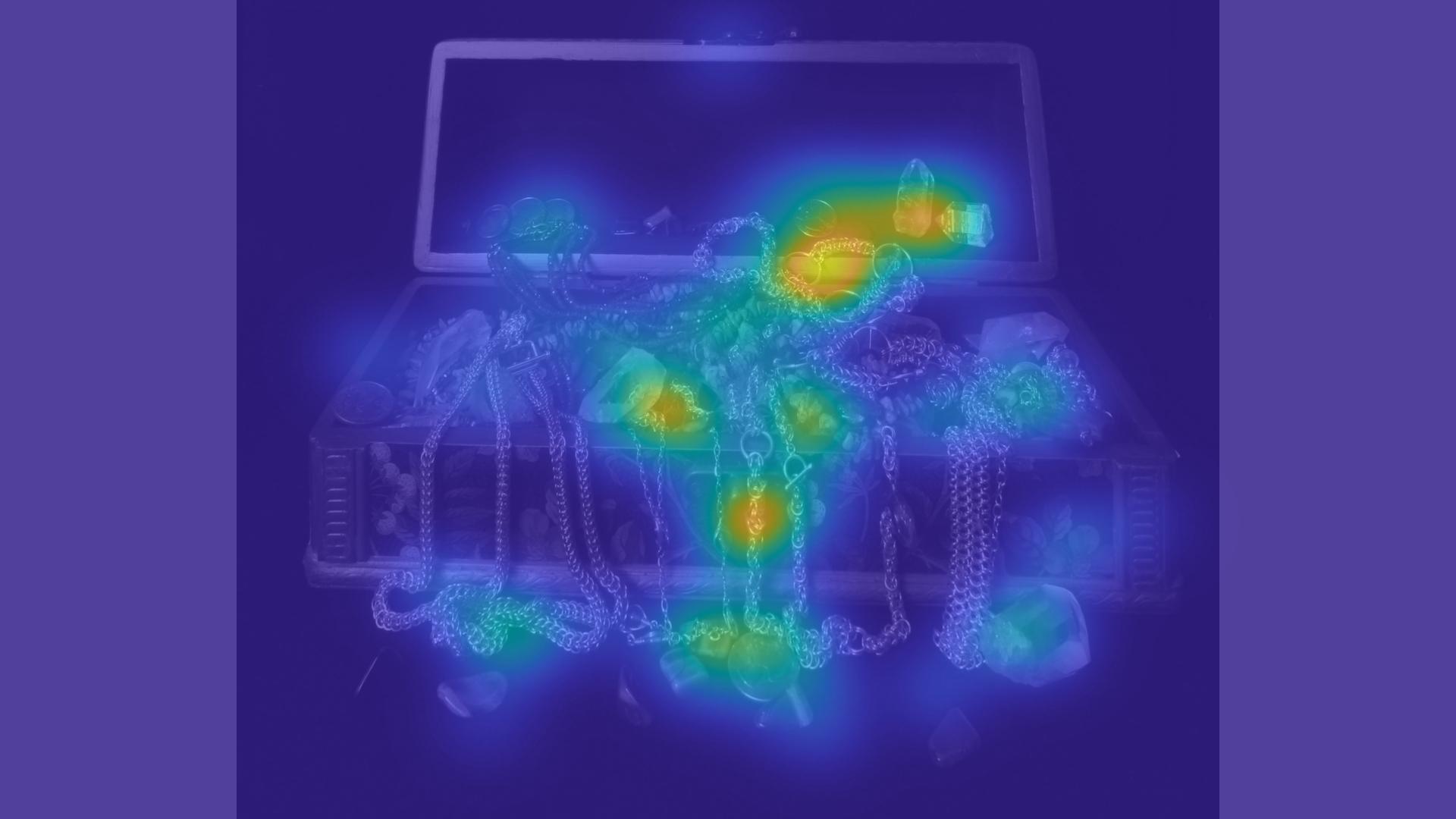}
  \end{subfigure}\hfil
  \begin{subfigure}{0.1555\linewidth}
    \centering
    \includegraphics[width=\linewidth]{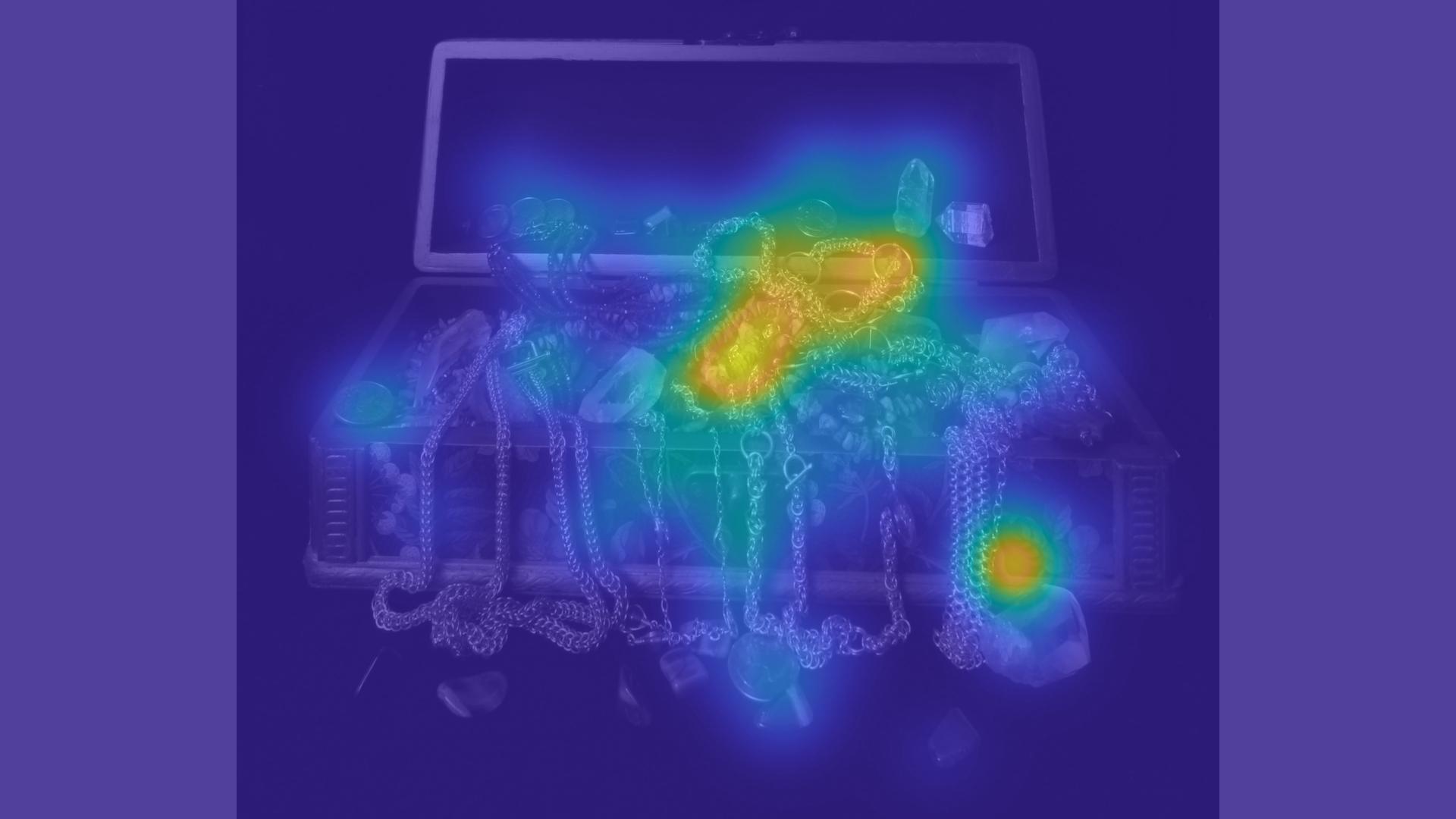}
  \end{subfigure}\hfil
  \begin{subfigure}{0.1555\linewidth}
    \centering
    \includegraphics[width=\linewidth]{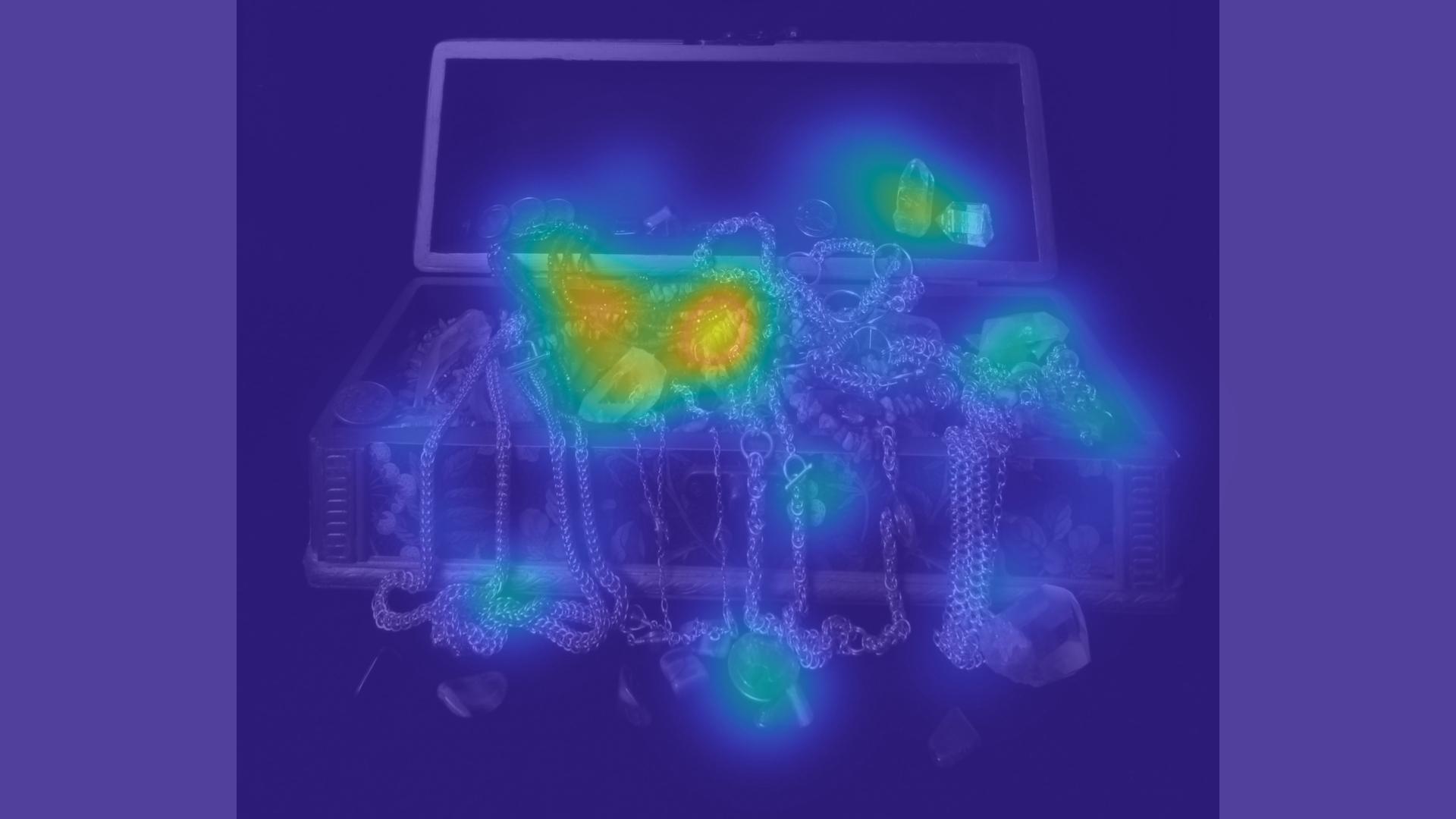}
  \end{subfigure}\hfil

  \vspace{0.4 mm} 
  \hrule %\noindent\hrulefill \\
  \vspace{0.4 mm} 
  \rotatebox[origin = c]{90}{(f)}
\rotatebox[origin = c]{90}{Treasure}
\rotatebox[origin = c]{90}{F2B}
  \begin{subfigure}{0.1555\linewidth}
    \centering
    \includegraphics[width=\linewidth]{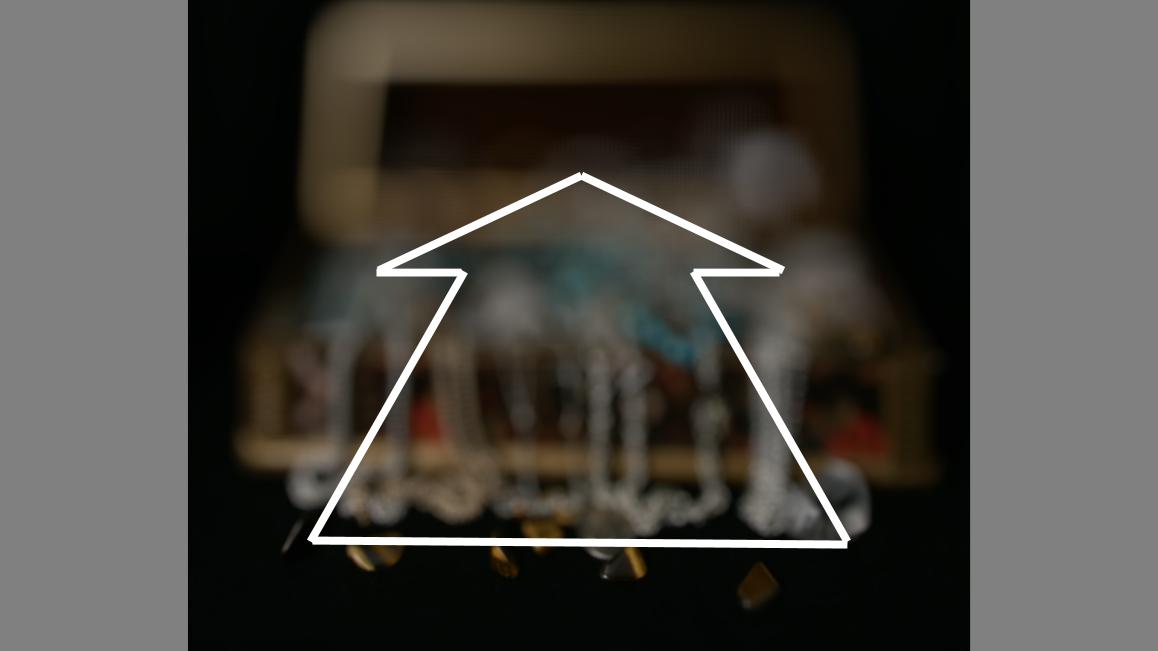}
  \end{subfigure}\hfil
  \begin{subfigure}{0.1555\linewidth}
    \centering
    \includegraphics[width=\linewidth]{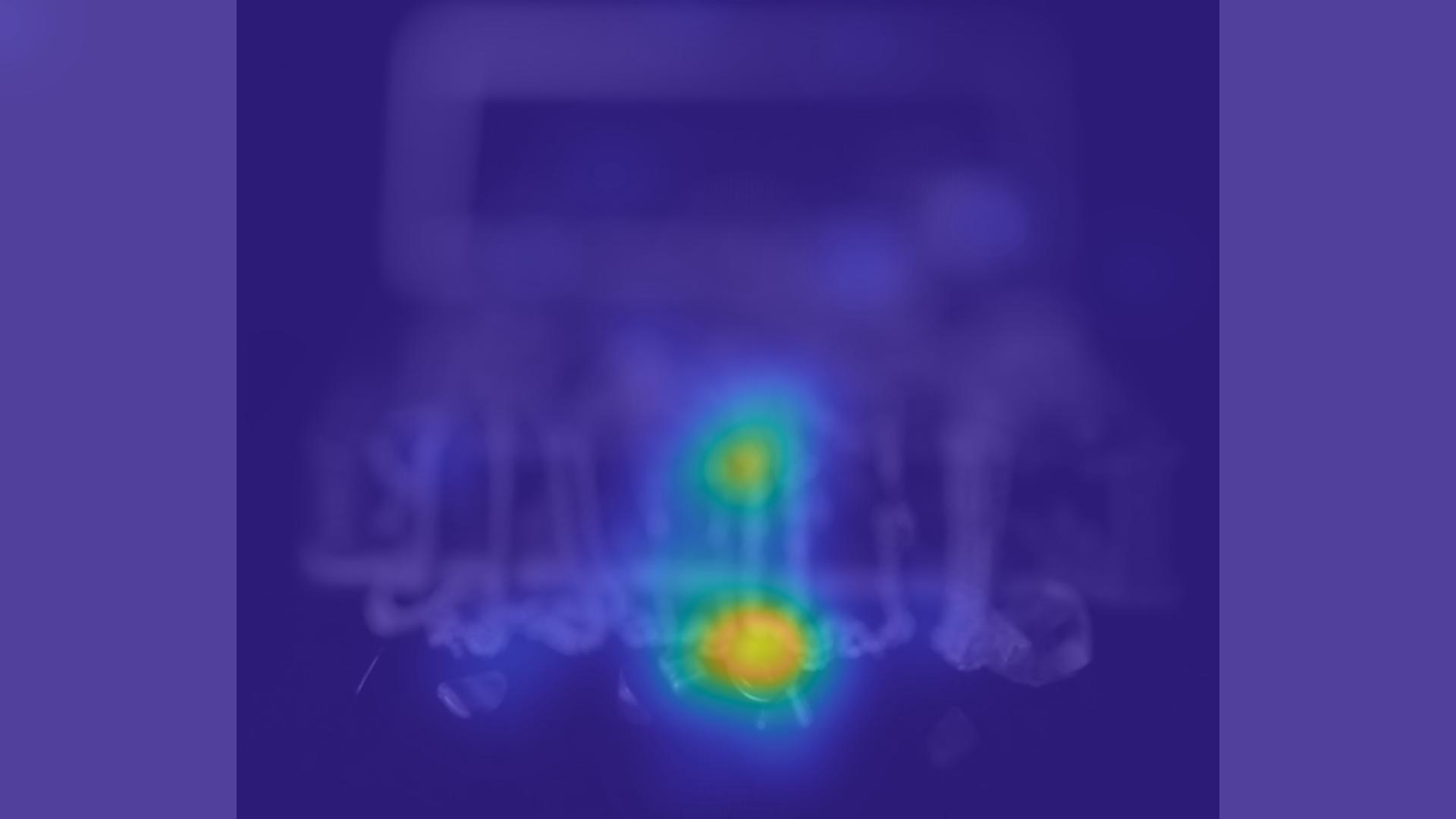}
  \end{subfigure}\hfil
  \begin{subfigure}{0.1555\linewidth}
    \centering
    \includegraphics[width=\linewidth]{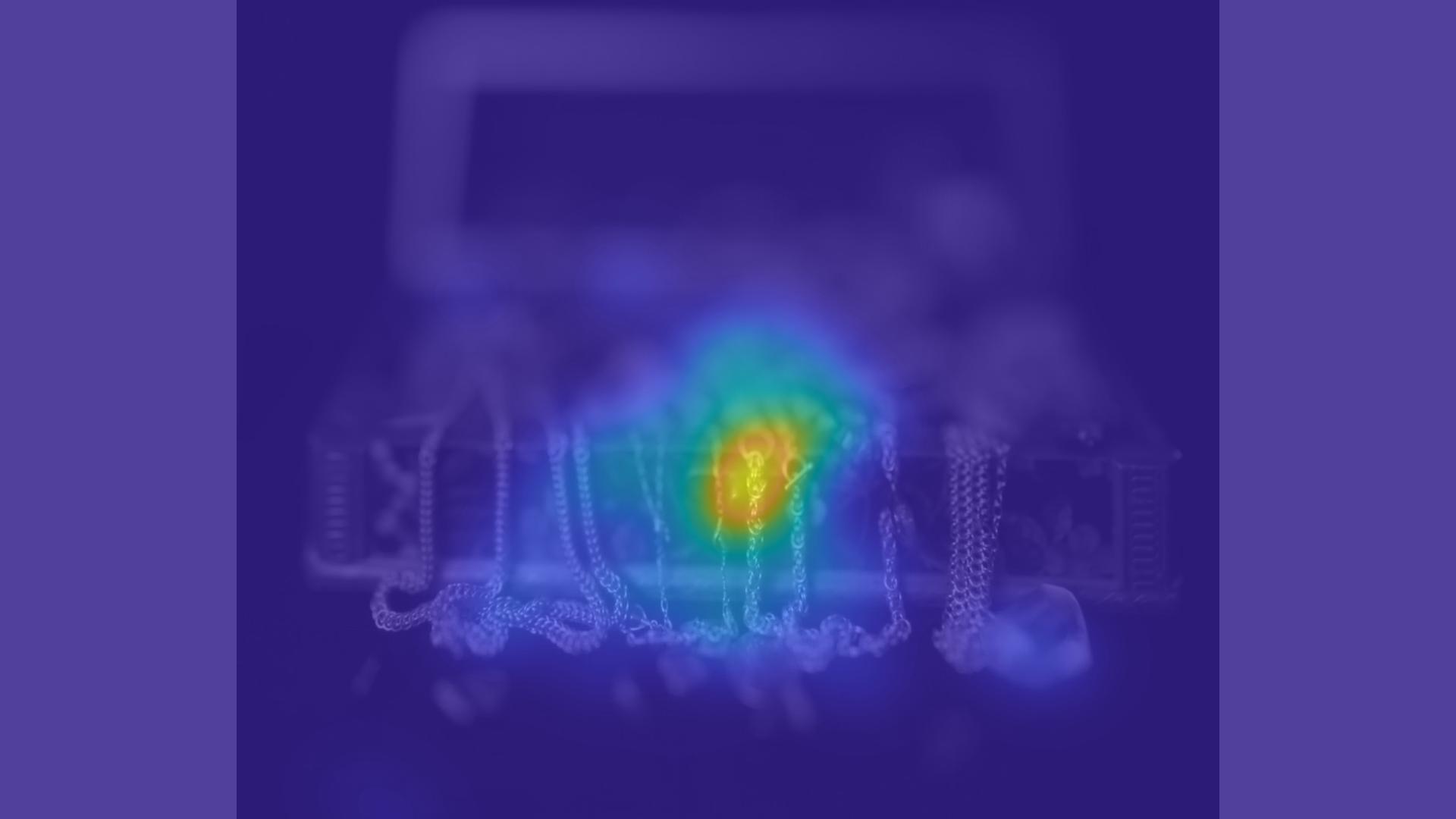}
  \end{subfigure}\hfil
  \begin{subfigure}{0.1555\linewidth}
    \centering
    \includegraphics[width=\linewidth]{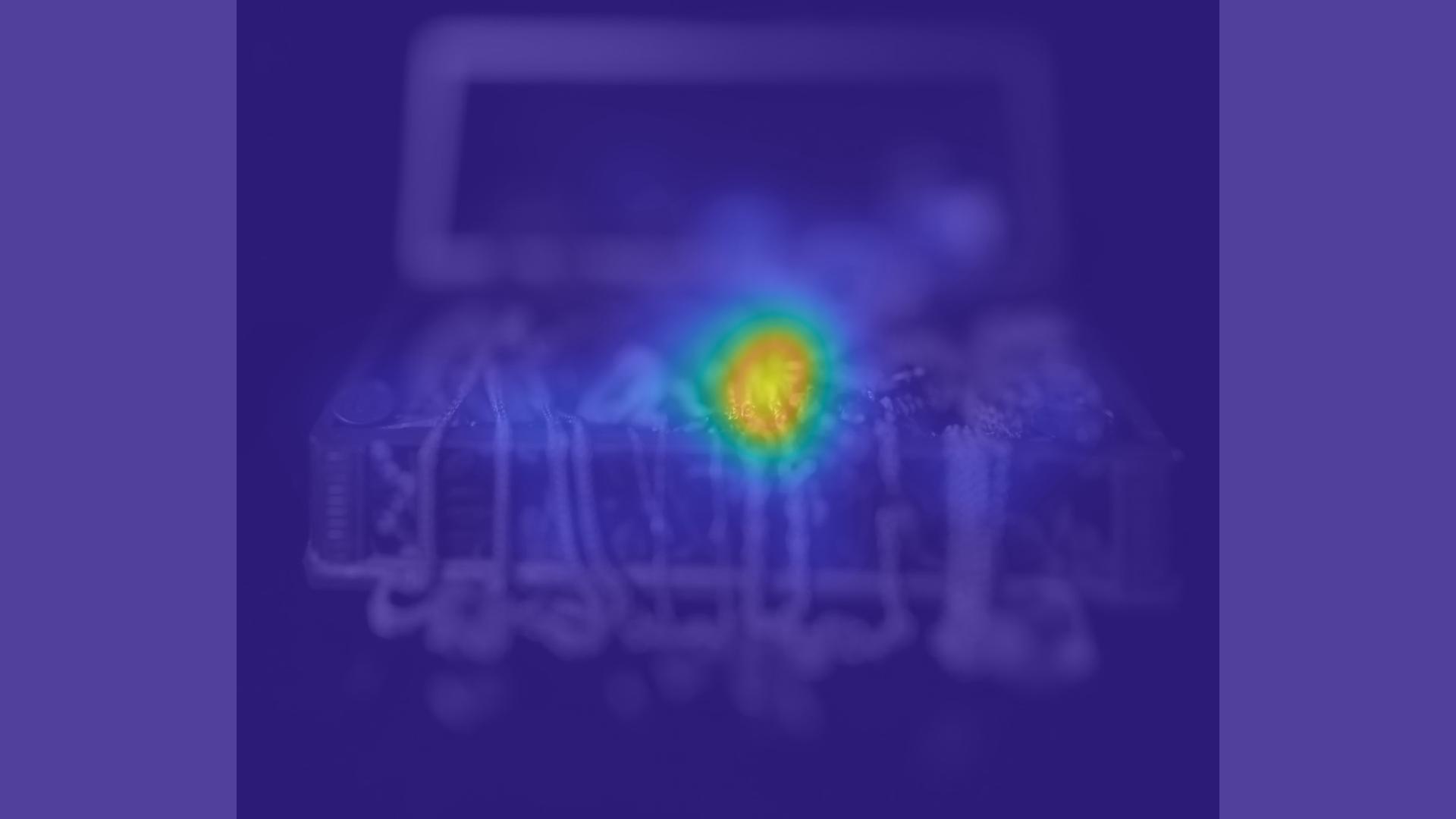}
  \end{subfigure}\hfil
  \begin{subfigure}{0.1555\linewidth}
    \centering
    \includegraphics[width=\linewidth]{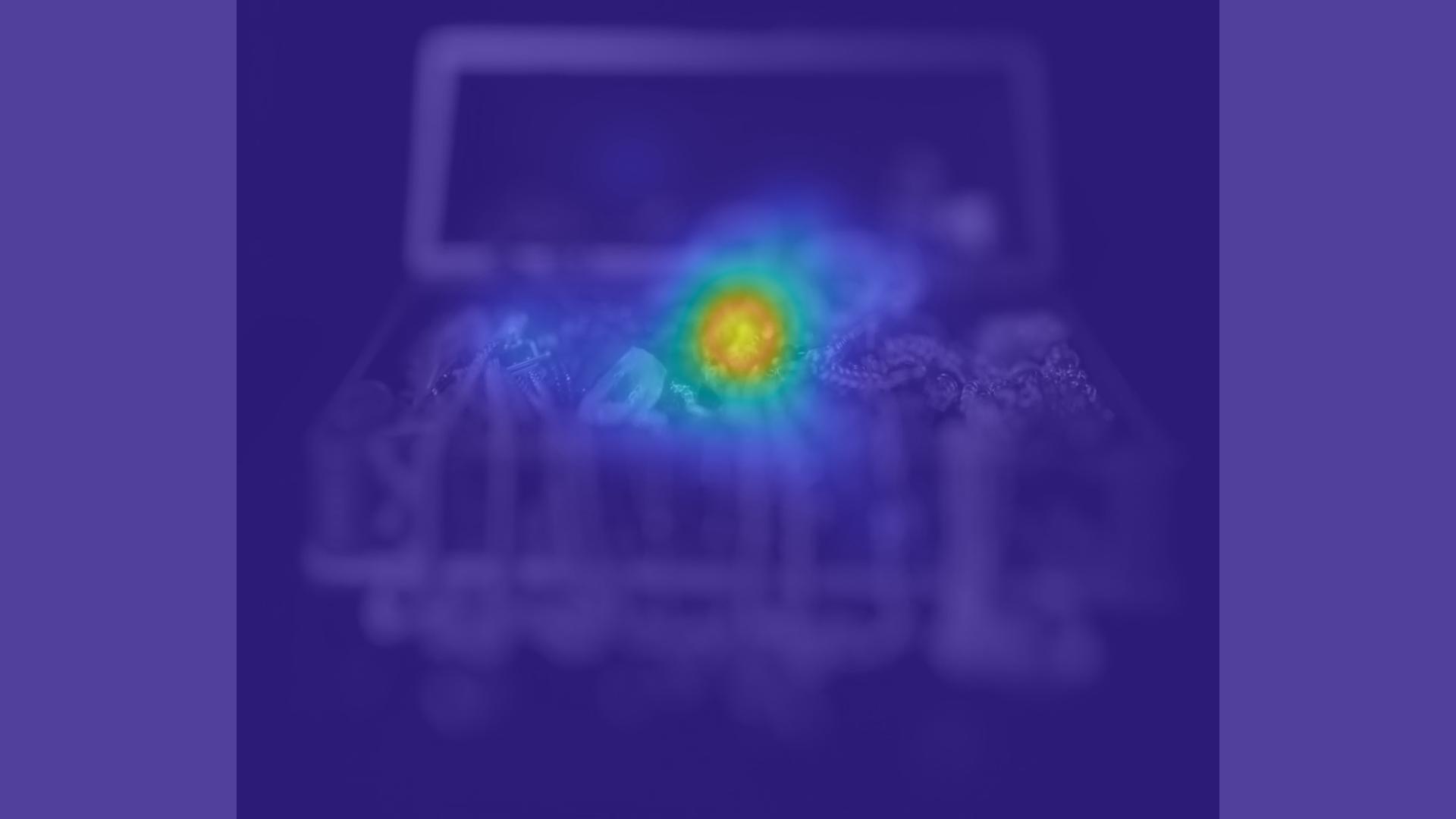}
  \end{subfigure}\hfil
  \begin{subfigure}{0.1555\linewidth}
    \centering
    \includegraphics[width=\linewidth]{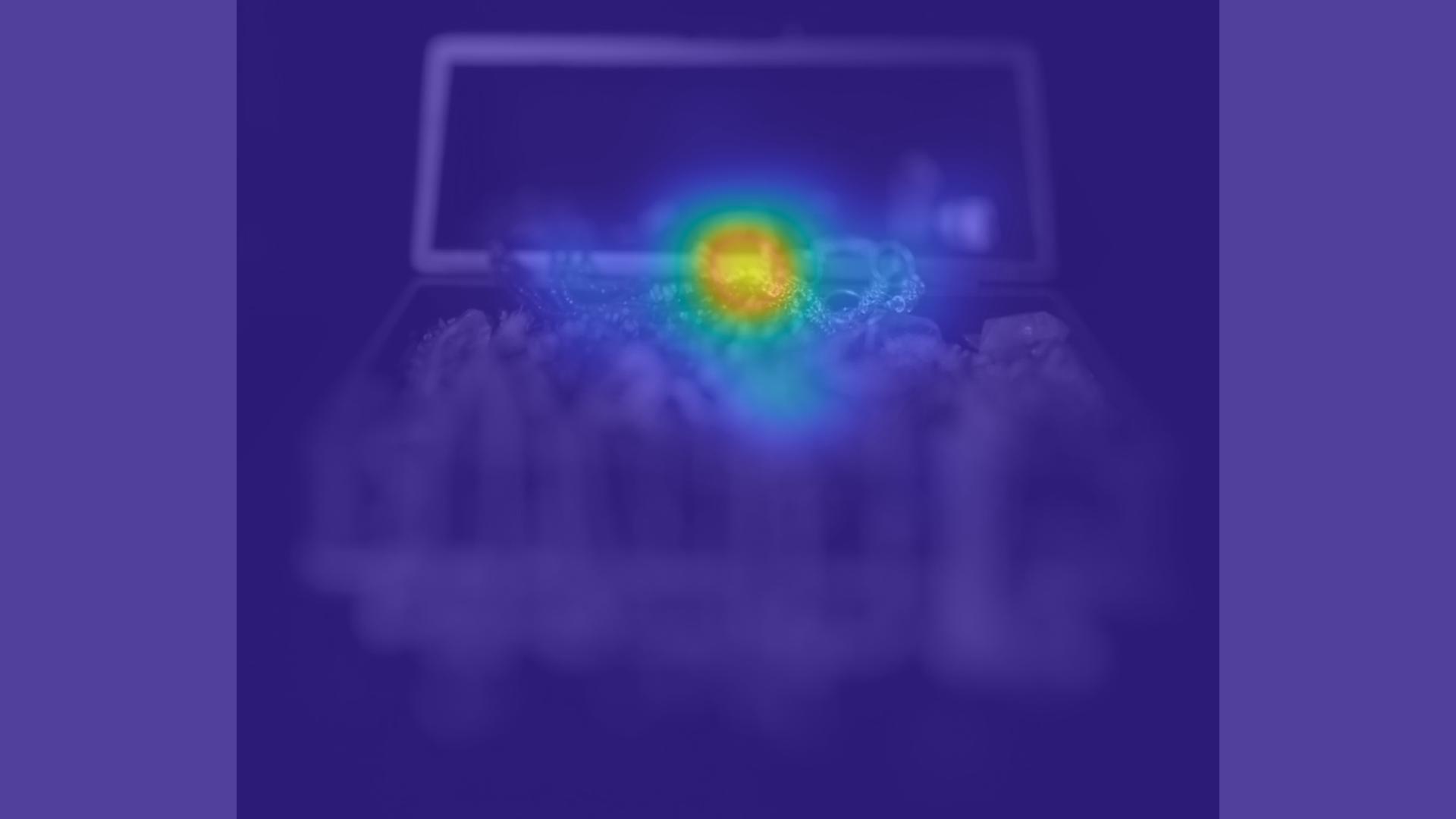}
  \end{subfigure}\hfil

  \vspace{0.3 mm} 
  \hrule %\noindent\hrulefill \\
  \rotatebox[origin = c]{90}{(g)}
\rotatebox[origin = c]{90}{Couch}
\rotatebox[origin = c]{90}{AiF}
  \begin{subfigure}{0.1555\linewidth}
    \centering
    \includegraphics[width=\linewidth]{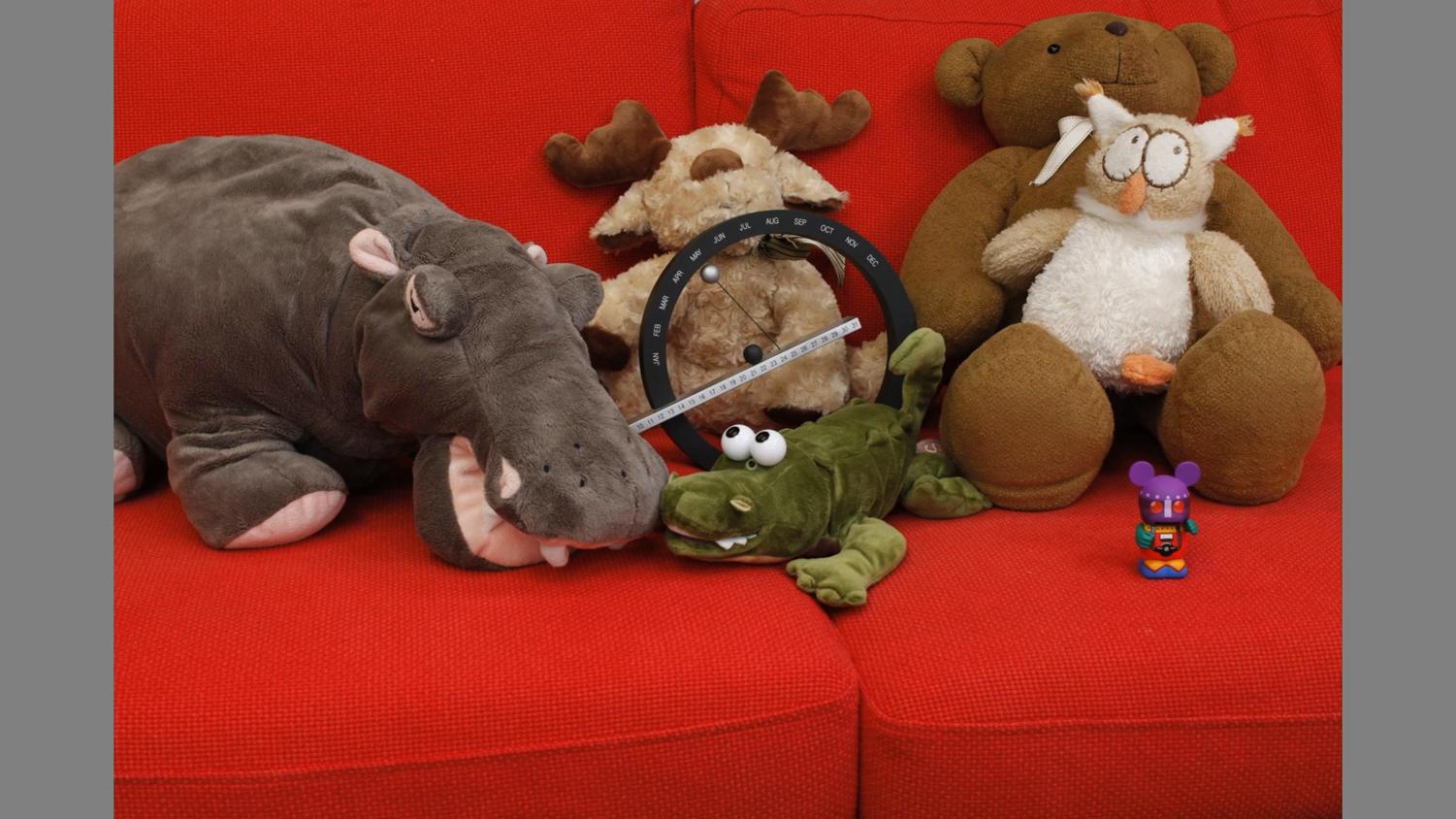}
  \end{subfigure}\hfil
  \begin{subfigure}{0.1555\linewidth}
    \centering
    \includegraphics[width=\linewidth]{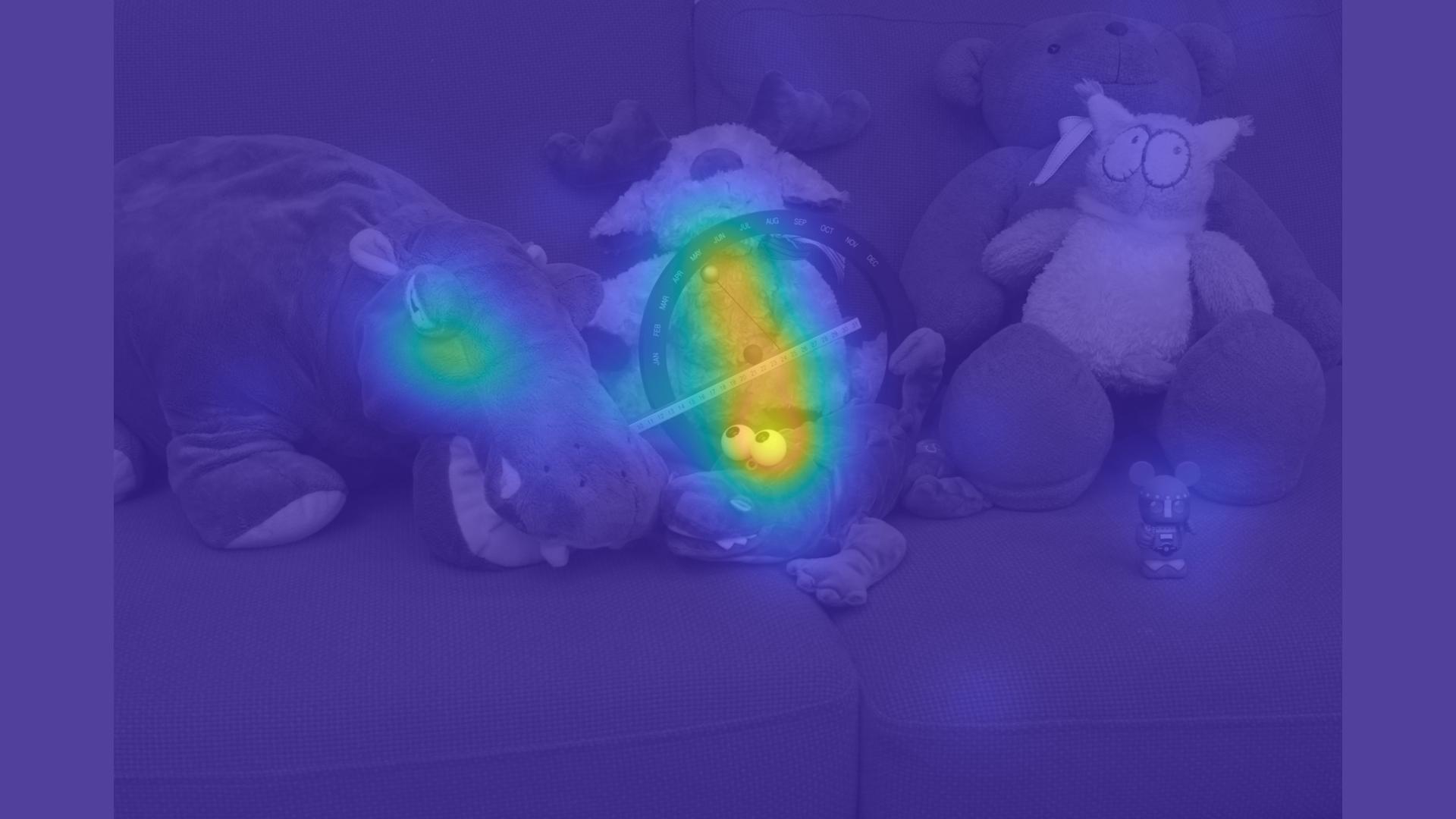}
  \end{subfigure}\hfil
  \begin{subfigure}{0.1555\linewidth}
    \centering
    \includegraphics[width=\linewidth]{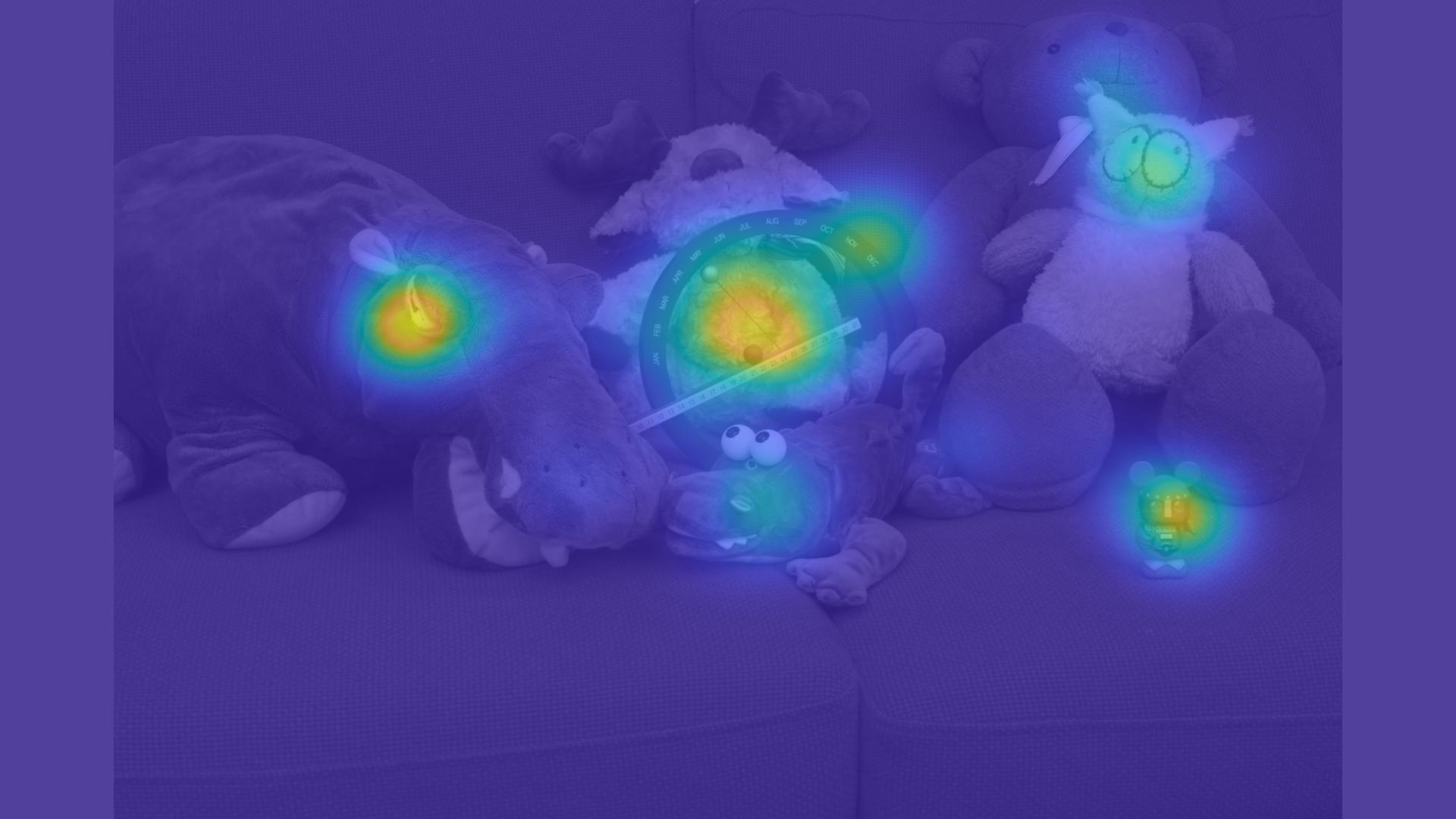}
  \end{subfigure}\hfil
  \begin{subfigure}{0.1555\linewidth}
    \centering
    \includegraphics[width=\linewidth]{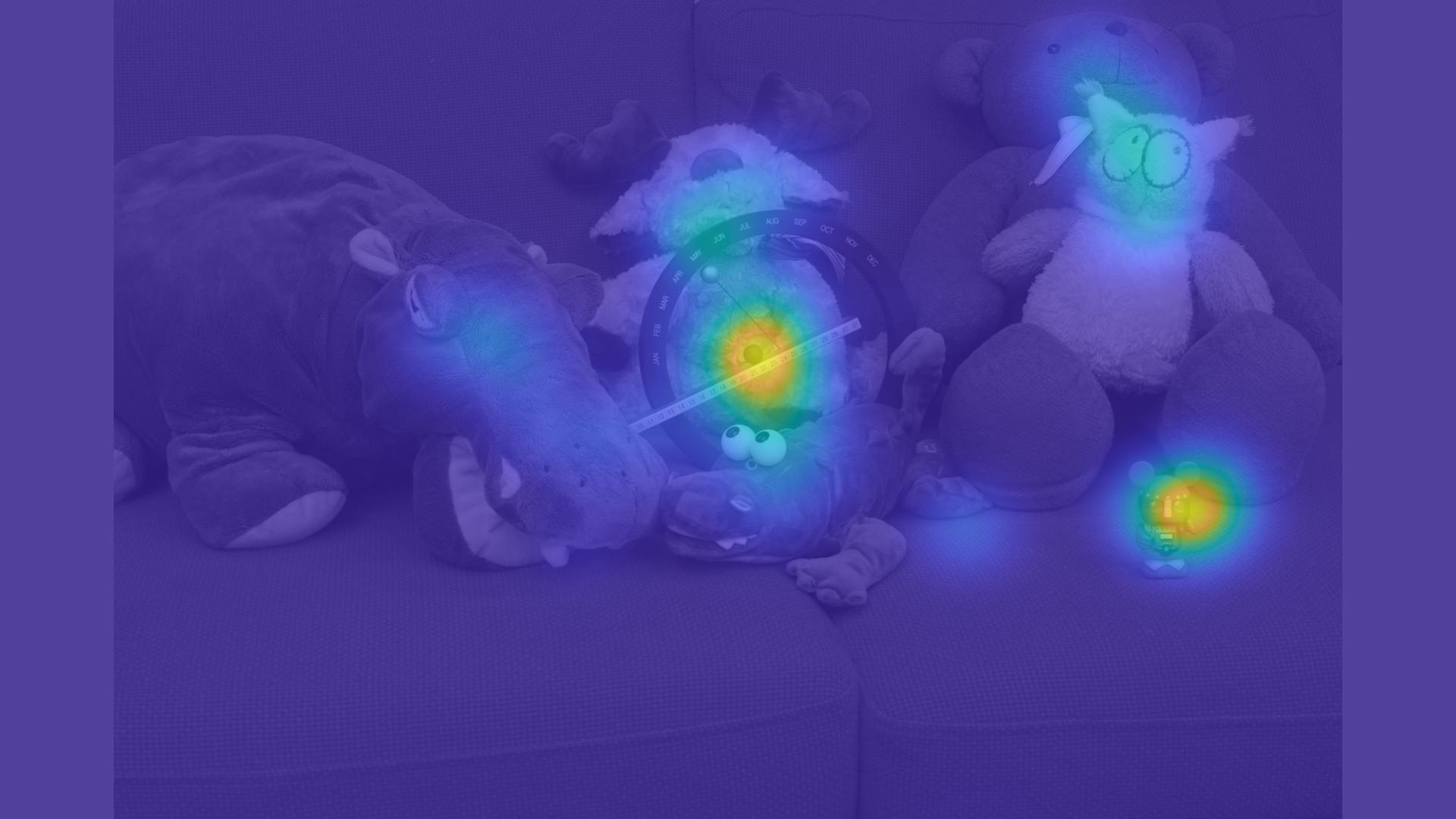}
  \end{subfigure}\hfil
  \begin{subfigure}{0.1555\linewidth}
    \centering
    \includegraphics[width=\linewidth]{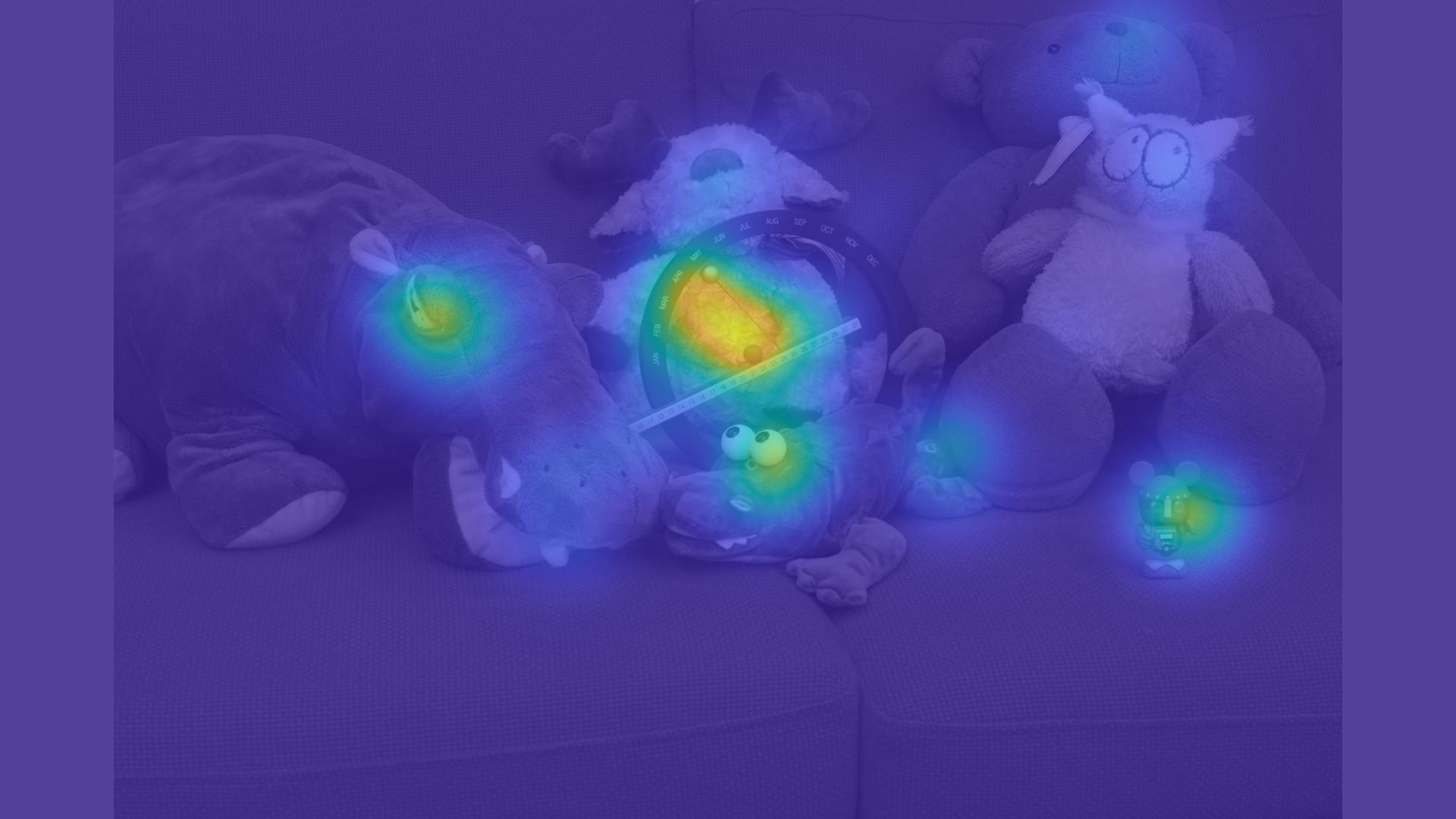}
  \end{subfigure}\hfil
  \begin{subfigure}{0.1555\linewidth}
    \centering
    \includegraphics[width=\linewidth]{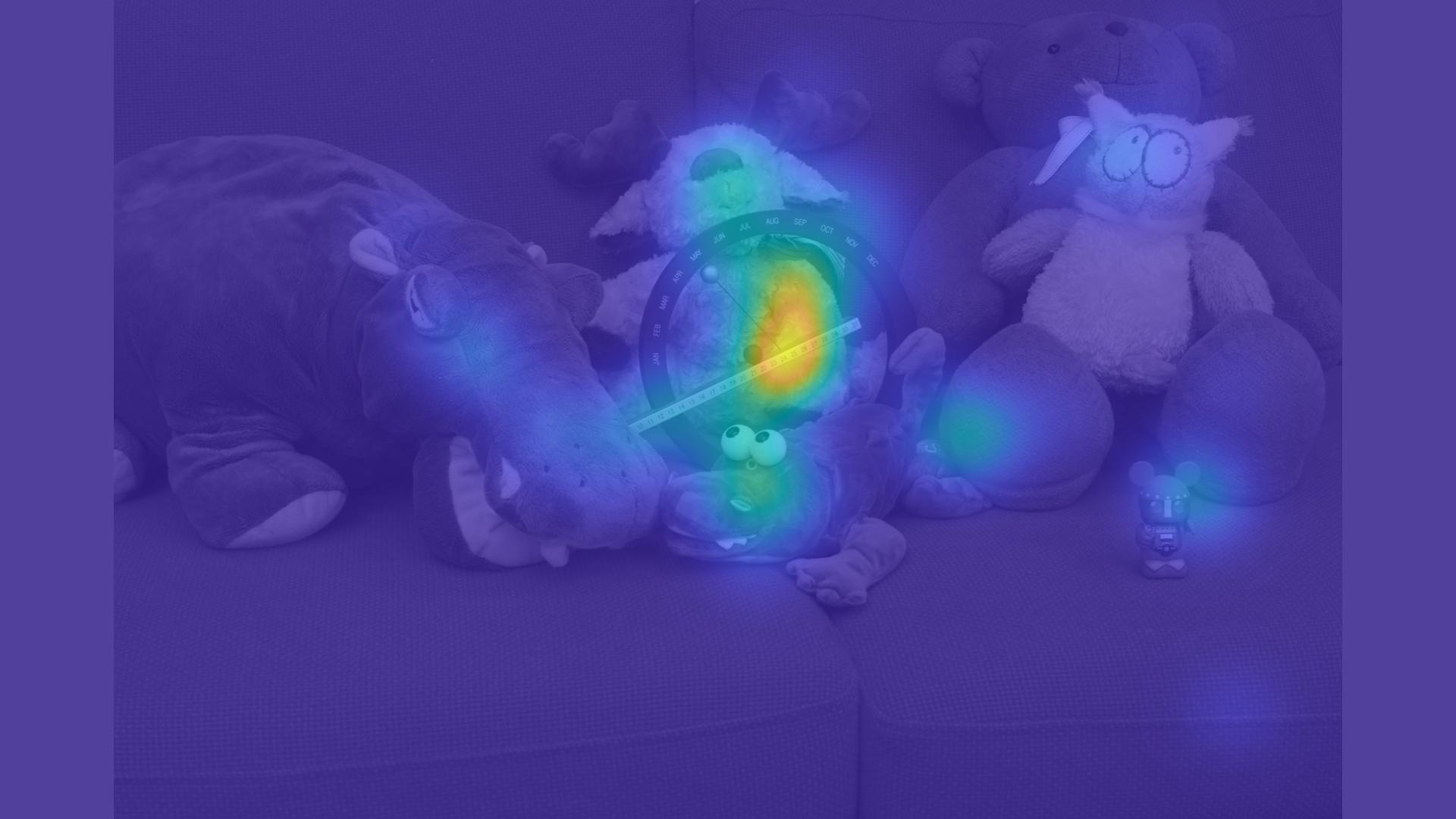}
  \end{subfigure}\hfil

  \vspace{0.3 mm} 
  \rotatebox[origin = c]{90}{(h)}
\rotatebox[origin = c]{90}{\hspace*{3 mm}Couch}
\rotatebox[origin = c]{90}{\hspace*{3 mm}F2B}
  \begin{subfigure}{0.1555\linewidth}
    \centering
    \includegraphics[width=\linewidth]{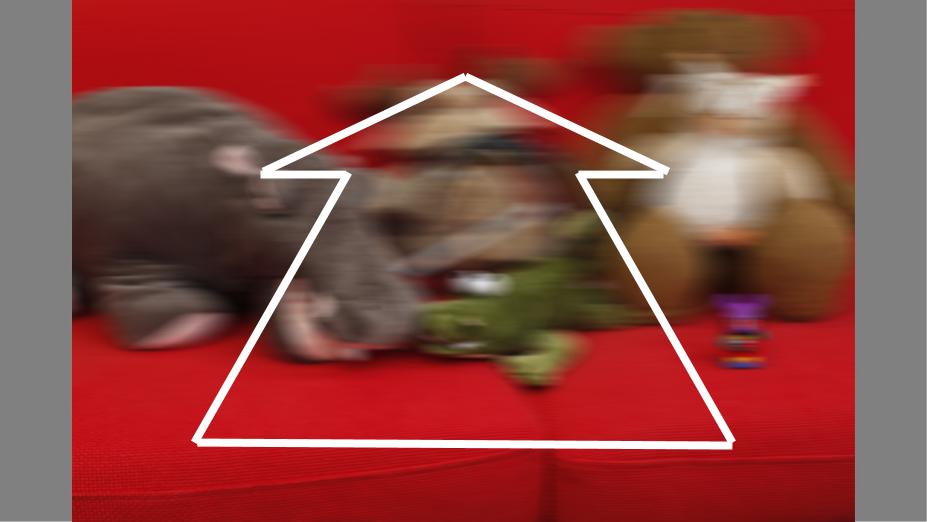}
    \caption*{Reference}
  \end{subfigure}\hfil
  \begin{subfigure}{0.1555\linewidth}
    \centering
    \includegraphics[width=\linewidth]{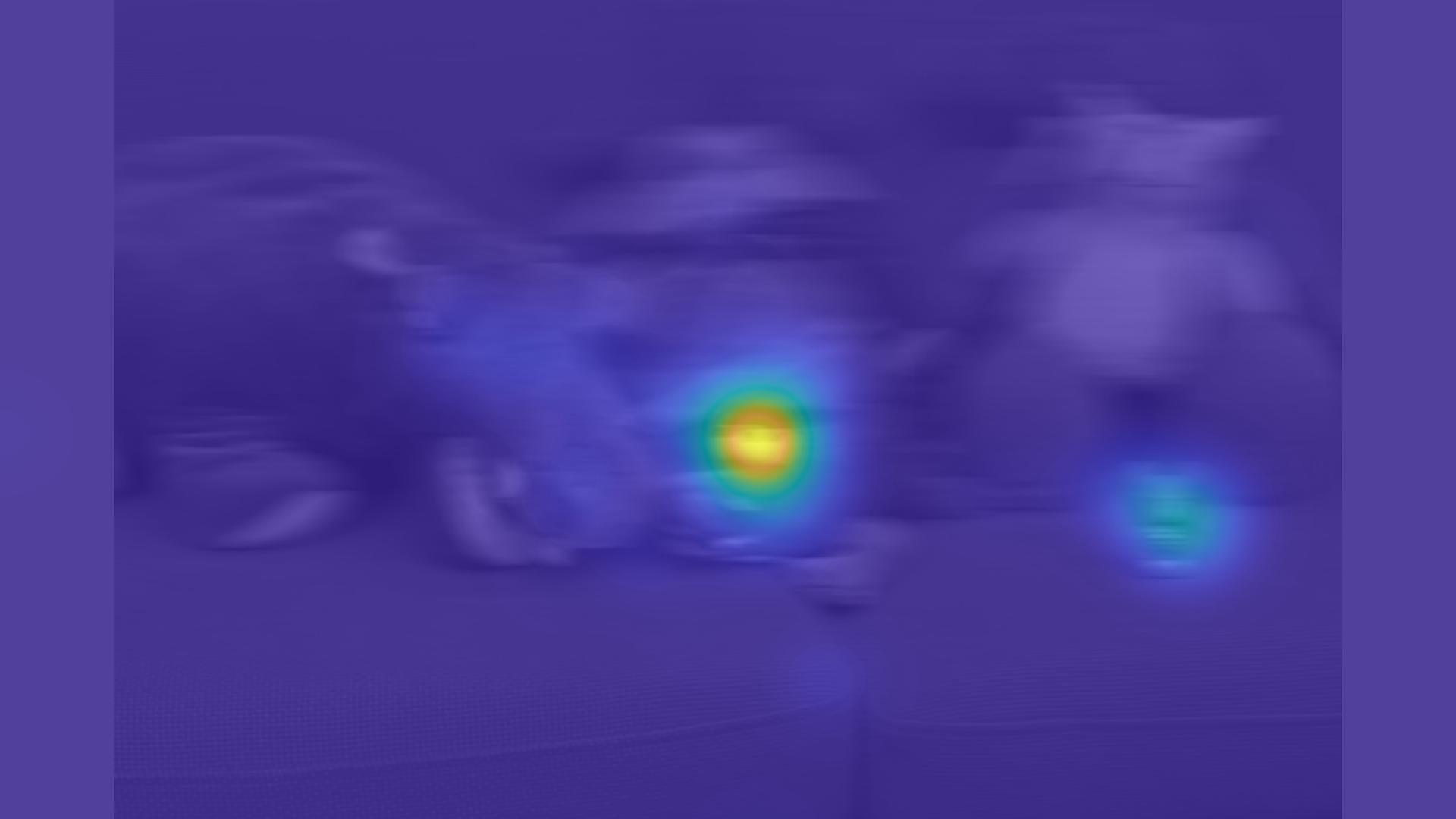}
    \caption*{Segment 1}
  \end{subfigure}\hfil
  \begin{subfigure}{0.1555\linewidth}
    \centering
    \includegraphics[width=\linewidth]{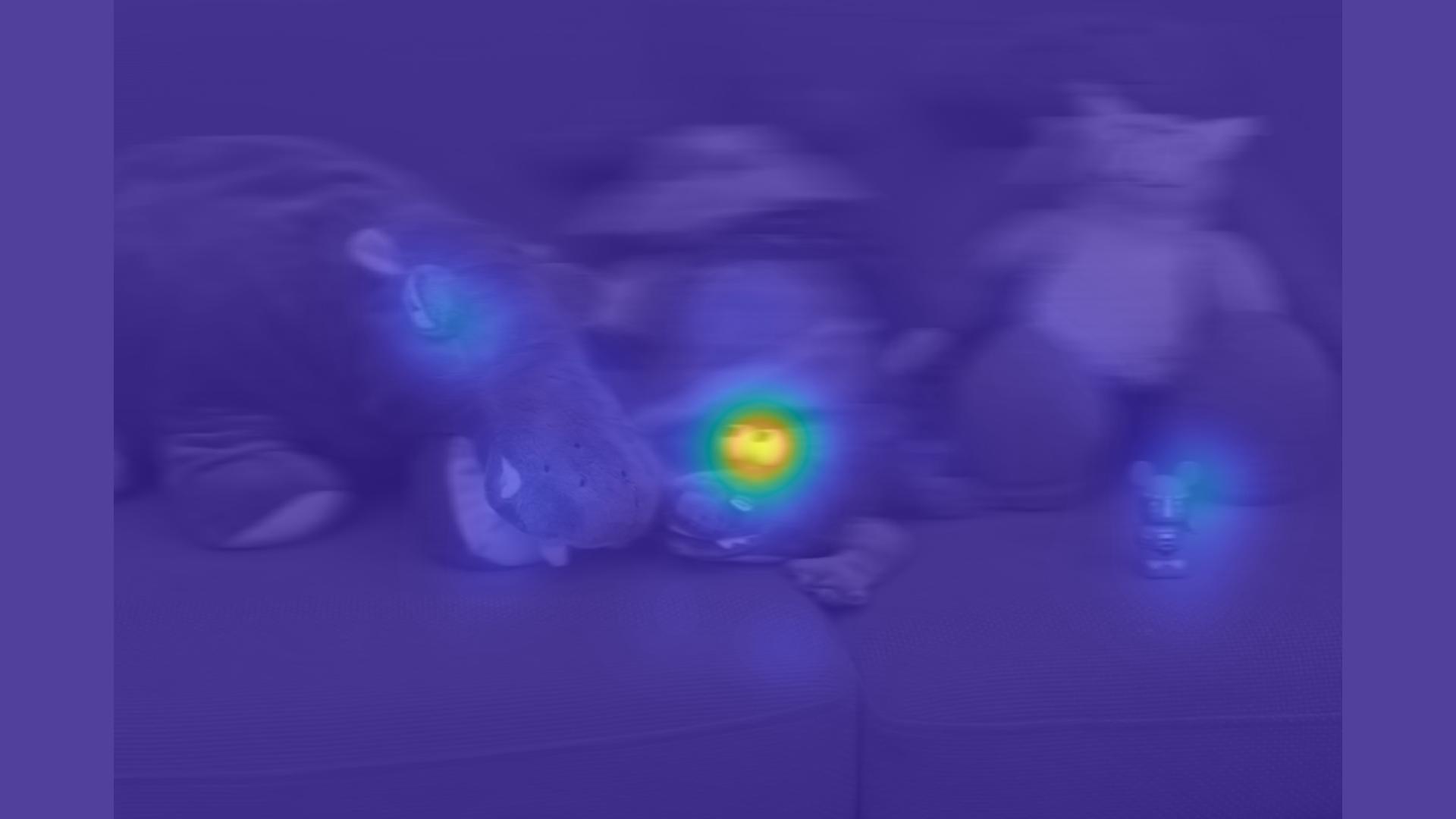}
    \caption*{Segment 2}
  \end{subfigure}\hfil
  \begin{subfigure}{0.1555\linewidth}
    \centering
    \includegraphics[width=\linewidth]{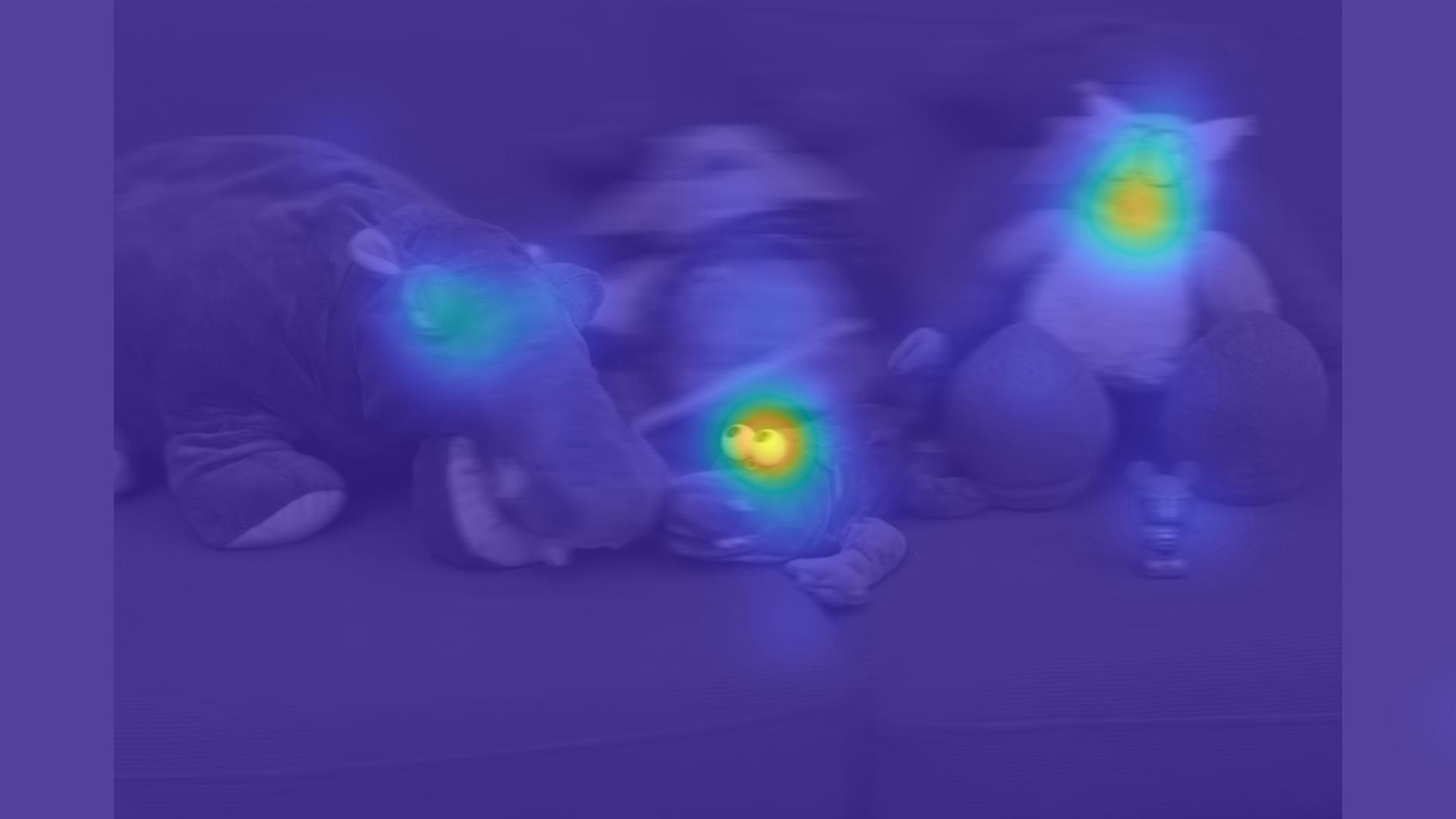}
    \caption*{Segment 3}
  \end{subfigure}\hfil
  \begin{subfigure}{0.1555\linewidth}
    \centering
    \includegraphics[width=\linewidth]{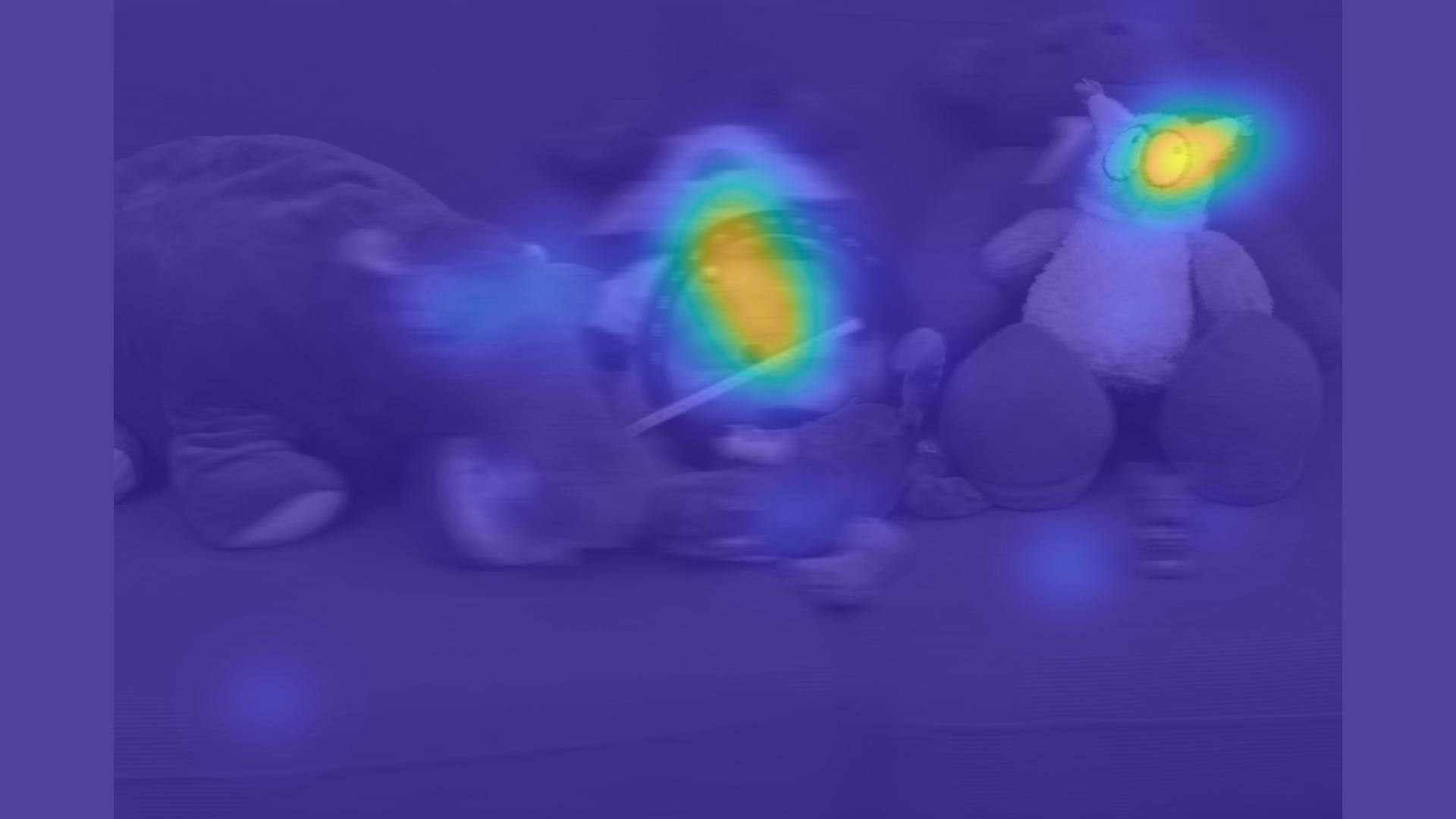}
    \caption*{Segment 4}
  \end{subfigure}\hfil
  \begin{subfigure}{0.1555\linewidth}
    \centering
    \includegraphics[width=\linewidth]{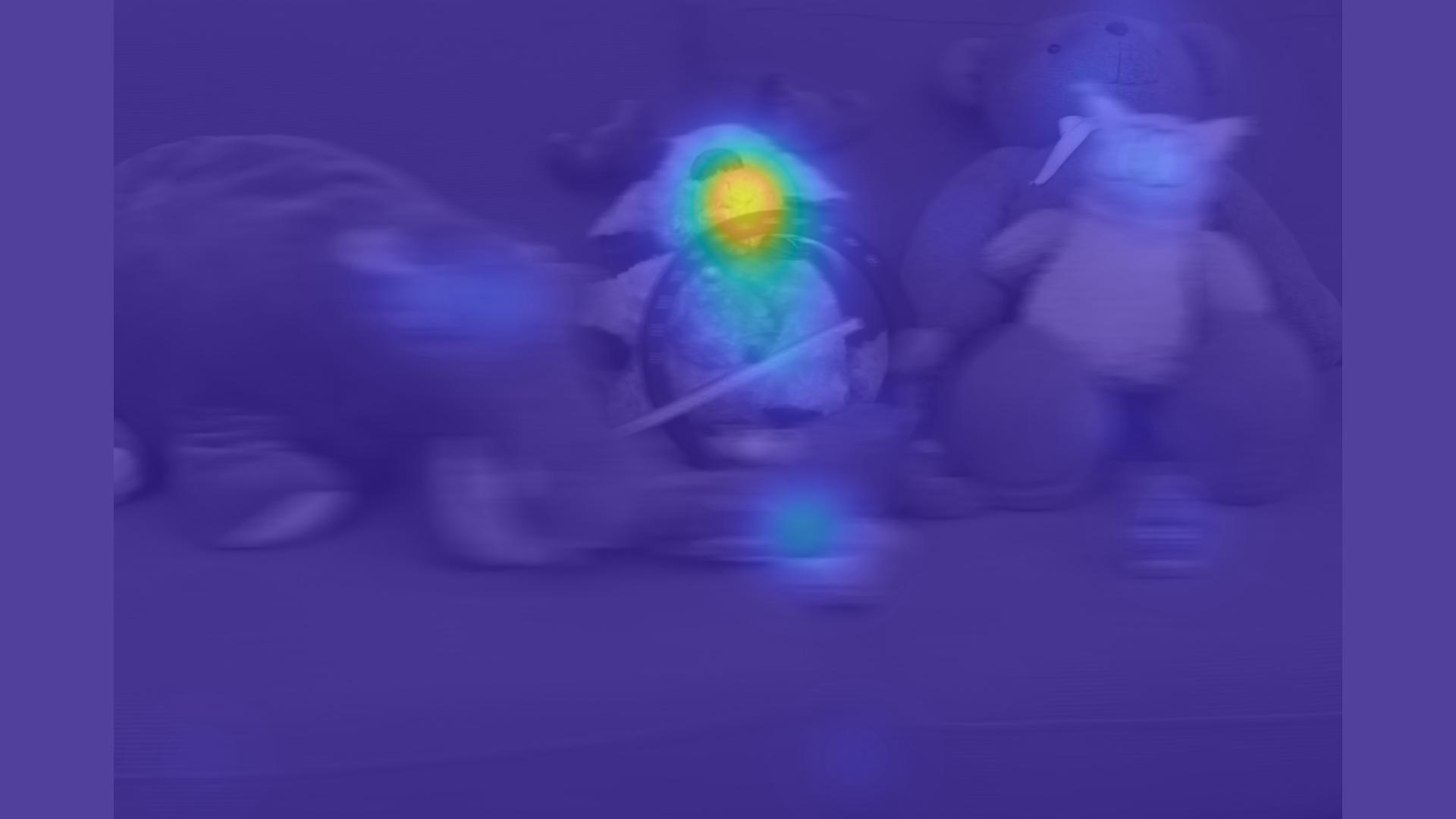}
    \caption*{Segment 5}
  \end{subfigure}\hfil

  \caption{Light Field renderings overlayed with a heatmap generated from all participants and split over time into five 2-second segments. All-in-focus, region-in-focus and front-to-back renderings labelled \textit{AiF}, \textit{RiF}, \textit{F2B}}
    \vspace{-5pt}

  \label{fig:ChunkExamples}
\end{figure*}

Moreover,
to understand what causes participants to fixate on regions of focus and whether or not this is always the case, we compared the segments over time of static data (all-in-focus and region-in-focus renderings) to those of focally-varying data (focal-sweeps) for each light field. We found that gaze is held on objects that are in focus when they are also salient in the all-in-focus rendering. For example, the basketball and centre shoe in the Sideboard light field shown in Fig.~\ref{fig:ChunkExamples} (a)  and (b).

There are other cases where there are objects in a scene that have a higher level of saliency in the all-in-focus rendering and they pull the viewers attention away from the region of focus in other renderings. For example, observe the region-in-focus rendering of the Tarot-S light field in Fig.~\ref{fig:ChunkExamples} (c). Although the focus is in the foreground, the saliency map is also concentrated on the centre ball. As the centre ball has a high saliency in the all-in-focus rendering Fig.~\ref{fig:ChunkExamples} (d), we can deduce that it is salient independent of its level of focus.

Some scenes do not depict a specific object/ exhibit a region of high saliency. These tend to produce highly dispersed saliency maps. This is demonstrated in the Treasure light field in Fig.~\ref{fig:ChunkExamples} (e). The saliency dispersion in the all-in-focus rendering suggests that the viewer is likely to follow the region of focus almost exclusively in other renderings as seen in Fig.~\ref{fig:ChunkExamples} (f). 

This trend of following gaze is also evident when there are a few objects with similar levels of saliency in the all-in-focus rendering. This is seen in the animal heads and the central object in the Couch light field  Fig.~\ref{fig:ChunkExamples} (g). In the focal-sweep Fig.~\ref{fig:ChunkExamples} (h), we can see the viewers gaze following the path of focus as above but not as smoothly, rather jumping between the salient objects that are in focus in each segment.

\subsection{Quantitative Analysis}

%\temp{[Entropy results]}
In this subsection, we provide a quantitative analysis to understand the relationship between human visual perception and light field rendering. %Even though the qualitative analysis above shows that the participants mostly agree on the parts of the video to look at, we provide the readers with quantitative results.

\begin{wraptable}{r}{8cm}
\vspace{-3pt}
%\vspace{-15pt}
%\centering
\caption{Entropy analysis results. The five cases are given in the columns: All-in-focus (\textit{AiF}), region-in-focus \#1 (\textit{RiF1}), region-in-focus \#2 (\textit{RiF2}), back-to-front focal-sweep (\textit{B2F}), and front-to-back focal-sweep (\textit{F2B}).}
\vspace{-5pt}
\begin{tabular}{l|ccccc} \hline \hline %\noalign{\smalllskip}
\textbf{Light fields}
              & \textbf{AiF}   
                      & \textbf{RiF1}  
                              & \textbf{RiF2}  
                                      & \textbf{B2F}   
                                              & \textbf{F2B} \\ \hline%\noalign{\smalllskip} \hline \noalign{\smalllskip}
Boardgames    & 4.06  & 3.87  & 3.81  & 3.96  & 3.82       \\
Church        & 4.12  & 3.46  & 3.64  & 3.48  & 3.41       \\
Couch         & 4.27  & 4.04  & 3.82  & 3.64  & 3.52       \\
Dino          & 3.92  & 3.82  & 3.61  & 3.82  & 3.67       \\
Dishes        & 4.19  & 4.06  & 3.26  & 3.48  & 3.50       \\
Friends       & 4.55  & 4.49  & 4.45  & 4.25  & 4.45       \\
LegoKnights   & 4.02  & 3.45  & 3.62  & 3.27  & 3.34       \\
Mansion       & 4.12  & 4.04  & 3.67  & 3.48  & 3.24       \\
Medieval      & 4.02  & 3.71  & 3.83  & 3.65  & 3.76       \\
Pens          & 3.91  & 3.68  & 3.79  & 3.84  & 3.80       \\
Platonic      & 3.79  & 3.40  & 3.14  & 3.53  & 3.55       \\
Sideboard     & 3.79  & 3.75  & 3.81  & 3.98  & 3.90       \\
Table         & 3.85  & 3.47  & 3.85  & 3.76  & 3.65       \\
Tarot-L   & 4.92  & 3.46  & 4.39  & 3.44  & 3.73       \\
Tarot-S   & 4.62  & 3.85  & 3.80  & 3.91  & 4.16       \\
Tower         & 3.77  & 3.72  & 3.84  & 3.63  & 3.42       \\
Town          & 4.02  & 4.03  & 4.06  & 4.02  & 4.02       \\
Treasure      & 4.01  & 3.47  & 3.66  & 3.54  & 3.46       \\
Vespa         & 3.81  & 3.84  & 3.78  & 3.45  & 3.82       \\
Vinyl         & 4.18  & 3.60  & 3.87  & 4.16  & 3.77       \\ \hline %\noalign{\smalllskip} \hline \noalign{\smalllskip}
Average       & 4.10  & 3.76  & 3.78  & 3.71  & 3.70       \\ \hline \hline %\noalign{\smalllskip} \hline \hline
\end{tabular}
\label{tab:entropy}
\vspace{-10pt}
\end{wraptable}

We infer from our qualitative analysis that viewers are likely to fixate on and follow regions in focus and that fixations of all-in-focus renderings were more dispersed. To further examine these observations, we used the  calculated entropy of fixations recorded by the eye tracker to determine if participants' fixations varied more or less for each rendering per light field.
%For this purpose, we only use fixations, which are extracted by the EyeLink 1000 Plus Core System using the velocity and the acceleration of the gaze information.

To calculate entropy, we first created a \textit{fixation map} with the same spatial resolution as the stimuli (i.e., $1920 \times 1080$), %. %and was initialised with zeroes. 
%The fixation map was then 
which was populated with the fixations from all the users for a specific case, using Eqn. 1: % the following equation:
\begin{equation}
      F_C(i,j) =
      \begin{cases}
                                       1 & \text{if there is a fixation at } I(i,j) \\
                                       0 & \text{otherwise} 
      \end{cases}
\end{equation}
where $F$ is the fixation map, $i$ and $j$ are the row and column pixels, $I$ is the image stimulus, and $C \in \{\text{AiF}, \text{RiF1}, \text{RiF2}, \text{B2F}, \text{F2B}\}$ denotes the rendering. This yields a fixation map which is sparse. The entropy values were generated for each of the five cases and reported in Table~\ref{tab:entropy} using Matlab's \texttt{entropy(}$F_C$\texttt{)} function, computing the probability of 1s occuring in $F_C$.

The results show that all-in-focus renderings have higher entropy values on average compared to region-in-focus and focal-sweep ones, which suggests participants were more focused on average in the \textit{focal-sweep} and \textit{region-in-focus} cases, compared to the \textit{all-in-focus} case. A two-tailed t-test confirmed that \textit{all-in-focus} was significantly different than others ($\alpha = 0.05$). The differences among the other cases were not statistically significant.

% ------------- Can be deleted -----
%We conducted a two-tailed t-test which yielded  \textit{all-in-focus} was significantly different than others ($\alpha = 0.05$). The differences among the other cases were not statistically significant.}

%A two-tailed t-test found only \textit{all-in-focus} was significantly different than others with 95~\% confidence, but the numbers show} %This suggests -----> the calculations state
%that participants were on average} %collectively 
%more focused in the \textit{focal-sweep} cases than in \textit{region-in-focus} cases and \textit{all-in-focus} cases where the entropy values indicate participants were looking around more randomly. 

% The participants were more focused in the \textit{focal-sweep} and \textit{region-in-focus} cases, compared to the \textit{all-in-focus} cases. This is also confirmed with a two-tailed t-test which yielded that only \textit{all-in-focus} was significantly different than others with 95~\% confidence.
% ------------- Can be deleted -----

%[Find objects with disparity values and compare heatmaps to those]

%\temp{[Here we present entropy of PCC KLD with the heatmaps]}

%====================
%       DISCUSSION
%====================
\section{Conclusion}
\label{sec:conclusion}
In this paper, we outline our investigation into how attention is affected by changes in focus to verify whether characteristics specific to light fields influence visual attention. From analysis of the scanpaths of light fields, we conclude that there is a difference in visual attention of static renderings of light fields when compared to focally-varying renderings. This was reinforced by examining the saliency maps of the different rendering types. We found that visual attention was often guided by focus and objects/regions at the focal plane by observing saliency maps computed on segments of the renderings over time.

The ground truth data of previous light field saliency works are segmented objects of all-in-focus renderings. We found that these do not fully capture the visual attention of light fields. Salient information not present in the all-in-focus planar rendering is often revealed by saliency maps of focal-sweep renderings. It is also apparent in saliency maps that viewers did not only fixate on objects. Furthermore, the saliency maps of segmented all-in-focus renderings were more dispersed than those of other renderings. This observation was supported by analysis of the entropy of the fixation data for each rendering where we found that all-in-focus data had highest entropy which suggested greater randomness in the data. This variation in the visual attention of different renderings shows the limitations in the use of a saliency map of only one rendering type as ground truth.

We plan to use our saliency maps as ground truth data in future work for eye fixation prediction for light fields. As they depict the likelihood of eye fixation at every point of a light field capture they have applications in this field among others such as light field rendering and compression.

%\vfill
%\pagebreak


\bibliographystyle{apalike}

\bibliography{strings,imvip}

\begin{thebibliography}{}

\bibitem[{Alain} et~al., 2019]{lfstreaming}
{Alain}, M., {Ozcinar}, C., and {Smolic}, A. (2019).
\newblock A study of light field streaming for an interactive refocusing
  application.
\newblock In {\em 2019 IEEE International Conference on Image Processing
  (ICIP)}, pages 3761--3765.

\bibitem[Broxton et~al., 2019]{broxton2019low}
Broxton, M., Busch, J., Dourgarian, J., DuVall, M., Erickson, D., Evangelakos,
  D., Flynn, J., Overbeck, R., Whalen, M., and Debevec, P. (2019).
\newblock A low cost multi-camera array for panoramic light field video
  capture.
\newblock In {\em SIGGRAPH Asia 2019 Posters}, SA ’19, New York, NY, USA.
  Association for Computing Machinery.

\bibitem[Buswell, 1935]{buswell1935people}
Buswell, G.~T. (1935).
\newblock {\em How people look at pictures: a study of the psychology and
  perception in art.}
\newblock Univ. Chicago Press.

\bibitem[Gortler et~al., 1996]{Gortler1996}
Gortler, S.~J., Grzeszczuk, R., Szeliski, R., and Cohen, M.~F. (1996).
\newblock The lumigraph.
\newblock In {\em Proceedings of the 23rd Annual Conference on Computer
  Graphics and Interactive Techniques}, SIGGRAPH ’96, page 43–54, New York,
  NY, USA. Association for Computing Machinery.

\bibitem[Greenwald, 1976]{greenwald1976within}
Greenwald, A.~G. (1976).
\newblock Within-subjects designs: To use or not to use?
\newblock {\em Psychological Bulletin}, 83(2):314.

\bibitem[Honauer et~al., 2016]{honauer2016dataset}
Honauer, K., Johannsen, O., Kondermann, D., and Goldluecke, B. (2016).
\newblock A dataset and evaluation methodology for depth estimation on {4D}
  light fields.
\newblock In {\em Asian Conference on Computer Vision}, pages 19--34. Springer.

\bibitem[Itti, 2005]{itti2005quantifying}
Itti, L. (2005).
\newblock Quantifying the contribution of low-level saliency to human eye
  movements in dynamic scenes.
\newblock {\em Visual Cognition}, 12(6):1093--1123.

\bibitem[{Itti} et~al., 1998]{itti1998model}
{Itti}, L., {Koch}, C., and {Niebur}, E. (1998).
\newblock A model of saliency-based visual attention for rapid scene analysis.
\newblock {\em IEEE Transactions on Pattern Analysis and Machine Intelligence},
  20(11):1254--1259.

\bibitem[Jones et~al., 2007]{jones2007rendering}
Jones, A., McDowall, I., Yamada, H., Bolas, M., and Debevec, P. (2007).
\newblock Rendering for an interactive 360 light field display.
\newblock {\em ACM Transactions on Graphics (TOG)}, 26(3):40.

\bibitem[Judd et~al., 2009]{Judd_2009}
Judd, T., Ehinger, K., Durand, F., and Torralba, A. (2009).
\newblock Learning to predict where humans look.
\newblock In {\em IEEE International Conference on Computer Vision (ICCV)}.

\bibitem[Kim et~al., 2013]{kim2013scene}
Kim, C., Zimmer, H., Pritch, Y., Sorkine-Hornung, A., and Gross, M.~H. (2013).
\newblock Scene reconstruction from high spatio-angular resolution light
  fields.
\newblock {\em ACM Trans. Graph.}, 32(4):73--1.

\bibitem[Kleiner et~al., 2007]{kleiner2007s}
Kleiner, M., Brainard, D., Pelli, D., Ingling, A., Murray, R., Broussard, C.,
  et~al. (2007).
\newblock What’s new in {P}sychtoolbox-3.
\newblock {\em Perception}, 36(S).

\bibitem[Koch and Ullman, 1987]{koch1987shifts}
Koch, C. and Ullman, S. (1987).
\newblock Shifts in selective visual attention: towards the underlying neural
  circuitry.
\newblock In {\em Matters of intelligence}, pages 115--141. Springer.

\bibitem[Lanman et~al., 2011]{lanman2011polarization}
Lanman, D., Wetzstein, G., Hirsch, M., Heidrich, W., and Raskar, R. (2011).
\newblock Polarization fields: Dynamic light field display using multi-layer
  lcds.
\newblock {\em ACM Trans. Graph.}, 30(6):1–10.

\bibitem[Le~Meur and Baccino, 2013]{le2013methods}
Le~Meur, O. and Baccino, T. (2013).
\newblock Methods for comparing scanpaths and saliency maps: strengths and
  weaknesses.
\newblock {\em Behavior research methods}, 45(1):251--266.

\bibitem[{Le Pendu} et~al., 2019]{lePendu2019fourier}
{Le Pendu}, M., {Guillemot}, C., and {Smolic}, A. (2019).
\newblock A fourier disparity layer representation for light fields.
\newblock {\em IEEE Transactions on Image Processing}, 28(11):5740--5753.

\bibitem[Lee et~al., 2016]{lee2016additive}
Lee, S., Jang, C., Moon, S., Cho, J., and Lee, B. (2016).
\newblock Additive light field displays: Realization of augmented reality with
  holographic optical elements.
\newblock {\em ACM Trans. Graph.}, 35(4).

\bibitem[Levoy and Hanrahan, 1996]{Levoy1996}
Levoy, M. and Hanrahan, P. (1996).
\newblock Light field rendering.
\newblock In {\em Proceedings of the 23rd Annual Conference on Computer
  Graphics and Interactive Techniques}, SIGGRAPH '96, pages 31--42, New York,
  NY, USA. ACM.

\bibitem[Li et~al., 2016]{li2016saliency}
Li, N., Ye, J., Ji, Y., Ling, H., and Yu, J. (2016).
\newblock Saliency detection on light field.
\newblock {\em IEEE Transactions on Pattern Analysis and Machine Intelligence},
  39(8):1605--1616.

\bibitem[{Matysiak} et~al., 2020]{matysiak2020high}
{Matysiak}, P., {Grogan}, M., {Le Pendu}, M., {Alain}, M., {Zerman}, E., and
  {Smolic}, A. (2020).
\newblock High quality light field extraction and post-processing for raw
  plenoptic data.
\newblock {\em IEEE Transactions on Image Processing}.

\bibitem[Ng et~al., 2005]{ng2005light}
Ng, R., Levoy, M., Br{\'e}dif, M., Duval, G., Horowitz, M., Hanrahan, P.,
  et~al. (2005).
\newblock Light field photography with a hand-held plenoptic camera.
\newblock {\em Computer Science Technical Report CSTR}, 2(11):1--11.

\bibitem[Overbeck et~al., 2018]{overbeck2018welcome}
Overbeck, R.~S., Erickson, D., Evangelakos, D., and Debevec, P. (2018).
\newblock Welcome to light fields.
\newblock In {\em ACM SIGGRAPH 2018 Virtual, Augmented, and Mixed Reality},
  SIGGRAPH ’18, New York, NY, USA. Association for Computing Machinery.

\bibitem[Parkhurst et~al., 2002]{parkhurst2002modeling}
Parkhurst, D., Law, K., and Niebur, E. (2002).
\newblock Modeling the role of salience in the allocation of overt visual
  attention.
\newblock {\em Vision research}, 42(1):107--123.

\bibitem[Reinagel and Zador, 1999]{reinagel1999natural}
Reinagel, P. and Zador, A.~M. (1999).
\newblock Natural scene statistics at the centre of gaze.
\newblock {\em Network: Computation in Neural Systems}, 10(4):341--350.

\bibitem[Rerabek and Ebrahimi, 2016]{Rerabek:218363}
Rerabek, M. and Ebrahimi, T. (2016).
\newblock New light field image dataset.
\newblock In {\em 8th International Conference on Quality of Multimedia
  Experience (QoMEX)}.

\bibitem[Sheng et~al., 2016]{sheng2016relative}
Sheng, H., Zhang, S., Liu, X., and Xiong, Z. (2016).
\newblock Relative location for light field saliency detection.
\newblock In {\em 2016 IEEE International Conference on Acoustics, Speech and
  Signal Processing (ICASSP)}, pages 1631--1635. IEEE.

\bibitem[{SR Research}, 2016]{Eyelink2016}
{SR Research} (2016).
\newblock Eyelink 1000.
\newblock \url{http://www.sr-research.com/eyelink1000.html}.
\newblock [online].

\bibitem[Vaish and Adams, 2008]{StanfordLF}
Vaish, V. and Adams, A. (2008).
\newblock The (new) stanford light field archive.
\newblock \url{http://lightfield.stanford.edu}.
\newblock [online].

\bibitem[Wang et~al., 2019]{wang2019deep}
Wang, T., Piao, Y., Li, X., Zhang, L., and Lu, H. (2019).
\newblock Deep learning for light field saliency detection.
\newblock In {\em Proceedings of the IEEE International Conference on Computer
  Vision}, pages 8838--8848.

\bibitem[Yarbus, 2013]{yarbus2013eye}
Yarbus, A.~L. (2013).
\newblock {\em Eye movements and vision}.
\newblock Springer.

\bibitem[{Zhang} et~al., 2020]{Zhang2020}
{Zhang}, J., {Liu}, Y., {Zhang}, S., {Poppe}, R., and {Wang}, M. (2020).
\newblock Light field saliency detection with deep convolutional networks.
\newblock {\em IEEE Transactions on Image Processing}, 29:4421--4434.

\bibitem[Zhang et~al., 2015]{zhang2015saliency}
Zhang, J., Wang, M., Gao, J., Wang, Y., Zhang, X., and Wu, X. (2015).
\newblock Saliency detection with a deeper investigation of light field.
\newblock In {\em IJCAI}, pages 2212--2218.

\bibitem[Zhang et~al., 2017]{zhang2017saliency}
Zhang, J., Wang, M., Lin, L., Yang, X., Gao, J., and Rui, Y. (2017).
\newblock Saliency detection on light field: A multi-cue approach.
\newblock {\em ACM Transactions on Multimedia Computing, Communications, and
  Applications (TOMM)}, 13(3):1--22.

\end{thebibliography}

\end{document}